%% file: main.tex
\documentclass[10pt,twocolumn,letterpaper]{article}

\usepackage[pagenumbers]{cvpr} 

\input{preamble}

\definecolor{cvprblue}{rgb}{0.21,0.49,0.74}
\usepackage[pagebackref,breaklinks,colorlinks,allcolors=cvprblue]{hyperref}

\def\confName{CVPR}
\def\confYear{2027}

\usepackage{cuted}
\usepackage{capt-of}

\usepackage{graphicx}
\usepackage[table]{xcolor}
\usepackage{booktabs, adjustbox}

\usepackage{multirow}

\providecommand{\fup}[1]{\textsubscript{\textcolor{green!50!black}{$\uparrow$#1}}}
\providecommand{\fdown}[1]{\textsubscript{\textcolor{red!70!black}{$\downarrow$#1}}}
\providecommand{\fstd}[1]{{\scriptsize$\pm$#1}}   

\ifdefined\fithalves\else
\newsavebox\fitboxA\newsavebox\fitboxB\newdimen\fitmax
\newcommand{\fithalves}[2]{\sbox\fitboxA{#1}\sbox\fitboxB{#2}%
  \fitmax=\wd\fitboxA\relax\ifdim\wd\fitboxB>\fitmax\fitmax=\wd\fitboxB\relax\fi
  \ifdim\fitmax>\linewidth
    \leavevmode\resizebox{\dimexpr\wd\fitboxA*\numexpr\linewidth\relax/\numexpr\fitmax\relax\relax}{!}{\usebox\fitboxA}\par\medskip
    \leavevmode\resizebox{\dimexpr\wd\fitboxB*\numexpr\linewidth\relax/\numexpr\fitmax\relax\relax}{!}{\usebox\fitboxB}\par
  \else\leavevmode\usebox\fitboxA\par\medskip\leavevmode\usebox\fitboxB\par\fi}
\fi

\title{Projective Steering of In Situ-Identified Attention Heads\\Improves Directional Relation Prediction in Vision-Language Models}

\author{Patricia Stöhr\\
Ulm University\\
{\tt\small patricia.stoehr@uni-ulm.de}
\and
Hannah Kniesel\\
Ulm University\\
{\tt\small hannah.kniesel@uni-ulm.de}
\and
Timo Ropinski\\
Ulm University\\
{\tt\small timo.ropinski@uni-ulm.de}
}

\begin{document}
\maketitle
 
\begin{strip}
     \centering
     \includegraphics[width=\textwidth]{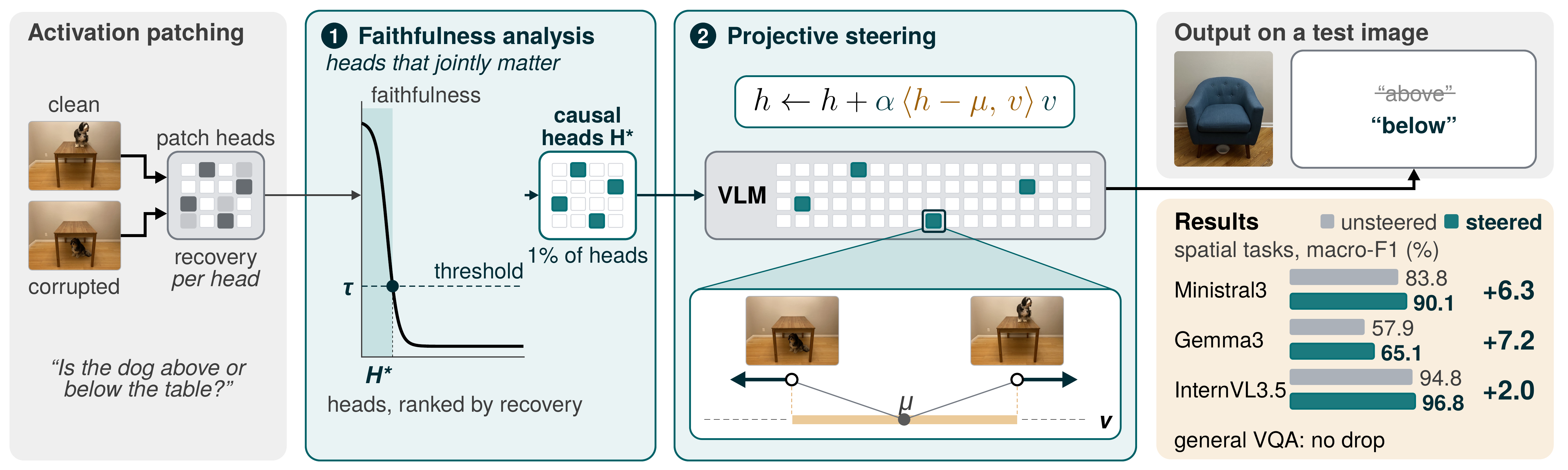}
     \captionof{figure}{\textbf{Overview.} We combine activation patching with a faithfulness analysis to identify, in situ, causally relevant attention heads to steer, and introduce projective steering, which amplifies a head's mean-centered activation along a spatially discriminative direction. Steering causally important heads improves the VLMs' directional relation prediction without harming the general VQA capabilities.\\}
     \label{fig:teaser}
\end{strip}

\input{sec/0_abstract}  
\input{sec/1_intro}
\input{sec/2_related_work}
\input{sec/3_methods}
\input{sec/4_experiments}
\input{sec/5_discussion}
\input{sec/6_conclusion}
\input{sec/7_LLM_usage_statement}
{
    \small
    \bibliographystyle{ieeenat_fullname}
    \bibliography{main}
}

\input{sec/X_suppl}

\end{document}

%% file: preamble.tex
\usepackage{microtype}



%% file: sec/0_abstract.tex
\begin{abstract}
Despite their strong general capabilities, vision-language models (VLMs) struggle to accurately predict directional relations such as left of, right of, above, below, in front of, and behind. Steering offers a lightweight way to improve model behavior at inference time without expensive weight optimization by shifting internal activations of selected model components (e.g., attention heads). Its effectiveness depends on finding suitable components that influence the desired behavior. Therefore, probing is commonly used to identify steering targets. However, this probing-steering pipeline has two weaknesses. First, probing establishes correlation, not causation: a concept being decodable from a representation does not necessarily influence the final answer. Second, most steering methods add a fixed vector to the activations regardless of the input, which can degrade output quality through over- or under-steering. In this paper, we are the first to leverage activation patching to reliably identify causally important attention heads to steer, and we propose projective steering, which amplifies a head's mean-centered activation along a spatially discriminative direction. By combining activation patching with a faithfulness analysis, we find that 1\% of the attention heads in the language model decoder drive directional relation prediction. Steering these heads improves accuracy on directional relation tasks by 1.4, 6.5, and 7.9 percentage points on average for InternVL3.5, Ministral3, and Gemma3, respectively, without degrading general visual question answering performance. Our projective steering outperforms all evaluated state-of-the-art inference-time methods designed for spatial tasks. Thus, it offers a minimally invasive, training-free, yet effective approach to mitigate directional relation failures in VLMs. More broadly, these findings suggest that steering targets should be chosen by causal relevance rather than decodability alone, and that steering should adapt to each input rather than applying a uniform shift.
\end{abstract}

%% file: sec/1_intro.tex
\section{Introduction}
\label{sec:intro}
Understanding object positions and their spatial relations is essential for intelligent models that operate in the real world. Despite their strong performance across many domains, vision-language models (VLMs) struggle to correctly predict directional relations (i.e., left vs. right, above vs. below, in front of vs. behind)~\cite{whatsup,picture_worth_thousand_words,eyes_shut,VSR}. When asked about a directional relation between objects in an image, VLMs often hallucinate or rely on learned priors while ignoring the image~\cite{wolf_your_other_left}. These failures pose serious risks in safety-critical applications, such as in autonomous driving, where a VLM must reason accurately about relative spatial positions. Reporting a pedestrian on the wrong side of a car can result in a life-threatening situation. In medical image analysis, a VLM reporting a tumor on the wrong side can cause severe consequences for the patient.

To overcome failure in directional relation prediction, prior work fine-tunes VLMs on newly generated spatial datasets or changes their architecture~\cite{spatialvlm,spatialrgpt,spatialmllm}. However, these approaches require expensive dataset generation and weight optimization. 
In contrast, leveraging knowledge about the model's internal mechanisms enables a targeted, highly efficient intervention at inference time~\cite{adaptvis,same_task_diff_circuits,dual_mechanisms,reading_img,spatial_IDs}. Common tools for understanding and controlling the model's internal representations are probing and steering~\cite{ITI,ICT,attention_in_space}. Probing~\cite{linear_probes} trains a classifier on model activations to evaluate whether they contain information about a particular concept. Steering modifies these activations during inference to control the model’s behavior~\cite{act_add,latent_steering_vectors}. For example, Ma \etal~\cite{attention_in_space} use probing to identify spatial attention heads to steer. However, the probing-steering pipeline has two main limitations. First, probing shows correlation but not necessarily causation~\cite{probing_shortcomings}. For example, probing may show that a spatial position is decodable from a head's activation, but this does not imply that the head contributes to the task or that the VLM uses the decoded information to find an answer. Second, most steering approaches shift a head’s activation by a constant vector independent of the current input~\cite{act_add,steering_llama,ICT,ITI}. Moreover, probing cannot distinguish whether decodable information is visually grounded or a plain language prior. Consequently, steering heads with high probing accuracy can improve performance by amplifying language biases~\cite{decodable_not_grounded}.

To address these limitations, we propose projective steering of causally important attention heads to improve directional relation prediction. We combine activation patching with a faithfulness analysis to identify causally relevant steering targets in situ, i.e., directly in the model's own components. In contrast to methods that train probing classifiers or sparse autoencoders~\cite{ITI,attention_in_space,ICT,marks_faithfulness}, this requires no auxiliary model. Activation patching~\cite{heimersheim_nanda, zhang_best_practices} measures how strongly each attention head influences the model's answer. However, it scores heads individually but does not show whether the highest-scoring heads together form a self-contained circuit.
Therefore, we construct circuits from the $k$ heads with the highest recovery scores and measure faithfulness~\cite{marks_faithfulness, IOI_GPT} (necessity and sufficiency) as a function of the circuit size $k$. We find a sparse set of attention heads, 1\% of the attention heads in the language model (LM) decoder, that drive the directional relation prediction. We steer these heads with projective steering, which amplifies a head's mean-centered activation along a spatially discriminative direction instead of adding a fixed steering vector.
By steering at most 1\% of the attention heads in the LM decoder, projective steering improves the F1 score on directional relation tasks on average by $2.0$, $6.3$, and $7.2$ percentage points for InternVL3.5-4B~\cite{internvl3.5}, Ministral3-3B~\cite{ministral3}, and Gemma3-4B~\cite{gemma3}, respectively. It improves the accuracy on directional relation prediction of InternVL, Ministral, and Gemma on average by $1.4$, $6.5$, and $7.9$ percentage points, respectively. With these results, projective steering outperforms additive steering~\cite{attention_in_space,ITI}, \textsc{ScalingVis}~\cite{adaptvis}, and \textsc{AdaptVis}~\cite{adaptvis}. Moreover, it does not harm the model's performance on the general visual question answering (VQA) datasets VQAv2~\cite{VQA}, GQA~\cite{GQA}, and POPE~\cite{POPE}.\\

\noindent
In summary, our \textbf{main contributions} are
\begin{itemize}
    \item We propose an in situ approach that combines activation patching with faithfulness analysis to identify causally relevant steering targets.
    \item We introduce projective steering, which amplifies a head's mean-centered activation along a spatially discriminative direction.
    \item We provide a novel directional relations dataset including image-question pairs, which can be easily used to investigate directional relation prediction in VLMs.
\end{itemize}

%% file: sec/2_related_work.tex
\section{Related work}
\label{sec:related_work}
\textbf{Mechanistic interpretability of spatial reasoning.}\hspace{0.2cm}VLMs struggle to perform spatial reasoning with natural images~\cite{whatsup,picture_worth_thousand_words,eyes_shut,VSR} as well as with medical images~\cite{wolf_your_other_left, med_img_spatial,6_fingers}. This finding has motivated a growing body of literature to investigate VLMs' spatial reasoning through mechanistic interpretability. The first line of work analyzes where the model looks and which components perform spatial reasoning. Chen \etal~\cite{adaptvis} find that spatial reasoning fails when the model directs its attention to irrelevant image regions. They propose two inference-time methods, \textsc{ScalingVis} and \textsc{AdaptVis}, to improve spatial understanding by adapting the attention distribution over the image. Moreover, Wang \etal~\cite{V-SEAM} and Ma \etal~\cite{attention_in_space} identify a sparse set of attention heads that are important for spatial reasoning. While Wang \etal~\cite{V-SEAM} evaluate the effect of masking attention heads to find relevant heads, Ma \etal~\cite{attention_in_space} use probing to characterize a head’s function. The second line of work investigates spatial information processing. 
The vision encoder establishes positional geometry, fostering the model's spatial perception~\cite{reading_img}. The vision encoder is the main source distributing spatial information across visual tokens, while the LM decoder adds spatial information when it is degraded. While the vision encoder spreads spatial information across all vision tokens, the LM decoder stores this information in object-related tokens~\cite{dual_mechanisms}. Many VLMs use \textit{spatial IDs}, i.e., latent structures that bind spatial information about visual objects to textual object tokens~\cite{spatial_IDs}. After this cross-modal binding, the VLM performs spatial reasoning via text tokens. Relatedly, Nikankin \etal~\cite{same_task_diff_circuits} find that models align image information with corresponding linguistic information in late layers and implement spatial reasoning circuits that are structurally disjoint but functionally equivalent across modalities. 

\textbf{Model steering.}\hspace{0.2cm}Fine-tuning or architectural changes improve VLMs' spatial reasoning but require large datasets and compute-intensive training~\cite{spatialvlm,spatialrgpt,spatialmllm}. In contrast, steering methods adjust a targeted set of model components at inference time without optimizing weights. While prior work proposes steering methods for LMs~\cite{act_add,refusal_steering, ITI,steering_llama} or hallucination mitigation in VLMs~\cite{ICT,latent_space_steering,dynamic_mm_steering}, we focus on the largely unexplored field of steering a VLM’s directional relation prediction. Compared with approaches that change a layer’s activation~\cite{act_add,steering_llama,refusal_steering,latent_space_steering}, our method is less invasive. It modifies activations of a limited set of attention heads, in line with prior work~\cite{ITI,dynamic_mm_steering,ICT,attention_in_space}.
Moreover, existing approaches select model components to steer via hyperparameter search~\cite{act_add,steering_llama,refusal_steering}, activation changes~\cite{dynamic_mm_steering}, or linear probing~\cite{ITI,ICT,attention_in_space}. These approaches do not ensure that the steered targets are causally relevant. We steer only heads that causally affect the model's answer. Most steering methods~\cite{act_add,refusal_steering,steering_llama,ICT,dynamic_mm_steering,latent_space_steering} compute the difference between activations of contrastive inputs, scale the difference by a steering strength, and add the scaled vector to the layer's or head's activations during the forward pass of a new sample. ITI~\cite{ITI} chooses a different approach. ITI extracts the direction along which true and false features are separable, computes the standard deviation of activations along this direction, and shifts a head's activation along this direction by $\alpha$ times the standard deviation, where $\alpha$ is the steering strength. Ma \etal~\cite{attention_in_space} adopt this steering rule for spatial reasoning. All these steering methods shift an activation by adding a fixed, input-independent steering vector. Unlike additive steering, our steering is input-dependent. The sign and magnitude of the applied shift depend on the data sample. For each test sample, our rule amplifies a head's deviation from its mean activation along the head's direction separating opposite spatial positions. Unlike steering a VLM's spatial prediction with \textit{spatial IDs}~\cite{spatial_IDs}, our approach does not inject label information. 

%% file: sec/3_methods.tex
\section{Methods}
\label{sec:methods}
\Cref{sec:target_selection} explains how we identify steering targets in situ.~\Cref{sec:steering} introduces our projective steering method.~\Cref{sec:directional_rel_dataset} describes the novel directional relation datasets.

\subsection{In situ identification of steering targets}
\label{sec:target_selection}
\textbf{Activation patching.}\hspace{0.2cm}We perform activation patching, where we patch each attention head’s output (at all token positions) individually from a clean run into a corrupted run and compute the recovery score~\cite{zhang_best_practices}. If the recovery score is positive, patching the head’s activation restores the clean run’s answer. Thus, the head causally influences the model’s answer. As we investigate directional relation prediction, our input is an image depicting objects in a directional relation combined with a question about their relation (see~\cref{fig:dataset_overview_main}). This image-question pair builds the clean input. In the corrupted run, we ask the same question but provide an image with swapped object positions such that the model gives the opposite answer. Therefore, a change in the answer after patching is attributable to the visual input, and activation patching identifies heads that carry visually grounded information about the directional relation. We aim to find heads that drive directional relation prediction in general, not heads that prefer one specific answer. So, we balance the answer distribution across the clean and corrupted runs. For example, the clean answer is ``left'' in $50\%$ of the cases and ``right'' in the other cases. Furthermore, we restrict activation patching to input pairs for which the model answers both the clean and corrupted runs correctly without patching. When the VLM predicts the same answer for both images, it fails on one of two inputs. In this case, its answer reflects a bias rather than a correct spatial perception. Additionally, failure on the clean input leaves no correct behavior to restore. Failure on the corrupted input means this run already yields the clean answer, and patching cannot produce a meaningful change. If the model fails on both inputs, its answers differ between the images but are inverted, and patching would restore an incorrect prediction.

\textbf{Faithfulness analysis.}\hspace{0.2cm}Activation patching assigns a recovery score to each head. According to these scores, we rank the heads in descending order (i.e., top-$1$ is the head with the highest recovery score). We use the ranking to build circuits iteratively of increasing size. A circuit of size $k$ includes the top-$k$ heads ranked by recovery. For each circuit size $k$, we measure the faithfulness~\cite{marks_faithfulness, IOI_GPT}.

Let $M$ denote the model, $C\subseteq M$ the hypothesized circuit (i.e., a task-specific subgraph of the model’s computational graph), $D={x_1,\dots,x_N}$ the dataset, and
$F$ a per-sample metric (here, the logit difference). For a set of heads $H \subseteq M$, we define $F_H(x_i)$ as the value of $F$ when running the model on input $x_i \in D$, while all heads in $M \setminus H$ are ablated. Specifically, $F_M(x_i)$ denotes the value of metric $F$ when running the original, unablated model on input $x_i \in D$, whereas $F_\emptyset(x_i)$ is the result for the completely ablated model.
We use the normalized faithfulness score~\cite{marks_faithfulness}
\begin{equation}
\mathrm{Faithfulness}(H) = \frac{\frac{1}{N}\sum_{i=1}^N \big(F_H(x_i) - F_\emptyset(x_i)\big)}{\frac{1}{N}\sum_{i=1}^N \big(F_M(x_i) - F_\emptyset(x_i)\big)}
\label{eq:faithfulness}
\end{equation}
to evaluate a circuit $C$.
We perform two complementary faithfulness analyses that differ in how $H$ is defined:
\begin{enumerate}
\item \textbf{Necessity} ($H=M\setminus C$): All heads in the circuit $C$ are ablated, while every other head’s output remains unchanged. A faithfulness value close to 0 means the heads in the circuit are necessary because the model’s behavior collapses without them.
\item \textbf{Sufficiency} ($H = C$): All heads, except those in circuit $C$, are ablated. A faithfulness close to 1 indicates that the heads within the circuit are sufficient for the task and restore the full model’s behavior.
\end{enumerate}
In line with related work~\cite{IOI_GPT, marks_faithfulness}, we apply mean ablation, i.e., replacing a head’s output with its mean output, averaged over all data samples. Specifically, we compute two separate means over image and text tokens, which reflect precisely the activation patterns of the two modalities. We choose mean ablation instead of zero ablation (i.e., setting the head’s output to zero). Zero ablation shifts the head’s activations off-distribution and can confound the result~\cite{rushing_self_repair,zero_ablation_DINO}.
The goal is to find a sparse set of heads that is necessary and sufficient, representing an optimal target set for steering.

\subsection{Projective steering}
\label{sec:steering}
We aim to steer directional relation prediction in a controlled, targeted way. Projective steering intervenes on a small set of causally important attention heads during inference. We split the dataset into training and test sets to compute steering parameters on the training set only, and finally evaluate the steering effect on the test set. Our approach uses two steering parameters: the mean $\mu$ and the answer direction $v$ of each head to steer.

Let $h \in \mathbb{R}^{d_{\text{head}}}$ be the output of a given attention head
at a token position, and let $h_i$ denote the output for the $i$-th training
sample of $\mathcal{D} = \{(x_i, y_i)\}_{i=1}^{N}$.
The mean
\begin{equation}
\mu = \frac{1}{N} \sum_{i=1}^{N} h_i
\label{eq:mean}
\end{equation}
is the head’s output averaged over the training samples.
To extract the answer direction $v$, we compute the difference between the head’s average output over samples labeled with the same answer (\eg, left) and the head’s average output over samples labeled with the opposite answer (\eg, right).
For an answer $a$ and its opposite $\bar{a}$, let
$\mathcal{I}_a = \{i \mid y_i = a\}$ and
$\mathcal{I}_{\bar{a}} = \{i \mid y_i = \bar{a}\}$.
Then, the answer direction $v$ is defined as
\begin{equation}
v = \frac{1}{|\mathcal{I}_a|} \sum_{i \in \mathcal{I}_a} h_i
- \frac{1}{|\mathcal{I}_{\bar{a}}|} \sum_{i \in \mathcal{I}_{\bar{a}}} h_i.
\label{eq:direction}
\end{equation}
Additionally, we normalize the answer direction $v$ to unit length. Semantically, $v$ describes the direction along which the head discriminates the two spatial positions in its activation space. The method steers each token type using modality-specific statistics and therefore computes $\mu$ and $v$ separately for image and text tokens.
Projective steering modifies the head’s output during the forward pass of an
unseen test sample via
\begin{equation}
h \leftarrow h + \alpha \langle h - \mu, v \rangle  v,
\label{eq:steering}
\end{equation}
where $\alpha$ denotes the steering strength. The steering rule amplifies the head’s deviation from the mean along $v$. The update depends on the head’s activation for the given input, which makes it more flexible than shifting the activation by a constant vector, as in previous work (see~\cref{sec:related_work}). Although the update depends on the input, it does not inject information about the test labels.

\subsection{Directional relations dataset}
\label{sec:directional_rel_dataset}
We generate three directional relation datasets of increasing complexity (see examples in~\cref{fig:dataset_overview_main}). We design them to be easily used to investigate directional relation prediction in VLMs. Thus, each dataset includes image pairs. Each pair contains an image with two objects in a directional relation and a complementary image in which their positions are interchanged.

\textbf{2D Shapes.}\hspace{0.2cm}We generate synthetic images that visualize two 2D objects in a directional relation on a white background. The objects can have three different shapes (dot, square, or triangle), two different colors (red or blue), and different sizes. We position the objects randomly in the image. Consequently, the model has to reason about the objects’ relative positions to find the correct answer. The dataset includes 300 image pairs per shape.

\textbf{3D Objects.}\hspace{0.2cm}We leverage the code from CLEVR~\cite{clevr} to render images showing two objects in a 3D scene. This dataset contains the same object shapes, materials, and colors as CLEVR. Compared to the 2D shapes dataset, it increases the number of colors from two to eight (red, blue, gray, green, brown, purple, cyan, yellow).  Additionally, the objects can have three different shapes (sphere, cube, or cylinder), varying sizes, and two different materials (metal or rubber). In each image, the positions of the objects, the camera, and the lights are chosen randomly. The dataset includes 300 image pairs per directional relation.

\textbf{WhatsUp.}\hspace{0.2cm}This dataset includes images of real-world objects taken from WhatsUp~\cite{whatsup}. WhatsUp includes 820 images with two objects in a directional relation. Each image has a corresponding image with identical objects in the opposite directional relationship. We group the image pairs by directional relation, resulting in three subsets: 205 left-right, 103 above-below, and 102 front-behind pairs. Additionally, we phrase a question for each image pair instead of using the image caption from the original WhatsUp dataset to ensure consistency across datasets.

Each image pair shares a question about the objects’ directional relation. The 2D dataset covers two directional relations (left-right, above-below), whereas the 3D datasets add a further relation (front-behind). We follow the convention of CLEVR~\cite{clevr} and WhatsUp~\cite{whatsup} and define spatial positions from the viewer’s perspective. All datasets have a balanced answer distribution.

%% file: sec/4_experiments.tex
\section{Experiments}
\begin{figure}[tb]
    \centering
    \includegraphics[width=\linewidth]{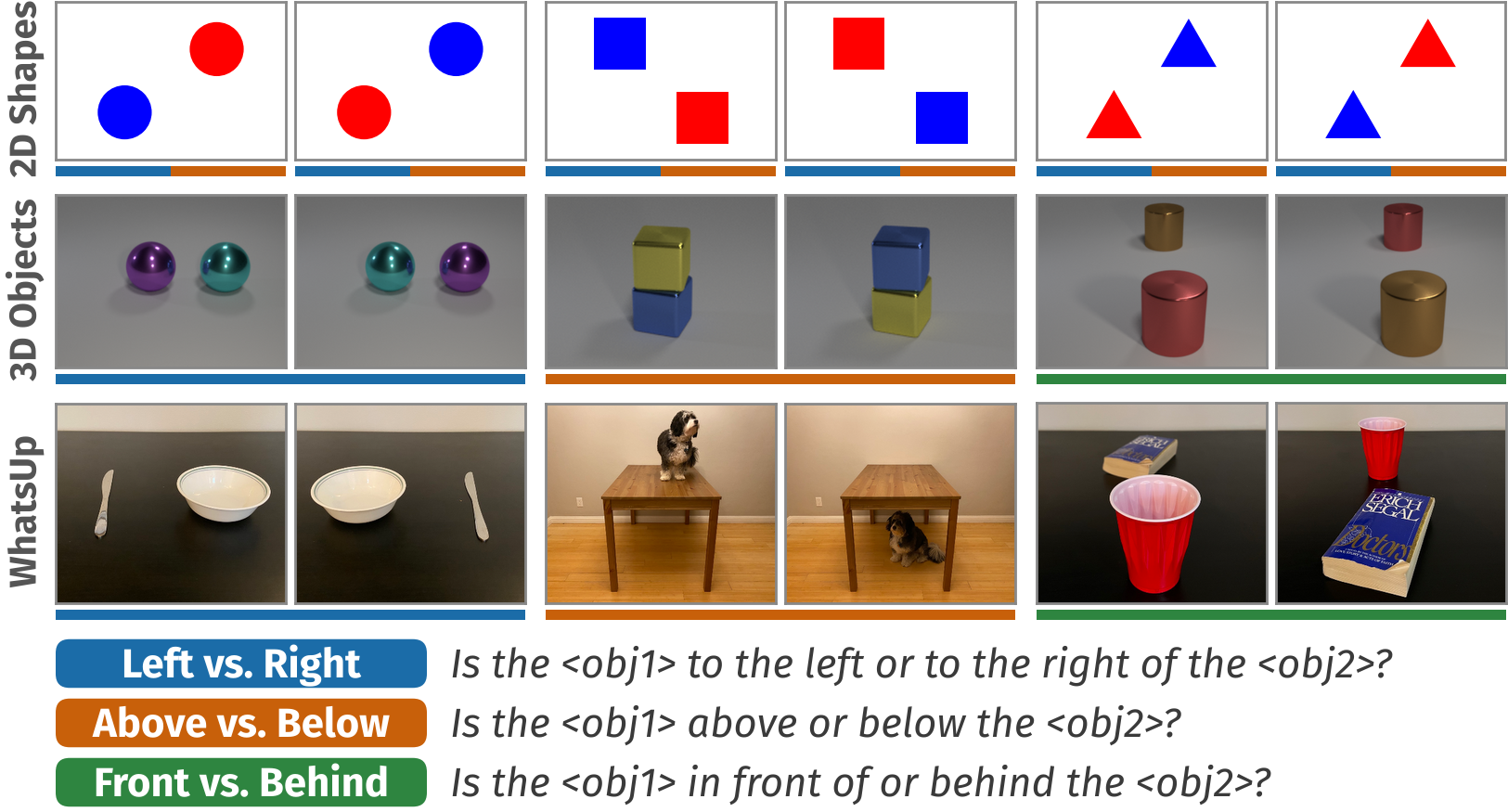}
    \caption{\textbf{Datasets.} We use three datasets of increasing complexity: 2D images of geometric shapes, 3D scenes with geometric objects, and images with real-world objects. Each dataset consists of image pairs: one image showing two objects in a directional relation, and its counterpart with the objects’ positions swapped. Each pair shares a directional-relation question.}
    \label{fig:dataset_overview_main}
\end{figure}
Our experiments select attention heads relevant for directional relation prediction via activation patching and faithfulness analysis. We steer causally important attention heads to improve a model's directional relation prediction. All experiments ran on NVIDIA A100 and H100 GPUs.

\textbf{Models.}\hspace{0.2cm}Our methods require access to attention heads, and activation patching is computationally expensive for a large number of heads. Hence, we choose three open-weight VLMs of manageable size: Ministral3-3B~\cite{ministral3}, Gemma3-4B~\cite{gemma3}, and InternVL3.5-4B~\cite{internvl3.5}. These VLMs are diverse, as the vision encoders and LM decoders differ in module types and the number of layers and heads (see~\cref{tab:vlm_overview}). Nevertheless, they are comparable as they have similar sizes and share a common architecture consisting of a vision encoder, followed by a multi-modal projector and an LM decoder. 

\textbf{Datasets.}\hspace{0.2cm}We utilize our three directional relation datasets (see~\cref{sec:directional_rel_dataset}) for all experiments. Additionally, we use VQAv2~\cite{VQA}, GQA~\cite{GQA}, and POPE~\cite{POPE} to evaluate the steering effect on the VLM's general VQA performance.

\subsection{Activation patching results}
\begin{figure}[b]
  \centering
  \includegraphics[width=\linewidth]{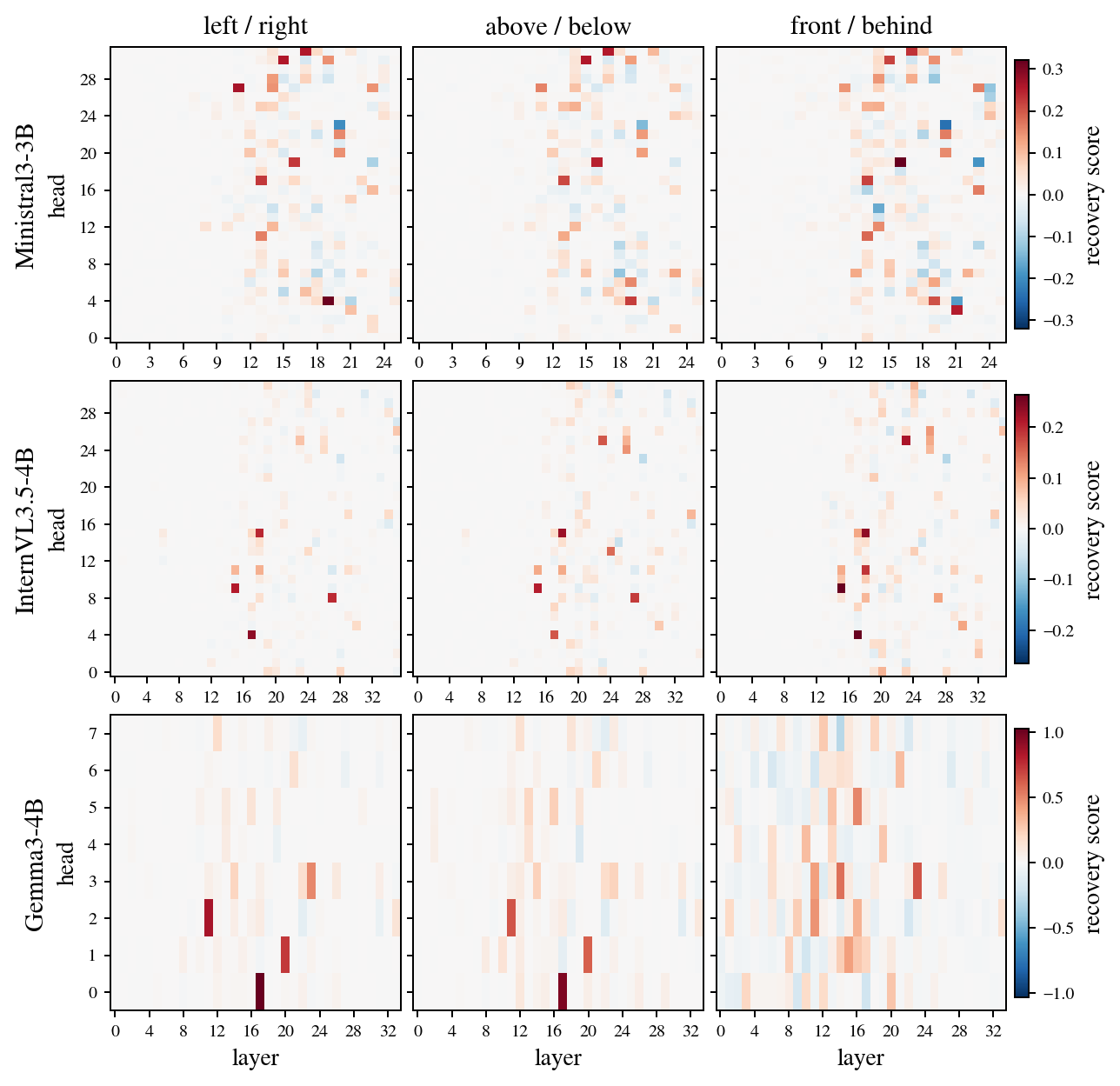}
  \caption{\textbf{Head-wise Activation Patching Results.} Performing activation patching on the attention head's activations identifies heads that restore the correct answer (high recovery scores). Thus, they are causally important for giving the correct answer.}
  \label{fig:head_wise_activation_patching_results}
\end{figure}
We perform activation patching for every attention head in the VLM. However, it does not reveal vision encoder heads that meaningfully influence the model's output (recovery $\approx 0$ for all heads).
In contrast, activation patching identifies several LM decoder heads that are causally important. For each model and task, we average the heads' recovery scores over the datasets to get a recovery score for every head per spatial axis (horizontal axis: left vs. right, vertical axis: above vs. below, depth axis: front vs. behind). We check that the average does not obscure a trend (the same heads have high recovery on each dataset).~\Cref{fig:head_wise_activation_patching_results} shows the heads' recovery scores in one heatmap per model and spatial axis. Heads highlighted in red achieve a positive recovery score, which means patching their activation restores the clean run's answer. The heatmaps also color some heads in blue (negative recovery score). These heads lead the model to prefer the wrong answer. The heatmaps show many heads with a weak impact and only a few with high recovery scores. Comparing the three VLMs shows that a single head’s impact decreases with increasing number of heads in the LM decoder. The LM decoder has $272$, $832$, and $1152$ heads in Gemma, Ministral, and InternVL, respectively. The recovery scores lie between $-0.28$ to $1.0$ in Gemma, $-0.24$ to $0.32$ in Ministral, and $-0.08$ to $0.26$ in InternVL. Several heads achieve high recovery scores in all tasks. These heads lie in the middle to late layers of the LM decoder. We compare the ten heads with the highest recovery scores in Ministral for each spatial axis and find six heads that appear in the top$10$ for every axis: L13H17, L15H30, L16H19, L17H31, L19H4, L20H22. We call them \textbf{universal directional relation heads}, as they contribute to correct directional relation predictions regardless of the axis. We also find universal directional relation heads in Gemma (L11H2, L14H3, L16H5, L21H6, L23H3) and InternVL (L15H9, L17H4, L18H11, L18H15, L23H25, L27H8). 

\subsection{Faithfulness analysis results}
\label{subsec:faithfulness_experiment}
\begin{figure}[tb]
  \centering
  \includegraphics[width=\linewidth]{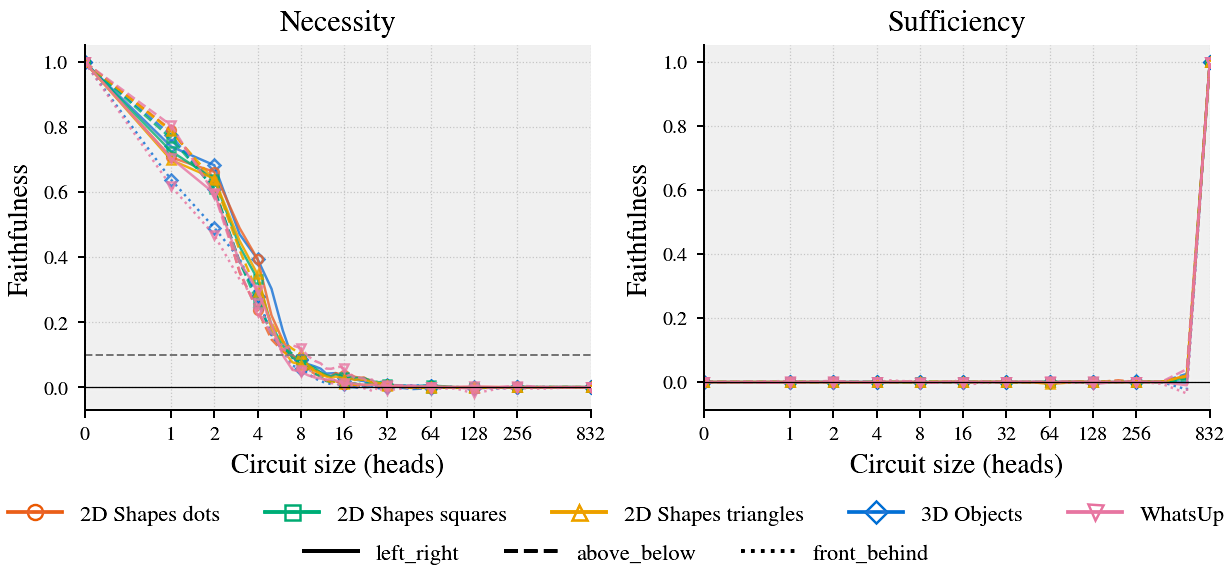}
  \caption{\textbf{Faithfulness Analysis of Ministral.} When ablating the top$7$ heads ranked by recovery, the model retains less than $10\%$ of its original performance, i.e., the top$7$ are necessary for the behavior (left panel). Ablating all heads, except the top$7$, is not sufficient to restore the behavior (right panel).}
  \label{fig:faithfulness_ministral}
\end{figure}

Activation patching identifies attention heads that are causally relevant for correct directional relation prediction (red heads in~\cref{fig:head_wise_activation_patching_results}). We then perform the faithfulness analysis as described in~\cref{sec:target_selection}. A circuit of size $k$ includes the top-$k$ heads ranked by recovery. We evaluate necessity and sufficiency on the same circuit sizes to ensure direct comparison of both metrics at each size. We add attention heads to the circuit in order of decreasing recovery. So, the effect of each additional head diminishes as the circuit size increases. Hence, we first evaluate every size from $k = 1$ to $k = 32$. After that, we apply an approximately geometric grid consisting of the powers of two and their midpoints, $k \in \{48, 64, 96, 128, 192, 256, \dots\}$, until the circuit consists of all heads of the LM decoder. These sizes resolve both curves evenly on a logarithmic scale while keeping the number of evaluations logarithmic rather than linear in the number of heads. Necessity is trivially $1$ for $k=0$ and $0$ for $k=n_{total}$, whereas sufficiency is trivially $0$ for $k=0$ and $1$ for $k=n_{total}$.~\cref{fig:faithfulness_ministral} shows the faithfulness results of Ministral as an example.

\textbf{Necessity.}\hspace{0.2cm}The necessity evaluation shows similar trends for all datasets. The faithfulness decreases steadily as the circuit size increases. Ablating as few as 7 heads in Ministral ($0.8\%$ of all heads in the LM decoder) reduces faithfulness below $0.1$, which means the model retains less than $10\%$ of its original performance (marked with a dotted horizontal gray line in~\cref{fig:faithfulness_ministral}). Above this threshold, the faithfulness decreases fast, and afterwards ablating more heads has a less drastic impact. This finding also holds for Gemma and InternVL. We define $k_{10}$ as the smallest circuit whose ablation removes $\geq90\%$ of the logit difference effect, i.e., faithfulness lies below $0.1$ at this size and at every larger size sampled.~\Cref{tab:circuit-k10} lists $k_{10}$ for all models and datasets. For Ministral, both the mean and median $k_{10}$ across datasets are $7$. This finding also holds for the other VLMs, as the median $k_{10}$ across datasets ranges from $0.4\%$ to $1.0\%$ of all attention heads in the LM decoder (Gemma: $0.4\%$, Ministral: $0.8\%$, InternVL: $1.0\%$). 

\textbf{Sufficiency.}\hspace{0.2cm}The sufficiency values show the same trend for all models (see right panels in~\cref{fig:faithfulness_ministral},~\cref{fig:faithfulness_gemma},~\cref{fig:faithfulness_internvl}). Retaining only the circuit and ablating every remaining head yields a sufficiency of zero for most circuit sizes. Sufficiency lies above $0.5$ only when $71-100\%$ of the heads are not ablated. Thus, a circuit is sufficient only when it is nearly the whole model.

In summary, the sufficiency analysis reveals \textbf{no small stand-alone circuit} implementing directional relation prediction. The computation is spread across the model. However, \textbf{a sparse set of heads is strongly necessary}. Ablating at most $1.0\%$ of the heads in the LM decoder (those with the highest recovery scores) leads to a collapsed directional relation prediction behavior, which the remaining heads cannot compensate for. These two findings guide our steering target selection. We aim for a controlled, minimally invasive steering. So, the necessary heads enable an efficient and effective intervention. As the prediction depends on them, steering them should change the prediction, while a large set of remaining heads stays untouched. Steering the sufficient set would intervene on at least $70\%$ of all heads, which is not targeted, more costly, and likely to cause side effects such as degraded performance on other tasks.

\ifdefined\dcolL\else\newlength{\dcolL}\newlength{\dcolR}\fi
\providecommand{\dcell}[3]{\makebox[\dcolL][r]{#1{#2}}\makebox[\dcolR][l]{#1{.#3}}}
\providecommand{\dplain}[1]{#1}
\providecommand{\dpos}[1]{\textcolor{green!50!black}{#1}}
\providecommand{\dneg}[1]{\textcolor{red!70!black}{#1}}
\providecommand{\dposB}[1]{\textbf{\dpos{#1}}}
\providecommand{\dposU}[1]{\underline{\smash{\dpos{#1}}\vphantom{0}}}
\providecommand{\Dgain}[2]{\dcell{\dpos}{+#1}{#2}}
\providecommand{\DgainB}[2]{\dcell{\dposB}{+#1}{#2}}
\providecommand{\DgainU}[2]{\dcell{\dposU}{+#1}{#2}}
\providecommand{\Dloss}[2]{\dcell{\dneg}{$-$#1}{#2}}
\providecommand{\Dzero}{\dcell{\dplain}{0}{0}}
\providecommand{\dbold}[1]{\textbf{#1}}
\providecommand{\dul}[1]{\underline{\smash{#1}\vphantom{0}}}
\providecommand{\Pgain}[2]{\dcell{\dplain}{+#1}{#2}}
\providecommand{\PgainB}[2]{\dcell{\dbold}{+#1}{#2}}
\providecommand{\PgainU}[2]{\dcell{\dul}{+#1}{#2}}
\providecommand{\Ploss}[2]{\dcell{\dplain}{$-$#1}{#2}}

\ifdefined\fitwidth\else\newsavebox\fitwidthbox
\newcommand{\fitwidth}[1]{\sbox\fitwidthbox{#1}\ifdim\wd\fitwidthbox>\linewidth\resizebox{\linewidth}{!}{\usebox\fitwidthbox}\else\usebox\fitwidthbox\fi}\fi

\begin{table*}[tb]\centering
\caption{\textbf{Change in F1 (in percentage points)} on the directional relation test set relative to each model's unsteered baseline (Gemma3-4B, InternVL3.5-4B, Ministral3-3B). Largest gain among the four methods per column and model in bold, second largest underlined (improvements only). N/A: shift steering was not run because every train stimulus is answered correctly and shift steering is undefined in this case. \textit{Average} is the mean over the eight preceding columns. On average, our steering method achieves the highest performance gains for all VLMs. L-R: left-right, A-B: above-below, F-B: front-behind.}
\label{tab:f1-delta-main}
\settowidth{\dcolL}{\textbf{+00}}\settowidth{\dcolR}{\textbf{.0}}
\fitwidth{%
\begin{tabular}{ll cc ccc ccc c}\toprule
 & & \multicolumn{2}{c}{2D Shapes} & \multicolumn{3}{c}{3D Objects} & \multicolumn{3}{c}{WhatsUp} & \\
\cmidrule(lr){3-4}\cmidrule(lr){5-7}\cmidrule(lr){8-10}
Method & Model & L--R & A--B & L--R & A--B & F--B & L--R & A--B & F--B & \textit{Average} \\
\midrule
\textsc{ScalingVis}~\cite{adaptvis} (ICML 2025) & Gemma & \PgainU{1}{0} & \PgainU{17}{6} & \PgainU{1}{2} & \PgainB{15}{1} & \PgainU{1}{5} & \Dzero & \PgainU{2}{6} & \PgainB{8}{4} & \Dgain{5}{9} \\
 & InternVL & \Dzero & \Dzero & \Dzero & \Dzero & \Dzero & \Dzero & \PgainU{1}{2} & \PgainU{1}{5} & \DgainU{0}{3} \\
 & Ministral & \Ploss{1}{8} & \Dzero & \Dzero & \Dzero & \Dzero & \Ploss{0}{6} & \Pgain{2}{2} & \PgainU{2}{9} & \Dgain{0}{3} \\
\midrule
\textsc{AdaptVis}~\cite{adaptvis} (ICML 2025)& Gemma & \Pgain{0}{9} & \Pgain{17}{5} & \PgainU{1}{2} & \PgainU{15}{0} & \PgainB{8}{3} & \PgainU{0}{1} & \Pgain{1}{8} & \PgainB{8}{4} & \DgainU{6}{7} \\
 & InternVL & \Dzero & \Dzero & \Dzero & \Dzero & \Dzero & \Dzero & \PgainU{1}{2} & \PgainU{1}{5} & \DgainU{0}{3} \\
 & Ministral & \Ploss{1}{8} & \Dzero & \Dzero & \Dzero & \Dzero & \PgainU{0}{1} & \PgainU{2}{9} & \Pgain{2}{5} & \DgainU{0}{5} \\
\midrule
Shift Steering~\cite{attention_in_space,ITI} (NeurIPS 2023)& Gemma & \Ploss{0}{2} & \Ploss{1}{6} & \Ploss{0}{6} & \Ploss{0}{3} & \Ploss{0}{1} & \Ploss{0}{4} & \Dzero & \Dzero & \Dloss{0}{4} \\
 & InternVL & N/A & N/A & N/A & N/A & N/A & N/A & \Ploss{0}{3} & \Dzero & \Dloss{0}{2} \\
 & Ministral & \Ploss{0}{8} & \Ploss{0}{1} & N/A & N/A & N/A & \PgainB{0}{8} & \Ploss{0}{4} & \Ploss{0}{6} & \Dloss{0}{2} \\
\midrule
\rowcolor{gray!15}Projective Steering (Ours) & Gemma & \PgainB{9}{8} & \PgainB{19}{1} & \PgainB{5}{4} & \Pgain{5}{8} & \Ploss{0}{1} & \PgainB{5}{2} & \PgainB{12}{1} & \Dzero & \DgainB{7}{2} \\
\rowcolor{gray!15} & InternVL & \Dzero & \Ploss{0}{1} & \Dzero & \Ploss{0}{1} & \Dzero & \Ploss{0}{2} & \PgainB{12}{6} & \PgainB{4}{1} & \DgainB{2}{0} \\
\rowcolor{gray!15} & Ministral & \PgainB{20}{4} & \PgainB{0}{2} & \Dzero & \Ploss{2}{3} & \Dzero & \Ploss{0}{5} & \PgainB{8}{6} & \PgainB{24}{0} & \DgainB{6}{3} \\
\bottomrule\end{tabular}%
}
\end{table*}

\subsection{Steering results on directional relation tasks}
\label{sec:experiment_steering}
For each model and spatial axis, we steer the median $k_{10}$ heads, which are only $1$, $7$, and $12$ heads for Gemma, Ministral, and InternVL, respectively. The supplementary material lists the steered heads.
We steer the selected heads using~\cref{eq:steering} and compare our projective steering with shift steering~\cite{attention_in_space,ITI}, \textsc{ScalingVis}~\cite{adaptvis}, and \textsc{AdaptVis}~\cite{adaptvis}. For projective steering, we compute the steering parameters $\mu$ and $v$ on a training set. Therefore, we randomly split each dataset into equally sized halves. As the datasets consist of image pairs, we split the data so that both members of a pair are in the training set or the test set to avoid data leakage. Based on the hyperparameter sweep, we choose $\alpha_{proj}=2.75$ for projective steering and $\alpha_{shift}=0.1$ for shift steering (see~\cref{sec:appendix_steering}). We repeat each experiment five times (with five random seeds) to compute the mean and standard deviation (std) of accuracy and F1 score.~\Cref{tab:f1-delta-main} presents the change in F1 scores across all models and datasets (detailed results in supplementary material).

Without steering, Ministral has on average an accuracy of $85\%$ and an F1 score of $83.8\%$ on the directional relation tasks. InternVL achieves the highest unsteered scores on average (accuracy: $96.5\%$, F1: $94.8\%$), whereas Gemma performs the worst on the directional relation tasks (accuracy: $69.4\%$, F1: $57.9\%$). We do not apply the shift steering for datasets with an unsteered score of $100\%$, since this steering rule is undefined when the model gives no incorrect answers (cf. Eq. (6) in~\cite{attention_in_space}). Projective steering improves Ministral's F1 score by $6.3$ percentage points (pp) on average (min: $-2.3$\,pp, max: $24.0$\,pp) compared to the unsteered model. Thus, it outperforms \textsc{ScalingVis}, \textsc{AdaptVis}, and shift steering, which increase the unsteered Ministral's F1 score on average by $0.3$\,pp, $0.5$\,pp, and $-0.2$\,pp, respectively. Ministral's accuracy increases on average by $6.5$\,pp (min: $-2.3$\,pp, max: $21.3$\,pp) when applying projective steering, whereas \textsc{ScalingVis}, \textsc{AdaptVis}, and shift steering change the accuracy by $-0.1$, $0.0$, and $-0.3$\,pp, respectively. Furthermore, projective steering achieves, on average, an accuracy gain of $1.4$\,pp (min: $-0.4$\,pp, max: $13.4$\,pp) for InternVL and $7.9$\,pp (min: $-0.1$\,pp, max: $18.8$\,pp) for Gemma. It improves the F1 scores on average by $2.0$\,pp (min: $-0.2$\,pp, max: $12.6$\,pp) for InternVL and $7.2$\,pp (min: $-0.1$\,pp, max: $19.1$\,pp) for Gemma. While \textsc{ScalingVis} enhances Gemma's F1 score by $5.9$\,pp and accuracy by $4.1$\,pp on average, its impact on InternVL is lower (F1: $0.3$\,pp on average, accuracy: $0.2$\,pp on average). The same finding holds for \textsc{AdaptVis}, which increases Gemma's accuracy by $4.1$\,pp and F1 score by $6.7$\,pp, but increases InternVL's accuracy and F1 score by $0.3$\,pp on average. 

Across all VLMs, projective steering is the most successful, followed by \textsc{AdaptVis} and \textsc{ScalingVis}, whereas shift steering is ineffective for every VLM. On directional relation tasks, projective steering achieves a higher F1 score on average for all VLMs than every other method. It outperforms \textsc{ScalingVis} and \textsc{AdaptVis} by $1.3$ and $0.5$ points on Gemma, by $1.7$ points each on InternVL, and by $6.0$ and $5.8$ points on Ministral. Projective steering surpasses shift steering by $6.5$ points on Ministral, $2.2$ points on InternVL, and $7.6$ points on Gemma.

\subsection{Steering results on general VQA tasks}
We evaluate whether projective steering harms the VLMs' performance on general VQA tasks. Therefore, we perform the same experiment as in the previous section, but instead of the directional relation test sets, we test the steering effect on GQA~\cite{GQA}, VQAv2~\cite{VQA}, and POPE~\cite{POPE}. From these datasets, we exclude questions about directional relations, so that they include only general questions about natural images. We steer the same heads and compute the steering parameters $\mu$ and $v$ on the same directional relation training sets. We compare projective steering with shift steering~\cite{attention_in_space,ITI}, \textsc{ScalingVis}~\cite{adaptvis}, and \textsc{AdaptVis}~\cite{adaptvis}. Additionally, we use the same steering strengths ($\alpha_{proj}=2.75$, $\alpha_{shift}=0.1$) as in the evaluation on directional relation tasks. We repeat each experiment five times (with five random seeds) to compute the mean and std of accuracy and F1 score. The supplementary material includes detailed results. 

Shift steering has a negligible effect on the VLMs’ general VQA capabilities. Accuracy and F1 score change by only $-0.5$\,pp to $0.3$\,pp relative to the unsteered baselines. In contrast, \textsc{ScalingVis} and \textsc{AdaptVis} lead to larger performance changes, especially for Gemma. \textsc{ScalingVis} decreases Gemma's F1 scores on average by $-3.3$\,pp (min: $-13.5$\,pp, max: $0.1$\,pp), $-1.0$\,pp (min: $-6.5$\,pp, max: $0.4$\,pp), and $-4.0$\,pp (min: $-14.3$\,pp, max: $0.3$\,pp) on POPE, GQA, and VQAv2, respectively. \textsc{AdaptVis} shows the same trends. \textsc{ScalingVis} and \textsc{AdaptVis} harm Ministral's and InternVL's performance on these VQA datasets less than Gemma. Nevertheless, the observed maximum decrease is higher compared to shift steering (accuracy: $-2.7$\,pp by \textsc{ScalingVis}, $-2.1$\,pp by \textsc{AdaptVis}, F1: $-4.1$\,pp by \textsc{ScalingVis}, $-3.2$\,pp by \textsc{AdaptVis}). On average, projective steering changes Gemma's F1 score by $-0.2$\,pp on POPE, $-0.3$\,pp on GQA, and by $0.1$\,pp on VQAv2. Applying projective steering to InternVL decreases the F1 score on average by $-0.4$\,pp, $-1.7$\,pp, and $-0.5$\,pp on POPE, GQA, and VQAv2, respectively. Lastly, our steering harms Ministral's F1 scores on average by $-0.9$\,pp, $-1.9$\,pp, and $-0.9$\,pp on POPE, GQA, and VQAv2, respectively.

\subsection{Ablation studies}
\label{sec:ablations}
Performing the faithfulness analysis with random head rankings shows that the recovery-ranking results depend on which heads are included and not solely on how many heads a circuit includes. Moreover, we perform two ablation studies for the steering experiments. First, we test other faithfulness thresholds for selecting a small set of necessary attention heads to steer. This study shows that choosing the $k_{10}$ head set (i.e., faithfulness threshold at $0.1$) is a valuable choice. Second, we steer an equally sized set of randomly chosen heads, which does not reproduce our gains. Details are explained in the supplementary material.

%% file: sec/5_discussion.tex
\section{Discussion}
\label{sec:discussion}
Our experiments evaluate projective steering of causally important attention heads in Ministral3-3B, Gemma3-4B, and InternVL3.5-4B across diverse tasks (directional relations and general VQA), revealing several valuable insights. 
Firstly, the experiments show two unexpected outcomes. Although the 2D shapes seem simpler than 3D objects to humans, Ministral makes more mistakes when answering questions about 2D shapes than about 3D objects. One possible explanation is that natural images dominate VLM training data, and rendered 3D objects are more similar to natural images than 2D shapes are.
Additionally, Ministral performs poorly on the horizontal axis of the 2D shapes dataset, while Gemma performs worse on the vertical axis than on the horizontal axis. Hence, difficulty depends more on the VLM than on the data. 
Secondly, the methods exhibit limitations. Shift steering is undefined when the model answers every training sample correctly. On the remaining directional tasks, it has a weak impact on every VLM and thereby trivially does not harm general VQA performance. \textsc{ScalingVis} and \textsc{AdaptVis} often lead the model to distribute its answers more evenly between the two possibilities without improving its performance. The F1 score increases but not the accuracy. For example, Gemma's F1 score rises by $8.4$\,pp on WhatsUp-front/behind, but its accuracy decreases by $0.2$\,pp. 
Projective steering amplifies a direction the head already has. If a head does not have a clear tendency towards an answer (i.e., its current activation is near its mean activation, $h-\mu \approx 0$), projective steering modifies the head's activation negligibly. Consequently, projective steering cannot steer the model towards the correct answer when the steered heads carry noisy signals, but this also prevents projective steering from amplifying noise.
Besides this limitation, projective steering achieves the most promising results. It outperforms the other approaches across all VLMs and directional relations in terms of accuracy and F1 score. Its positive impact generalizes across spatial axes. In contrast, \textsc{ScalingVis} and \textsc{AdaptVis} often enhance the vertical axis more than the others. Additionally, projective steering has small side effects on general VQA performance. Projective steering is the most effective approach and also highly efficient. While \textsc{ScalingVis} and \textsc{AdaptVis} intervene on all heads in the LM decoder, our approach steers at most $1\%$ of the LM decoder's attention heads. Furthermore, our steering gains are head-specific, as steering an equally sized set of randomly chosen heads does not reproduce these gains. This finding further supports that our mechanistic interpretability tools find steering targets with causal evidence. 
Leveraging knowledge about the model's internal mechanisms enables a controlled, highly efficient intervention. It improves a specific task without harming general performance on other tasks and is feasible during inference without weight optimization. Together with recent findings on weight-space programming~\cite{WSP}, our results add to growing evidence that knowledge of a model’s internal mechanisms can enable targeted, gradient-free interventions as an alternative to fine-tuning. While this work focuses on attention heads, investigating MLPs is an interesting avenue for future work. 

%% file: sec/6_conclusion.tex
\section{Conclusion}
\label{sec:conclusion}
We combine activation patching with a faithfulness analysis to select a small set of steering targets with a causal effect on directional relation prediction and introduce projective steering, which amplifies a head's mean-centered activation along a spatially discriminative direction. Combining activation patching with a faithfulness analysis yields a sparse set of attention heads ($1\%$ of the attention heads in the LM decoder) that are strongly necessary for directional relation tasks. We steer only this small set of causally important heads with projective steering. Our results show that projective steering improves the accuracy of InternVL3.5-4B, Ministral3-3B, and Gemma3-4B on directional relation tasks by $1.4$\,pp, $6.5$\,pp, and $7.9$\,pp on average, respectively. The F1 scores increase on directional relation tasks by $2.0$\,pp, $6.3$\,pp, and $7.2$\,pp on average for InternVL3.5-4B, Ministral3-3B, and Gemma3-4B, respectively. Additionally, our steering approach does not harm performance on general VQA datasets.

\section*{Acknowledgments}
This work was funded by the Deutsche Forschungsgemeinschaft (DFG, German Research Foundation) – GRK 3012 – 520750254 (kemai.uni-ulm.de). The authors acknowledge support by the state of Baden-Württemberg through bwHPC.

%% file: sec/7_LLM_usage_statement.tex
\section*{LLM Usage Statement}
\label{sec:LLM_usage}
The authors take full responsibility for all submitted content, regardless of any LLM or AI tool use. 
In this work, we used Claude (Opus and Fable) as well as Claude Code (Opus and Fable), as described below.

\noindent
\textbf{Research ideas, hypotheses, and conclusions.} No LLMs or generative AI
tools were used in developing the research ideas or hypotheses, which are
entirely the authors' own. Claude was used only to refine or sharpen the formulation of
author-written claims and conclusions.

\noindent
\textbf{Related work and citations.} The authors identified part of the cited works themselves. Furthermore, the authors used Claude to find papers related to this work. Claude proposed a list of related literature, from which the authors selected relevant works and read them before citing.
The authors selected the cited works and checked each reference against its source. All BibTeX entries were collected manually.

\noindent
\textbf{Writing.} The authors used Claude only to edit author-written text for grammar, clarity, and structure throughout the paper. ChatGPT was also used to check and edit grammar. The authors checked and approved every proposed edit. Claude provided LaTeX code for creating tables. The authors refined the proposed table layout. The correctness of table entries was manually checked by the authors. 

\noindent
\textbf{Figures.} The overview and dataset figures were generated with Claude based on the authors' ideas. For generating figures that visualize experimental results, 
Claude Code wrote code to extract results from CSV files, to compute summary statistics, and to generate matplotlib figures. The code was reviewed and edited
by the authors.

\noindent
\textbf{Datasets.} Claude Code wrote the code for generating the synthetic 2D Shapes images.
It modified the implementation provided by CLEVR~\cite{clevr} to generate our 3D Objects dataset. Additionally, it
generated JSON files storing information about all used images (\eg, visible objects and their positions).
The authors reviewed the code that generates images and inspected a representative sample of all image-text items. 

\noindent
\textbf{Code and implementation.} Claude Code carried out multi-step implementation tasks with limited step-by-step direction from the
authors. The authors specified the method and the tasks. The authors reviewed, tested, and edited the code.

\noindent
\textbf{Experiments.} Claude Code autonomously launched and monitored experimental runs. It also collected results into tables or figures. The authors defined the experimental protocol, models, datasets, hyperparameters, evaluation metrics, and verified the reported numbers against the raw logs. The authors interpreted the results and inferred conclusions.

%% file: sec/X_suppl.tex
\clearpage
\setcounter{page}{1}
\maketitlesupplementary
\section{Models}
\label{sec:appendix_models}

\begin{table*}[h]
\centering
\caption{\textbf{VLM architectures.} We analyze three VLMs that share a common architecture (vision encoder, multimodal projector, LM decoder) but differ in the module types, number of layers, and number of heads.}
\label{tab:vlm_overview}
\begin{tabular}{l lcc lcc}
\toprule
& \multicolumn{3}{c}{\textbf{Vision encoder}} & \multicolumn{3}{c}{\textbf{LM decoder}} \\
\cmidrule(lr){2-4} \cmidrule(lr){5-7}
\textbf{Model} & Type & Layers & Heads & Type & Layers & Heads (Q/KV) \\
\midrule
Ministral3-3B~\cite{ministral3}   & Pixtral ViT            & 24 & 16 & Ministral3  & 26 & 32/8 \\
Gemma3-4B~\cite{gemma3}           & SigLIP-So400m          & 27 & 16 & Gemma3      & 34 & 8/4  \\
InternVL3.5-4B~\cite{internvl3.5} & InternViT-300M         & 24 & 16 & Qwen3       & 36 & 32/8 \\
\bottomrule
\end{tabular}
\end{table*}

In this paper, we utilize three open-weight VLMs downloadable from Hugging Face:
Ministral-3-3B-instruct-2512-bf16~\cite{ministral3}, Gemma-3-4B-it~\cite{gemma3}, and InternVL-3.5-4B-hf~\cite{internvl3.5}.

\Cref{tab:vlm_overview} summarizes the models' architectures. To ensure all models receive consistent messages, we run all models on the same user prompts without a system prompt. Gemma and InternVL support system prompts but do not use one by default. In contrast, Ministral's chat template inserts the checkpoint's default system prompt. Hence, we disable Ministral's system prompt to apply consistent input messages across models. For reproducibility, we use greedy decoding (i.e., set the option `do\_sample’ to False) such that the models' decoding is deterministic. We do not run InternVL in thinking mode. Gemma implements 5-to-1 interleaved attention, consisting of five local attention layers with a sliding window of 1024 tokens, followed by a global attention layer. The local attention never applies in our experiments, as all inputs are shorter than 1024 tokens. Thus, all layers attend over the whole sequence. Activation patching requires that the clean and corrupted input differ only in the aspect being examined (here, the directional relation). Consequently, the number and layout of visual tokens must match in the clean-corrupted pair to avoid desynchronized token indices. Therefore, we ensure that all vision encoders process each image as a single view and pass the same number of image tokens for each image to their corresponding LM decoder. To achieve this, we disable the pan-and-scan mechanism in Gemma, disable the crop-to-patches mechanism in InternVL, and resize every image to $448\times448$ pixels before we feed it into the VLM.

We implement activation patching, ablation, and steering interventions with PyTorch forward hooks, which do not modify the VLMs’ implementations. The model code and weights remain unchanged. A hook can record an input or output of a model component during a forward pass. Alternatively, it can overwrite an input or output during a forward pass. All forward passes run without gradient computation. For activation patching, mean ablations, and steering, the hooks intervene on the input to the attention output projection, where the outputs of the individual heads are concatenated but separable.

\section{Directional relations dataset}
\label{sec:appendix_data}
We generate three datasets of increasing complexity. The first dataset includes 2D images depicting geometric shapes. The second dataset contains 3D scenes with geometric objects, whereas the third dataset includes 3D scenes with real-world objects. We keep the objects and scenes simple because our experiments focus on directional relations rather than object identification.
Each dataset includes pairs of images. Each pair consists of an original image with two objects in a directional relationship and a complementary image in which the objects' positions are interchanged.

\textbf{2D Shapes.}\hspace{0.2cm}We generate synthetic images that visualize two 2D objects in a directional relation (i.e., left-right, above-below) on a white background. The objects can have three different shapes (dot, square, triangle), two different colors (red, blue), and different sizes. Each image depicts two shapes of the same type (\eg, dot) so that they occupy the same space in the original and complementary image. The shapes are distinguishable only by color. We position the objects randomly in the image. Consequently, the model has to compare and reason about the objects' relative positions to find the correct answer. There is a small margin between objects to avoid overlap and ensure unambiguous directional relations. The image size is $448\times448$ pixels. The dataset includes 300 image pairs per shape.

\textbf{3D Objects.}\hspace{0.2cm}This dataset is a modified version of CLEVR~\cite{clevr}. We cannot use the CLEVR dataset directly for activation patching because it does not provide image pairs in which only two objects swap positions while everything else remains unchanged. Moreover, it does not show objects in an above-below relation. Hence, we leverage the CLEVR code to render image pairs showing two objects in a 3D scene. Our dataset contains the same object shapes, materials, and colors as CLEVR. Thus, this dataset is more complex than the first set. Firstly, it introduces an additional directional relation (in front of vs. behind) as we transition from 2D to 3D. Secondly, it increases the number of colors from two to eight (red, blue, gray, green, brown, purple, cyan, yellow). Thirdly, besides three different shapes (sphere, cube, cylinder) and varying sizes, the objects can have two different materials (metal, rubber). Similar to the first dataset, each image shows objects with the same shape and material. The objects differ only by color. In each image, the positions of the objects, the camera, and the lights are chosen randomly. The only placement constraint is that both objects are clearly visible. The image size is $448\times448$ pixels. The dataset includes 300 image pairs per directional relation.

\textbf{WhatsUp.}\hspace{0.2cm}While the previous datasets deal with simple geometric shapes, this dataset increases the complexity by using photos of real-world 3D objects (\eg, furniture, food, \dots). The WhatsUp dataset~\cite{whatsup} includes 820 images with two objects in a directional relation. The background is uniform to reduce noise and distraction. Each image has a corresponding image with identical objects in the opposite directional relationship. Hence, the dataset provides controlled image pairs suitable for our activation patching experiment.
We group the image pairs by directional relation, resulting in three subsets: 205 left-right, 103 above-below, and 102 front-behind pairs. Furthermore, we resize each image to $448\times448$ pixels such that the image size is consistent across all experiments. Additionally, we phrase questions about each image instead of using the provided image captions to ensure consistency across datasets. 

\begin{figure*}
  \centering
  \includegraphics[width=0.85\linewidth]{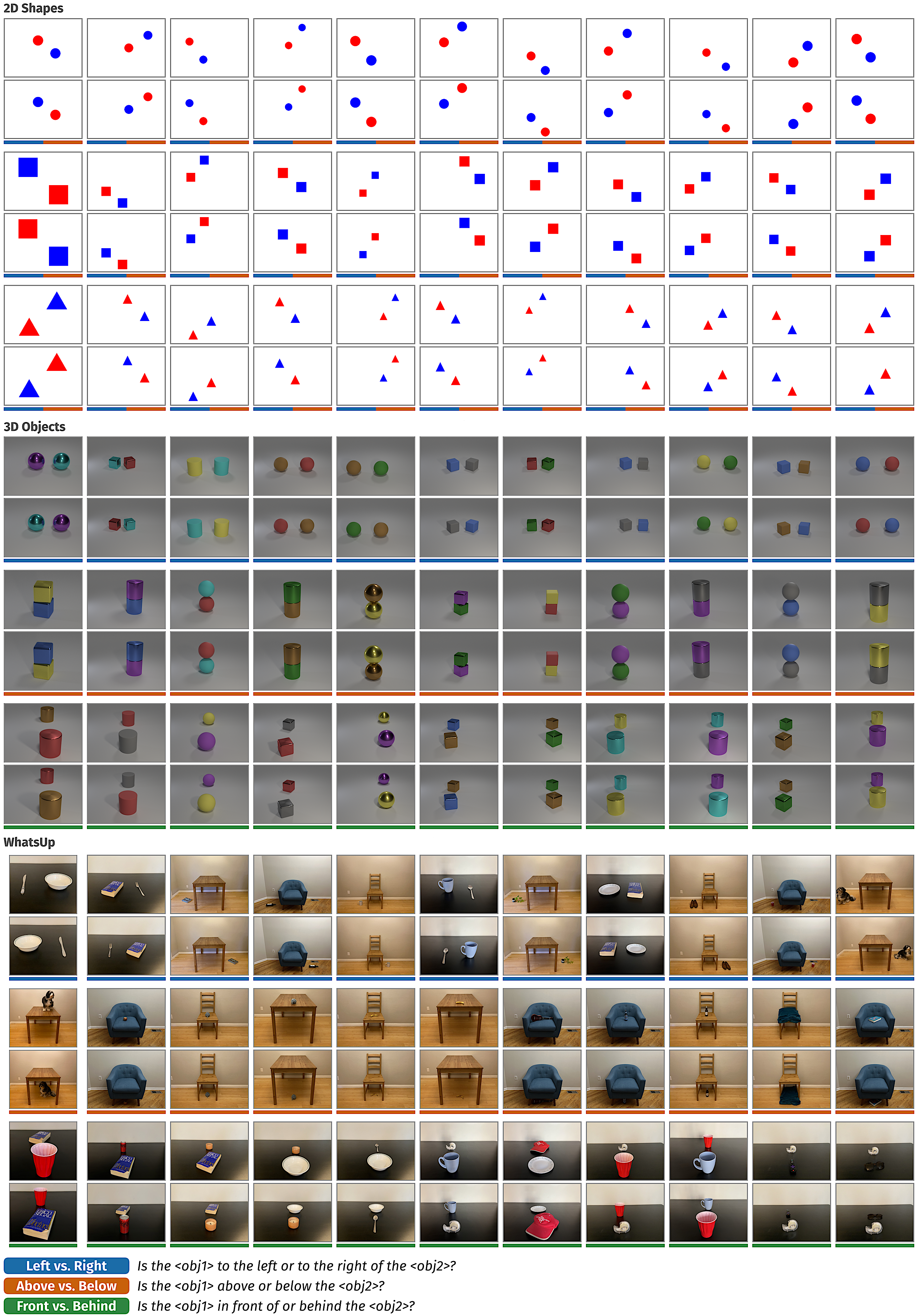}
  \caption{\textbf{Examples from Datasets.} We generate 3 datasets of increasing complexity: 2D images depicting geometric shapes, 3D scenes with geometric objects, and 3D scenes with real-world objects. Each dataset includes pairs of images. A pair consists of an original image with two objects in a directional relationship, and a complementary image in which the two objects' positions are swapped. We formulate a directional relation question for each image pair.}
  \label{fig:datasets}
\end{figure*}

\Cref{fig:datasets} shows some example image pairs from all datasets. Each image pair shares a question about the objects' directional relations. In activation patching, it is essential that only one attribute changes between the clean and corrupted run (either image or text, not both).
For all datasets, we utilize the following prompt templates:

\noindent
\textbf{Left vs. Right:} Look at the image. Is the \{first\} to the left or to the right of the \{second\}? Answer with one word: `left' for left and `right' for right.

\noindent
\textbf{Above vs. Below:} Look at the image. Is the \{first\} above or below the \{second\}? Answer with one word: `above' for above and `below' for below.

\noindent
\textbf{Front vs. Behind:} Look at the image. Is the \{first\} in front of or behind the \{second\}? Answer with one word: `front' for in front and `back' for behind.

The terms \{first\} and \{second\} are placeholders for the objects in the image.
The tokenizers encode `in front of' and `behind' with multiple tokens. Hence, we force the model to answer `front' or `back' to ensure a single-token answer, as in the other tasks. All used tokenizers encode `front' and `back' as a single token such that the logit difference is well-defined and calculated the same way in all tasks. For evaluating the model's answer, we follow the convention of CLEVR~\cite{clevr} and WhatsUp~\cite{whatsup} and define spatial positions from the viewer’s perspective. All datasets have a balanced answer distribution.

\section{Head-wise activation patching}
\label{sec:supp_hw_ap}
Activation patching~\cite{heimersheim_nanda,zhang_best_practices} is a mechanistic interpretability tool for localizing model components (\eg, layers, attention heads) that are important for a model's behavior. It reveals which activations causally restore an answer. Unlike linear probing~\cite{linear_probes}, which only assesses whether information is decodable from hidden states without examining whether the encoded knowledge is causally relevant or visually grounded~\cite{probing_shortcomings, decodable_not_grounded}, activation patching localizes where the model processes information that influences the output. 

\begin{figure}[t]
  \centering
  \includegraphics[width=\linewidth]{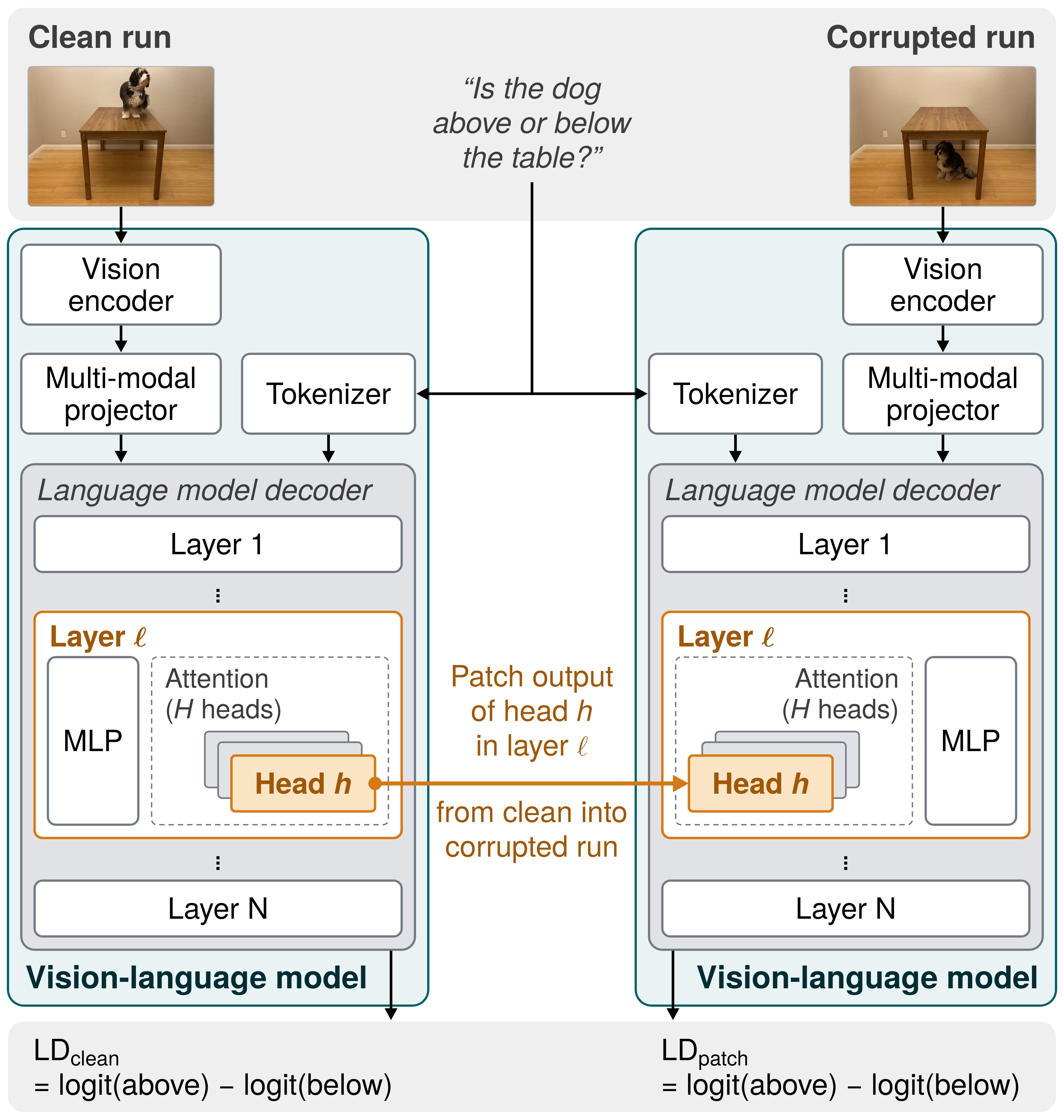}
  \caption{\textbf{Head-wise Activation Patching.} Activation patching inserts an attention head's activation from a clean run into a corrupted run and measures how much the patched activation restores the clean run's answer. }
  \label{fig:overview_ap}
\end{figure}

Activation patching requires generating a clean input  $x_{clean}$ and a corrupted input $x_{corr}$ that differ on a key fact, such that the model gives different answers $answer_{clean}$ and $answer_{corr}$, respectively. We use the logit difference
\begin{equation}
    LD = logit(answer_{clean}) - logit(answer_{corr})
\end{equation}
as a metric to quantify the model's preference between the two answers, as Zhang and Nanda~\cite{zhang_best_practices} recommend. The LD is always computed towards the clean answer.
The activation patching procedure includes the following steps~\cite{heimersheim_nanda}:
\begin{enumerate}
    \item \textbf{Activations:} choose the model activation to patch
    \item \textbf{Clean run:} run the model on the clean input, cache the component's activation, and calculate $LD_{clean}$
    \item \textbf{Corrupted run:} run the model on the corrupted input and calculate $LD_{corr}$
    \item \textbf{Patching:} run the model on the corrupted input while replacing the selected activation with its corresponding activation from the clean run, then compute $LD_{patch}$
    \item \textbf{Patching effect:} analyze how the patching intervention changes the model's answer by calculating the recovery 
\begin{equation}
    recovery = \frac{LD_{patch} - LD_{corr}}{LD_{clean} - LD_{corr}}
\end{equation}
    \item \textbf{Repeat:} repeat $1.-5.$ per component
\end{enumerate}
The patching effect shows how much patching a component's activation restores the behavior observed in the clean run. If the recovery score is positive, patching a component's activation restores the clean run's answer. Thus, this component causally influences the model's behavior.
If the recovery score is negative, patching the component leads the model to favor the corrupted run's answer even more. A component does not influence the model's answer for the given input when the recovery is zero.~\cref{fig:overview_ap} shows the activation patching procedure.

In our experiment, the clean and corrupted runs share the same question and differ only in the image. The objects' positions vary such that the correct answers are opposites in the clean and corrupted runs (\eg, $answer_{clean}=$ above, $answer_{corr}=$ below). Across pairs of clean and corrupted runs, we counterbalance two factors. First, we feed each image pair from our directional relations dataset to the model twice, such that each image serves once as the clean input and once as the corrupted input. Both answers occur equally often as clean answers, which prevents activation patching from highlighting heads whose activation induces a bias toward one answer. Second, we ask the same question with varying object order. For the example in~\cref{fig:overview_ap}, we ask ``Is the dog above or below the table?'' for one pair of runs and ``Is the table above or below the dog?'' for another pair. 
\begin{figure}[b]
  \centering
  \includegraphics[width=\linewidth]{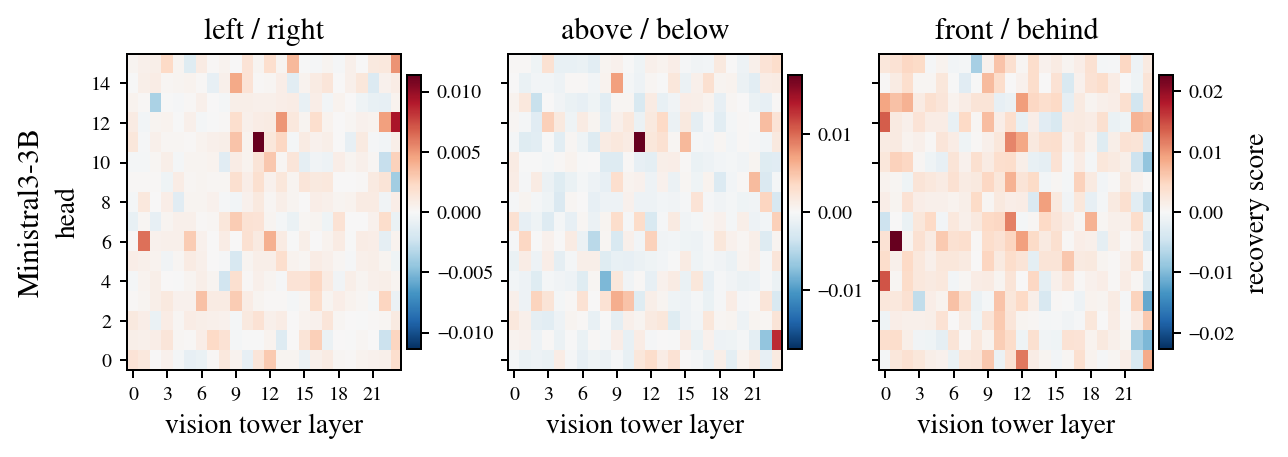}
  \includegraphics[width=\linewidth]{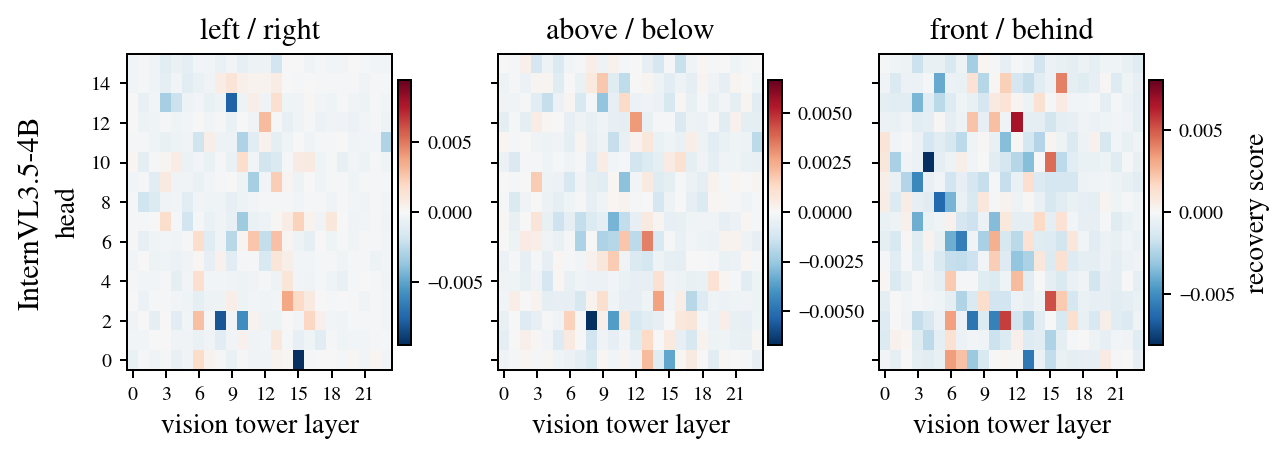}
  \includegraphics[width=\linewidth]{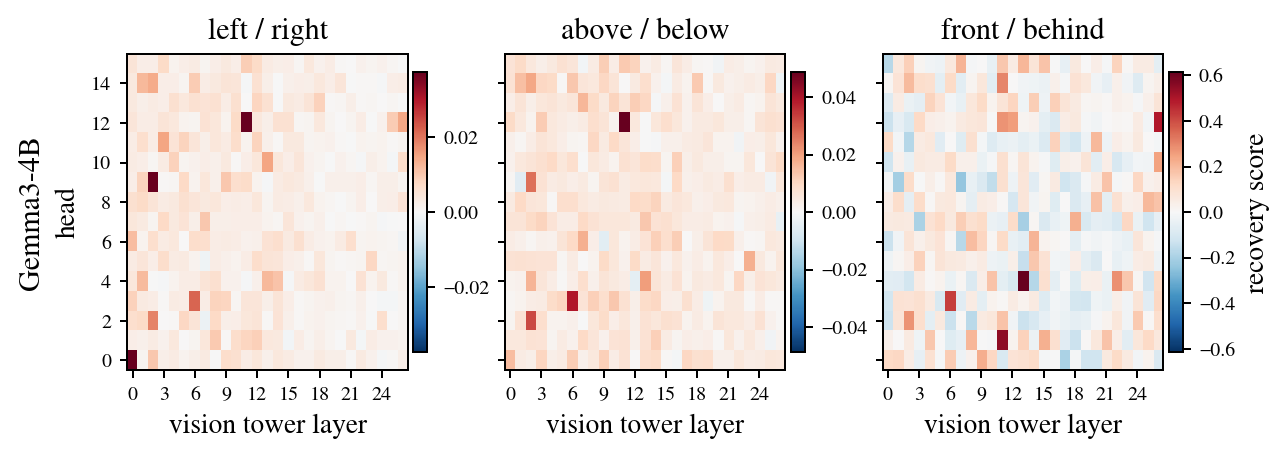}
  \caption{\textbf{Head-wise Activation Patching Results of Vision Encoder.} Performing activation patching on the vision encoder's attention heads shows low recovery scores. Patching does not influence the answer directly. The vision encoder provides information about directional relations to the LM decoder, but no head directly influences the answer.}
  \label{fig:ap_vision}
\end{figure}

\textbf{Activation patching results.}\hspace{0.2cm}We perform activation patching for every head in the vision encoder for all VLMs. We patch each head's output at all image token positions and compute the recovery score at the LM decoder's answer position. For each model and task, we average the head's recovery scores across datasets to get a recovery score for every head per spatial axis.~\Cref{fig:ap_vision} shows the heads’ recovery scores as one heatmap per model and spatial axis. The vision encoder heads have a weak direct impact on the answer because their recovery scores are extremely low. The heads in Ministral's and InternVL's vision encoder reach a recovery of at most $0.025$. The vision encoder heads in Gemma achieve a recovery of at most $0.05$ on the horizontal and vertical axes. Only the heatmap for Gemma on the depth axis shows higher values, but Gemma has an accuracy of $50\%$ on these datasets. Thus, we consider it a noisy result and neglect it in our choice. These results suggest that the vision encoder provides no valuable signal for the task. However, layer-wise activation patching rebuts this. Patching the input to the first LM decoder layer (i.e., the final output of the vision encoder) leads to a recovery of $0.98$ to $1.0$. Thus, the vision encoder passes information about directional relations to the LM decoder. Following prior work~\cite{spatial_IDs,reading_img,dual_mechanisms}, we hypothesize that the vision encoder spreads the spatial information across image tokens, while the LM decoder connects it to the textual information that influences the answer. As a result, activation patching identifies several influential LM decoder heads, whereas the vision encoder heads’ signal is weak and diffuse. Thus, we focus on the LM decoder to find influential attention heads to steer.

\section{Faithfulness results}
\label{sec:appendix_faithfulness_results}
This section extends the section in the main part, reporting additional faithfulness results. The main part includes the necessity and sufficiency curves for Ministral (see~\cref{fig:faithfulness_ministral}).
\begin{table}[b]
\centering
\caption{\textbf{Circuit Necessity.} Parameter $k_{10}$ is the smallest circuit whose ablation removes $\geq 90\%$ of the logit-difference effect, i.e., the smallest size at which necessity falls below $0.1$ and stays below at every larger size. The circuit of size $k$ includes $k$ heads with the highest recovery scores. $k_{10}$ is listed as the absolute number of heads and as a percentage of the total number of heads in the LM decoder. LR: left-right, AB: above-below, FB: front-behind.
}
\label{tab:circuit-k10}
\begin{tabular}{llcccccc}
\toprule
& & \multicolumn{2}{c}{Gemma} & \multicolumn{2}{c}{InternVL} & \multicolumn{2}{c}{Ministral}\\
\cmidrule(lr){3-4}\cmidrule(lr){5-6}\cmidrule(lr){7-8}
Dataset &  & $k_{10}$ & \% & $k_{10}$ & \% & $k_{10}$ & \%\\
\midrule
2D Dots & LR & 1 & 0.4 & 11 & 1.0 & 7 & 0.8\\
 & AB & 1 & 0.4 & 14 & 1.2 & 7 & 0.8\\
\addlinespace
2D Squares & LR & 1 & 0.4 & 10 & 0.9 & 7 & 0.8\\
 & AB & 2 & 0.7 & 11 & 1.0 & 8 & 1.0\\
\addlinespace
2D Triangles & LR & 1 & 0.4 & 11 & 1.0 & 7 & 0.8\\
 & AB & 1 & 0.4 & 12 & 1.0 & 9 & 1.1\\
\addlinespace
3D Objects & LR & 1 & 0.4 & 25 & 2.2 & 7 & 0.8\\
 & AB & 2 & 0.7 & 12 & 1.0 & 7 & 0.8\\
 & FB & 11 & 4.0 & 6 & 0.5 & 7 & 0.8\\
\addlinespace
WhatsUp & LR & 1 & 0.4 & 27 & 2.3 & 6 & 0.7\\
 & AB & 5 & 1.8 & 22 & 1.9 & 9 & 1.1\\
 & FB & 31 & 11.4 & 10 & 0.9 & 7 & 0.8\\
\midrule
\multicolumn{2}{l}{\emph{Median}} & 1 & 0.4 & 12 & 1.0 & 7 & 0.8\\
\bottomrule
\end{tabular}
\end{table}

\begin{figure*}[tb]
  \centering
  \includegraphics[width=0.9\linewidth]{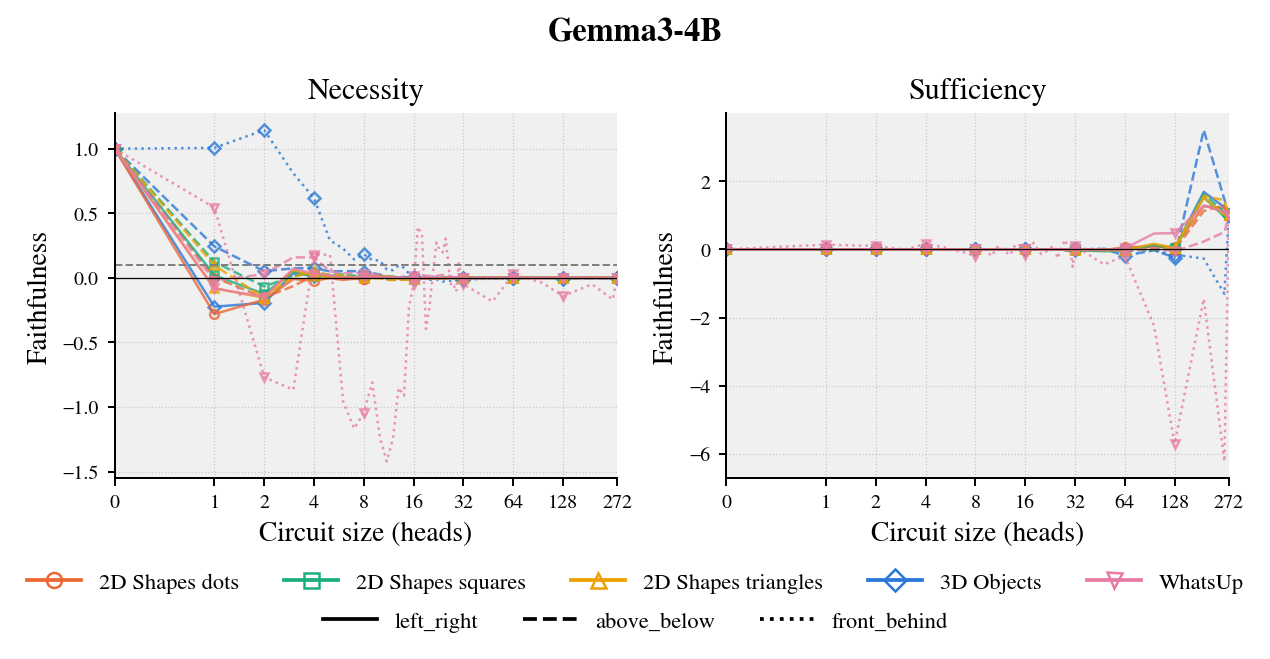}
  \caption{\textbf{Faithfulness Analysis of Gemma.} When ablating the top-$1$ head with the highest recovery, the model retains less than $10\%$ of its original performance on most datasets, i.e., the top-$1$ is necessary for the behavior (left panel). Sufficiency stays at zero for most circuit sizes (right panel). The circuit is only sufficient when it includes nearly all heads. Gemma has outlier curves, \eg 3D Objects-front/behind and WhatsUp-front/behind. On these datasets, the accuracy is at chance, and the model's behavior is noisy.}
  \label{fig:faithfulness_gemma}
\end{figure*}

\begin{figure*}[tb]
  \centering
  \includegraphics[width=0.9\linewidth]{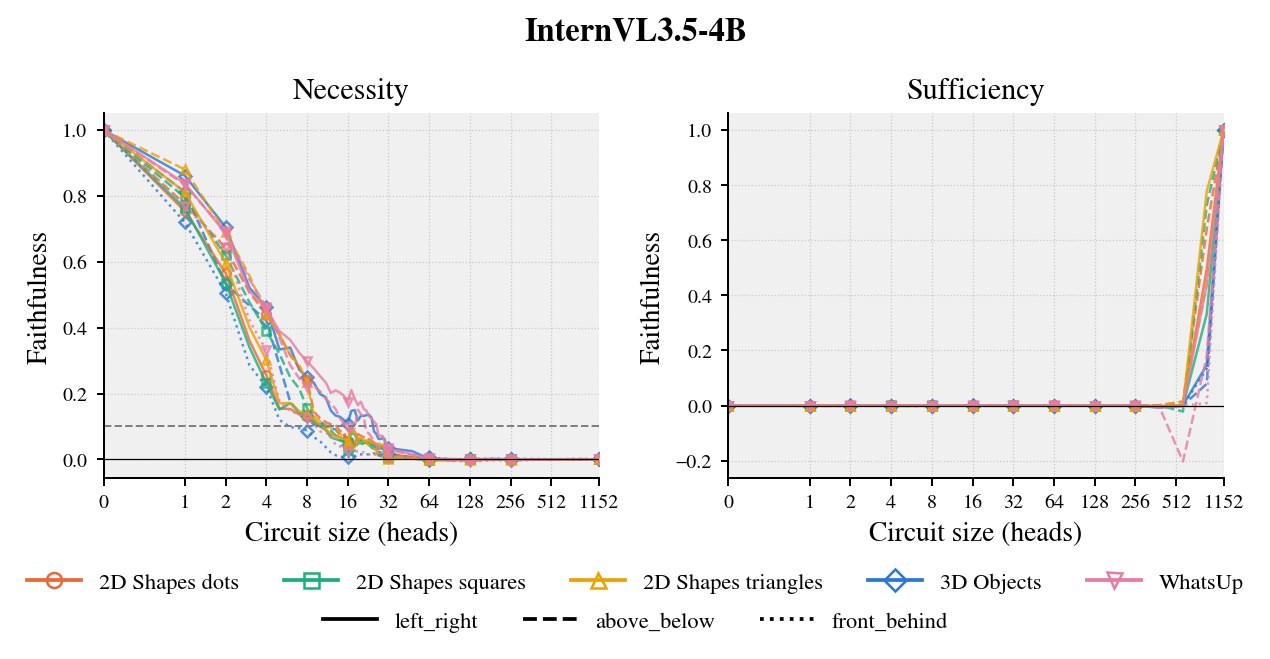}
  \caption{\textbf{Faithfulness Analysis of InternVL.} When ablating the top-$6$ to top-$27$ heads ranked by recovery, the model retains less than $10\%$ of its original performance on most datasets, i.e., these heads are necessary for the behavior (left panel). Ablating all heads, except the top-$6$ to top-$27$, is not sufficient to restore the behavior (right panel). Sufficiency increases to $1$ when the circuit includes nearly all heads. }
  \label{fig:faithfulness_internvl}
\end{figure*}

\textbf{Necessity.}\hspace{0.2cm}~\Cref{fig:faithfulness_gemma} (left panel) visualizes Gemma's results for the necessity analysis (one curve per dataset). The x-axis and y-axis show the circuit size and faithfulness, respectively. Gemma has two outlier curves, namely 3D Objects-front/behind and WhatsUp-front/behind, whose values differ considerably from the other datasets. On these datasets, Gemma achieves chance-level accuracy, which yields a small denominator in the faithfulness score (see~\cref{eq:faithfulness}). As a result, the small denominator amplifies small fluctuations that we observe in the noisy outlier curves. Excluding these two noisy runs, the results are similar across datasets. When we ablate the head with the highest recovery score, the faithfulness already falls below $0.1$ in most cases. For some datasets (\eg, 3D Objects-left/right or 2D Dots-left/right), the faithfulness is negative for circuit sizes $k=1$ and $k=2$ but then increases to zero at circuit size $k=4$. This progression means that ablating the top-$1$ or top-$2$ heads inverts the model's original behavior on the directional relation tasks. Ablating further heads restores parts of the original behavior. Hence, the head with the highest recovery score drives the overall behavior, while heads with lower recovery scores oppose the first head's signal. This finding agrees with the activation patching results, where the top-$1$ achieves a recovery score of $1$ (i.e., it perfectly restores the clean run's answer). If we ablate the top-$1$ head, the opposing heads are unopposed, resulting in negative faithfulness. Ablating up to four heads lets the curves converge to zero.  The values stay around zero when we ablate further heads (i.e., nothing of the original directional relation prediction behavior is left). Unlike Gemma, InternVL has no outliers, as shown in~\cref{fig:faithfulness_internvl} (left panel). The curves fall steeply and converge to zero already for small circuit sizes. The model retains less than $10\%$ of its original performance when we ablate at most $27$ heads (i.e., $2.3\%$ of all heads in the LM decoder). 

Although trends are similar across datasets, faithfulness values vary slightly across datasets for the same circuit sizes. For steering, we aim to find a criterion for selecting effective steering targets that generalizes across datasets and models. Therefore, we define $k_{10}$ as the smallest circuit whose ablation removes $\geq 90\%$ of the model's behavior (i.e., faithfulness $\leq 10\%$) at this size and at every larger size. We do not set the threshold to exactly zero because many heads have small diffuse effects and the faithfulness is noisy near zero. Setting the threshold to $0.1$ instead of $0.0$ has the advantage of yielding a smaller set of steering targets and excluding heads that add small noisy signals. As $k_{10}$ varies slightly across datasets, we choose the steering targets using the median $k_{10}$, i.e., the top-$k$ heads with $k = \operatorname{median}_d\, k_{10}^{(d)}$. This decision has two reasons. Firstly,  the median is robust to outliers. Outlier curves where the model's behavior is noisy, as seen for Gemma, can increase $k_{10}$ considerably. The median $k_{10}$ prevents noisy outlier curves from determining the steering targets. Secondly, it provides a uniform criterion that applies identically across datasets and models. We avoid per-dataset tuning and keep the resulting steering targets comparable.~\Cref{tab:circuit-k10} reports $k_{10}$ as the number of heads as well as a share of all attention heads in the LM decoder to facilitate comparability across models. The median $k_{10}$ is $1$, $7$, and $12$ heads for Gemma, Ministral, and InternVL, respectively.

\textbf{Sufficiency.}\hspace{0.2cm}~\cref{fig:faithfulness_gemma} and~\cref{fig:faithfulness_internvl} (right panels) show Gemma's and InternVL's sufficiency analysis results. Gemma has more outlier curves, and faithfulness curves vary more strongly across datasets, compared to InternVL and Ministral. Nevertheless, all models show a common trend: sufficiency stays zero for most circuit sizes and increases only when the circuit includes nearly all heads. Thus, our sufficiency analysis does not reveal a small stand-alone circuit that is sufficient to restore the model's behavior in directional relation prediction. As described in the main part, we focus on necessity results for selecting a small set of attention heads to steer due to these outcomes.

\clearpage

\clearpage
\section{Model steering}
\label{sec:appendix_steering}
This section provides additional details on model steering.

\subsection{Steering targets}
\label{sec:appendix_steered_heads}
As described in~\cref{sec:experiment_steering}, we steer the heads in the $k_{10}$ set. Specifically, these are the following heads.

\noindent
\textbf{Ministral (top-$7$ heads)}
\begin{itemize}
    \item Horizontal axis: L11H27, L13H11, L13H17, L15H30, L16H19, L17H31, L19H4
    \item Vertical axis: L11H27, L13H17, L15H30, L16H19, L17H31, L19H4, L19H6
    \item Depth axis: L13H11, L13H17, L15H30, L16H19, L17H31, L19H4, L21H3  
\end{itemize}

\noindent
\textbf{Gemma (top-$1$ head)}
\begin{itemize}
    \item Horizontal axis: L17H0
    \item Vertical axis: L17H0
    \item Depth axis: L23H3
\end{itemize}

\noindent
\textbf{InternVL (top-$12$ heads)}
\begin{itemize}
    \item Horizontal axis: L15H9, L15H11, L17H4, L18H11, L18H15, L23H25, L24H13, L26H24, L27H8,  L28H0, L30H5, L35H26
    \item Vertical axis:  L15H9, L15H11, L17H4, L18H11, L18H15, L23H25, L24H13, L26H24, L26H25, L27H8, L30H5, L34H17
    \item Depth axis: L15H9, L15H11, L17H4, L17H15, L18H11, L18H15, L20H0, L23H25, L26H25, L26H26, L27H8, L30H5
\end{itemize}

\subsection{Steering strength selection}
\begin{figure}[tb]
  \centering
  \includegraphics[width=\linewidth]{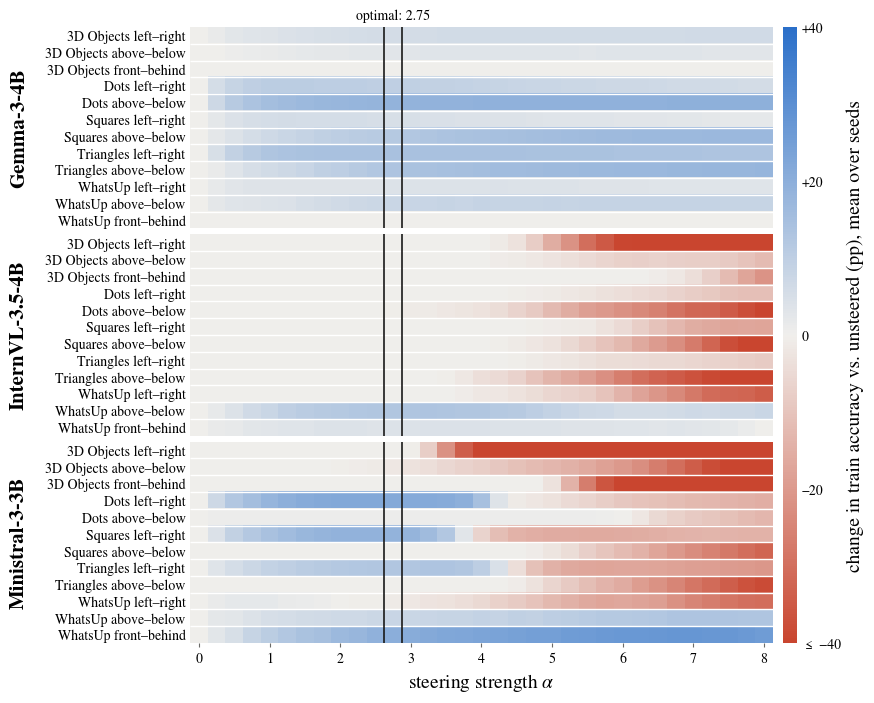}
  \caption{\textbf{Hyperparameter Search for Projective Steering Strength.} We choose the optimum of $\alpha=2.75$ for the projective steering experiments, as this steering strength achieves the highest accuracy gain averaged over all training sets.}
  \label{fig:alpha_heatmap}
\end{figure}
\textbf{Setup.}\hspace{0.2cm}This section summarizes our hyperparameter search to find an optimal steering strength for projective and shift steering~\cite{attention_in_space, ITI}. We perform the search on all VLMs and datasets. The steering runs intervene on the $k_{10}$ heads, i.e., the top-$1$, top-$12$, and top-$7$ heads for Gemma, InternVL, and Ministral, respectively. Projective steering intervenes at all token positions with individual means and answer directions per token group (image or text). We split the directional relation datasets pairwise into a training and test set (train: $50\%$, test: $50\%$). For projective steering, we compute the mean $\mu$ and answer direction $v$ only with the training samples. Additionally, we use the training samples to compute the shift steering parameters (i.e., the activation direction and standard deviation of activations). For both steering methods, we choose the optimal $\alpha$ based only on the models' performance on the training sets. Finally, we evaluate the steering effect on the test split. We sweep over 33 steering strengths: $\alpha=0.0$ to $\alpha=8.0$ in $0.25$-steps for projective steering. For shift steering, we sweep over the same values and the additional value $\alpha=0.1$ used by Ma \etal~\cite{attention_in_space}. 
We repeat each run five times with five different random seeds. For each steering strength, we average the train accuracy over the five seeds for each model and dataset, and report its change against $\alpha=0.0$ (i.e., unsteered baseline) in percentage points. Based on the results, we select the steering strength $\alpha$ that leads to the highest average gain without harming any model by more than $2.0$ percentage points. We aim to find a steering strength that generalizes well across models and datasets instead of picking an individual strength per condition.

\textbf{Projective steering strength.}\hspace{0.2cm}\Cref{fig:alpha_heatmap} shows the change in training accuracy compared to the unsteered baseline per steering strength for all models and datasets. For small steering strengths, projective steering has a weak effect. At $\alpha= 2.75$, the average gain on the training set is $5.2$\,pp (mean gain: $7.7$\,pp for Gemma, $1.3$\,pp for InternVL, $6.6$\,pp for Ministral). We choose $\alpha= 2.75$ as it leads to the highest average gain and generalizes well across models and datasets.
From $\alpha=3.0$ to $8.0$, the accuracy gains start to decrease. Thus, oversteering with high steering strengths harms accuracy. Only Gemma does not perform worse for steering strengths above $3.0$. The reason, therefore, could be that it achieves the lowest unsteered accuracy among the three VLMs. So, it has more room to improve compared to InternVL that achieves high accuracy without steering and an additional intervention influences the accuracy negatively.

\begin{figure}[tb]
  \centering
  \includegraphics[width=\linewidth]{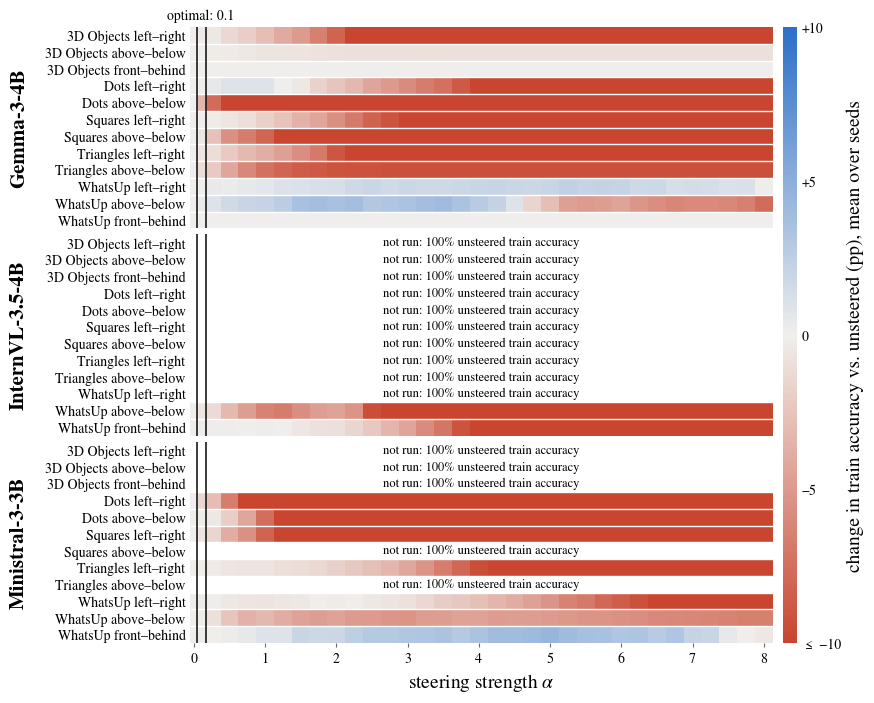}
  \caption{\textbf{Hyperparameter Search for Shift Steering Strength.} There is no one-fits-all optimum for the shift steering strength $\alpha$. So, we choose $\alpha=0.1$~\cite{attention_in_space}.}
  \label{fig:shift_alpha_heatmap}
\end{figure}

\textbf{Shift steering strength.}\hspace{0.2cm}\Cref{fig:shift_alpha_heatmap} shows the change in training accuracy compared to the unsteered baseline per steering strength for all models and datasets when we apply shift steering. The shift steering rule is undefined when no training sample is answered incorrectly. So, there are no values for datasets with $100\%$ training accuracy (labeled with ``not run: 100\% unsteered train accuracy'' in heatmap). The hyperparameter sweep for the shift steering strength does not show a one-fits-all optimum. For most model-dataset combinations, all choices of $\alpha$ decrease the training accuracy (marked in red). We choose $\alpha=0.1$ because any higher value of $\alpha$ harms performance considerably, and it is the value chosen by Ma \etal~\cite{attention_in_space}. For Gemma, the heatmap shows some blue regions on WhatsUp-left/right and WhatsUp-above/below. Ministral shows an accuracy gain on WhatsUp-front/behind for several $\alpha$ values. Hence, we evaluate the gain on the test sets when we choose an individual $\alpha$ for these models and datasets. For Gemma, the best $\alpha$ on WhatsUp-left/right and WhatsUp-above/below is $\alpha=5.3$ and $\alpha=3.5$, respectively. Then, the accuracy gains on the test sets are $2.0$\,pp and $3.1$\,pp for WhatsUp-left/right and WhatsUp-above/below, respectively. For Ministral on WhatsUp-front/behind, the best $\alpha$ of $5.0$ leads to an accuracy gain of $2.5$\,pp on the test split. Even with these individual $\alpha$ values, shift steering performs worse than projective steering on most test sets.

\subsection{General visual question answering datasets}
\label{sec:appendix_VQA_datasets}
Besides our directional relation datasets, we use diverse datasets to evaluate the steering effect on general VQA questions. This section describes the dataset preparation. We use three public datasets, namely GQA~\cite{GQA}, VQAv2~\cite{VQA}, and POPE~\cite{POPE}. From these datasets, we exclude all questions about spatial relations, so that they evaluate only effects on general capabilities and do not confound with the directional relation tasks that projective steering targets. Open-ended questions introduce language ambiguities. Hence, we focus on yes/no questions with well-defined answers such that we can score the answer as the model's preference between the two answers. Similar to the directional relation tasks, we apply an answer template: ``Look at the image. \{question\} Answer with one word: `yes' or `no'.'', where \{question\} is the placeholder for the question from the original dataset specified for the respective image. Additionally, we balance the answer distribution.

\textbf{GQA Dataset.}\hspace{0.2cm}We use the full testdev-balanced split of GQA v1.2, which consists of 12,578 questions on 398 images. After applying our filter criteria, we get 2728 questions (1364 yes, 1364 no).

\textbf{VQAv2 Dataset.}\hspace{0.2cm}We take the COCO val2014 images and questions. From this data, we keep yes/no-questions, where 8 of 10 annotators agree on the answer. As a result, we get 1948 question-image pairs (974 yes, 974 no). 

\textbf{POPE Dataset.}\hspace{0.2cm}This dataset includes three subsets, namely random, popular, and adversarial. These subsets differ in how the absent objects (objects named in the question that are not in the image) are chosen: random (random choice), popular (frequent objects in the dataset), and adversarial (objects that often co-occur with the present objects). Each subset contains 3000 questions (1500 yes, 1500 no) about 500 COCO val2014 images. The filtering excludes no image-question pairs because POPE includes only general yes/no questions by default.

\subsection{Steering results}
\label{sec:appendix_steering_results}
While we summarize the results in the main part, the supplementary material includes detailed results (see~\cref{tab:acc-gemma} to~\cref{tab:f1-pope-ministral}). For each model and spatial axis, we steer the median $k_{10}$ heads, which are only $1$, $7$, and $12$ heads for Gemma, Ministral, and InternVL, respectively.~\Cref{sec:appendix_steered_heads} lists the steered heads.
We steer the selected heads using~\cref{eq:steering} and compare our projective steering with shift steering~\cite{attention_in_space,ITI}, \textsc{ScalingVis}~\cite{adaptvis}, and \textsc{AdaptVis}~\cite{adaptvis}. For projective steering, we compute the steering parameters $\mu$ and $v$ on a directional-relation training set. Additionally, we use the training samples to compute the shift steering parameters (i.e., the activation direction and standard deviation of activations). To obtain training samples, we randomly split each directional relation dataset into equally sized halves. As the datasets consist of image pairs, we split the data so that both members of a pair are in the training set or the test set to avoid data leakage. Based on the hyperparameter sweep, we choose $\alpha_{proj}=2.75$ for projective steering and $\alpha_{shift}=0.1$ for shift steering. We evaluate the steering effect only on the test set. For both the directional relation and general VQA tasks, we specify prompts that ask the VLM to answer with one word, choosing between two possible answers (see prompt templates in~\cref{sec:appendix_data} and~\cref{sec:appendix_VQA_datasets}). Thus, each question queries the model to predict one of two answers, and we evaluate the model's preference between answers. We repeat each experiment five times (with five random seeds) to compute the mean and standard deviation (std) of accuracy and F1 score. As the two answer options for the directional relation datasets (\eg, left and right) have no positive class, we compute the macro-F1 score (i.e., the mean of the two answers' F1 scores). For general VQA datasets, we compute the standard F1 score with ``yes'' as the positive class, following the convention of POPE~\cite{POPE}. 

\Cref{tab:acc-gemma} to~\Cref{tab:acc-ministral} show the accuracy of Gemma, InternVL, and Ministral on directional-relation datasets. Without steering, the models have different weaknesses. Gemma performs better on the horizontal axis (left/right: $70.5\%$ to $88.3\%$) than on the vertical (above/below: $51.0\%$ to $78.5\%$) or depth (front/behind: $\approx 50\%$) axes. InternVL achieves an unsteered accuracy of $100\%$ on all datasets, except two sets (WhatsUp-AB: $62.7\%$, WhatsUp-FB: $95.8\%$). Ministral performs weakly on 2D shapes left/right ($67.3\%$ to $76.7\%$) and on WhatsUp above/below and front/behind ($55.2\%$ and $67.4\%$). Projective steering improves performance on most datasets, where accuracy is below ceiling ($2.8$ to $18.8$\,pp for Gemma, $4.1$ to $13.4$\,pp for InternVL, and $0.7$ to $21.3$\,pp for Ministral). However, when the model's accuracy is at chance level (\eg Gemma on 3D Objects-front/behind or WhatsUp-front/behind), projective steering achieves no accuracy gains. We hypothesize that the steered heads have no clear tendency towards an answer in these cases. Thus, projective steering does not shift their output considerably. Projective steering outperforms the other approaches in most cases.

\Cref{tab:acc-vqa-gemma} to~\Cref{tab:acc-pope-ministral} present the models’ accuracy on general VQA datasets. We compute the steering parameters on the directional relations training splits (given in the columns), but then evaluate the methods' (given in rows) effect on the general VQA datasets instead of the directional relation test split. \textsc{ScalingVis} and \textsc{AdaptVis} lead to the largest decrease in accuracy, especially for Gemma ($-17.2$\,pp and $-16.8$\,pp, both on VQAv2 with the 3D Objects-above/below training set). For InternVL and Ministral, the accuracy changes less drastically (from $-2.7$\,pp to $0.7$\,pp) when applying \textsc{ScalingVis} and \textsc{AdaptVis}. Projective steering harms Gemma's accuracy less than the previous two methods. It changes the accuracy by at most $-1.2$\,pp on Gemma (POPE popular), where it also gains up to $2.3$\,pp on GQA, by $-3.3$\,pp on InternVL (on GQA), and by $-1.5$\,pp on Ministral (on VQAv2). Shift steering decreases the accuracy on general VQA tasks by at most $0.2$\,pp on any model, in line with its lack of effect on the directional relation tasks.

\Cref{tab:f1-gemma} to~\Cref{tab:f1-ministral} show the macro-F1 score of Gemma, InternVL, and Ministral on the directional relations dataset. The F1 scores reveal that Gemma has a bias towards one answer, as its F1 scores are $33.8\%$ to $35.4\%$ on 3D Objects-above/below, 3D Objects-front/behind, and WhatsUp-front/behind. Additionally, Gemma's F1 scores are around chance level on 2D squares-above/below, 2D triangles-above/below, or WhatsUp-above/below. InternVL shows deficits on WhatsUp-above/below and WhatsUp-front/behind. Ministral has low F1 scores for the horizontal axis on 2D Shapes ($63.3\%$ to $75.3\%$) and on WhatsUp-above/below and front/behind ($54.3\%$, $63.4\%$). Shift steering improves the F1 scores by at most $0.8$\,pp. \textsc{ScalingVis} and \textsc{AdaptVis} outperform shift steering on most model and dataset combinations. Moreover, \textsc{ScalingVis} and \textsc{AdaptVis} surpass projective steering on Gemma with 2D triangles-above/below, 3D Objects-above/below or front/behind, and WhatsUp-front/behind. Nevertheless, projective steering achieves the highest average gains. It improves the F1 scores of Gemma by $4.2$\,pp to $19.9$\,pp, InternVL by $4.1$\,pp and $12.6$\,pp, and Ministral by $0.7$\,pp to $25.2$\,pp.

\Cref{tab:f1-vqa-gemma} to~\Cref{tab:f1-pope-ministral} present the F1 score of the models on the general VQA datasets. Shift steering changes the F1 scores by at most $-0.5$\,pp on all datasets and models. \textsc{ScalingVis} and \textsc{AdaptVis} harm Gemma drastically, but do not decrease InternVL's F1 scores considerably. Lastly, these methods harm Ministral's performance by at most $-4.1$\,pp (on GQA). Lastly, projective steering harms the F1 scores by at most $-4.4$\,pp for InternVL, $-4.0$\,pp for Ministral, and $-1.2$\,pp for Gemma.

\section{Ablation studies}
This section describes the setup and results of the ablation studies.
\subsection{Faithfulness results for random head rankings}
\label{sec:appendix_faith_ablation}
\begin{table}[b]
\centering
\caption{\textbf{Ablation - $k_{10}$ for Random Head Rankings.} Parameter $k_{10}$ is the smallest circuit whose mean-ablation removes $\ge$90\% of the logit-difference effect, i.e., the smallest size at which necessity falls below $0.1$ and stays below at every larger size sampled. The circuit includes random, not recovery-ranked, heads. The table shows results from three different runs with three different seeds, $k_{10}$ as the mean [min--max]. LR: left-right, AB: above-below, FB: front-behind.
}
\label{tab:control-k10}
\footnotesize
\setlength{\tabcolsep}{3pt}
\resizebox{\columnwidth}{!}{%
\begin{tabular}{@{}ll ccc@{}}
\toprule
& & \multicolumn{3}{c}{Random-ranking $k_{10}$ (mean [min--max])}  \\
\cmidrule(l){3-5}
Dataset & Relation & Gemma & InternVL & Ministral \\
\midrule
2D Dots & LR & 96 [64--128] & 576 [576--576] & 235 [128--384] \\
        & AB & 91 [48--128] & 576 [576--576] & 203 [96--256] \\
\addlinespace
2D Squares & LR & 85 [64--96] & 512 [384--576] & 235 [128--384]\\
           & AB & 91 [48--128] & 576 [576--576] & 203 [96--256] \\
\addlinespace
2D Triangles & LR & 75 [64--96] & 576 [576--576] & 235 [128--384] \\
             & AB & 101 [48--128] & 576 [576--576] & 203 [96--256] \\
\addlinespace
3D Objects & LR & 96 [64--128] & 512 [384--576] & 235 [128--384] \\
           & AB & 96 [64--128] & 576 [576--576] & 203 [96--256] \\
           & FB & 53 [48--64] & 512 [384--576] & 192 [128--256] \\
\addlinespace
WhatsUp & LR & 85 [64--96] & 576 [576--576] & 235 [128--384] \\
        & AB & 91 [48--128] & 512 [384--576] & 149 [96--256] \\
        & FB & 43 [1--128] & 512 [384--576] & 213 [128--384] \\
\midrule
\multicolumn{2}{l}{\emph{Median}} & 91 & 576 & 208 \\
\bottomrule
\end{tabular}
}
\end{table}

As a control study, we run the faithfulness experiment with three random head rankings per model (using three different seeds) instead of the recovery ranking. The results show whether the results in~\cref{subsec:faithfulness_experiment} depend on which heads are included or on how many heads the circuit includes. 

\begin{figure}
  \centering
 \includegraphics[width=\linewidth]{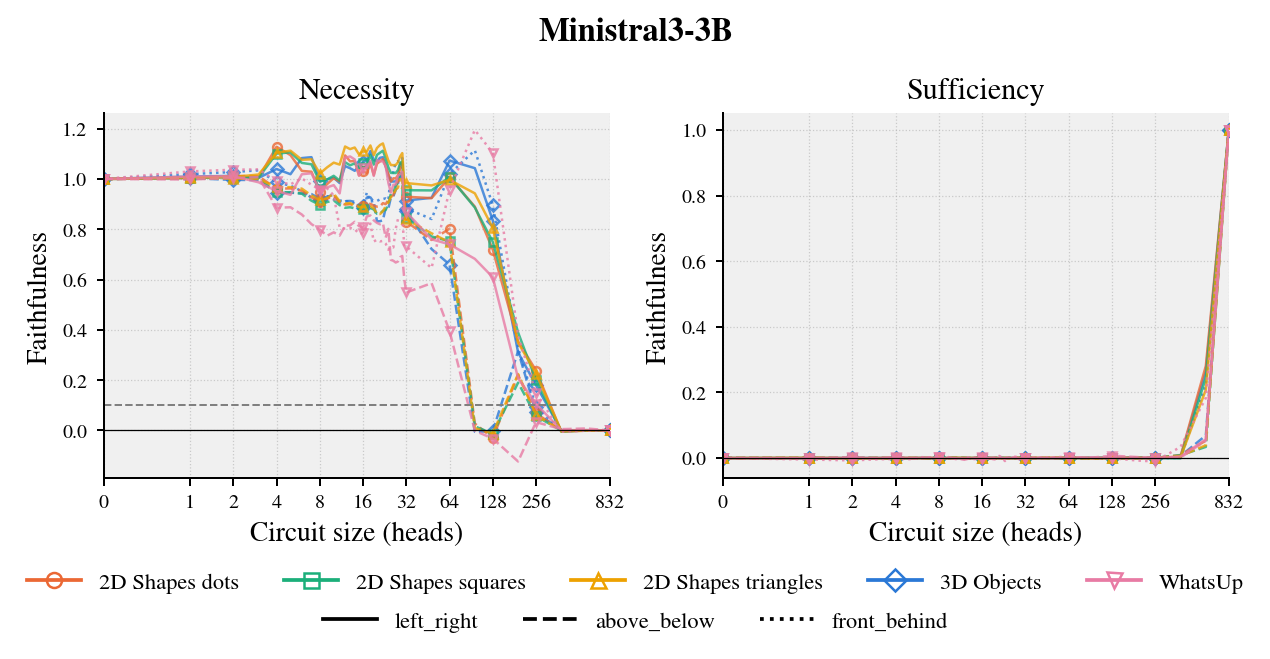}
 \includegraphics[width=\linewidth]{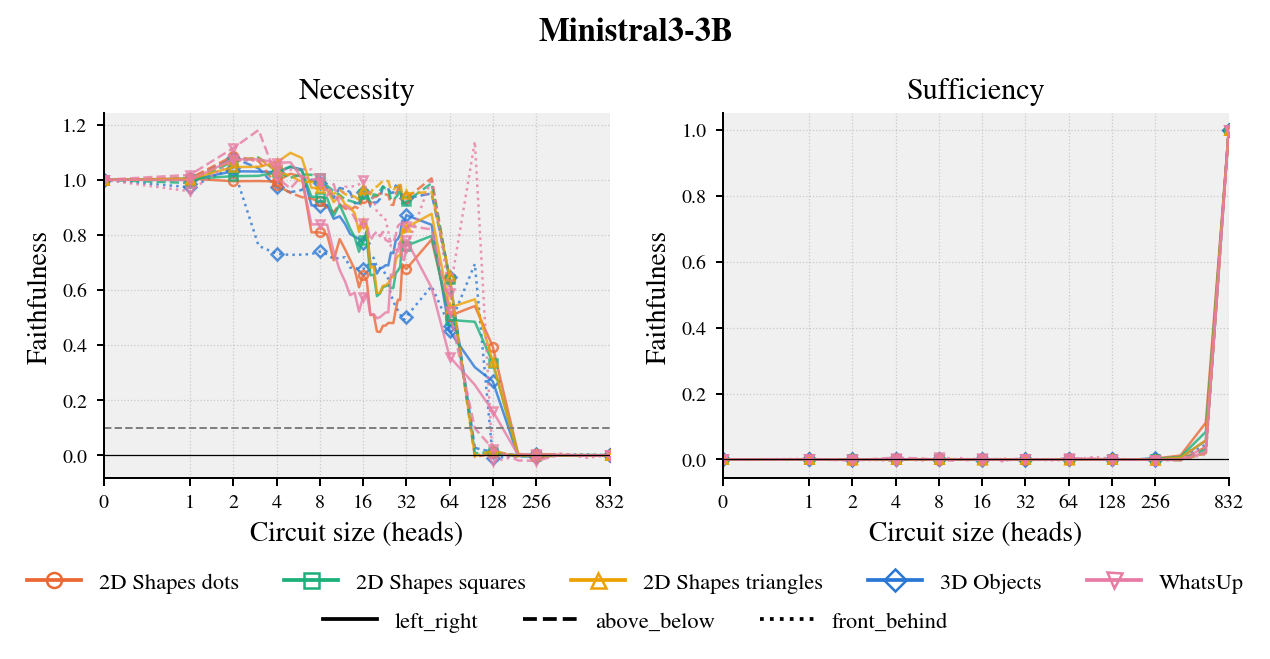}
 \includegraphics[width=\linewidth]{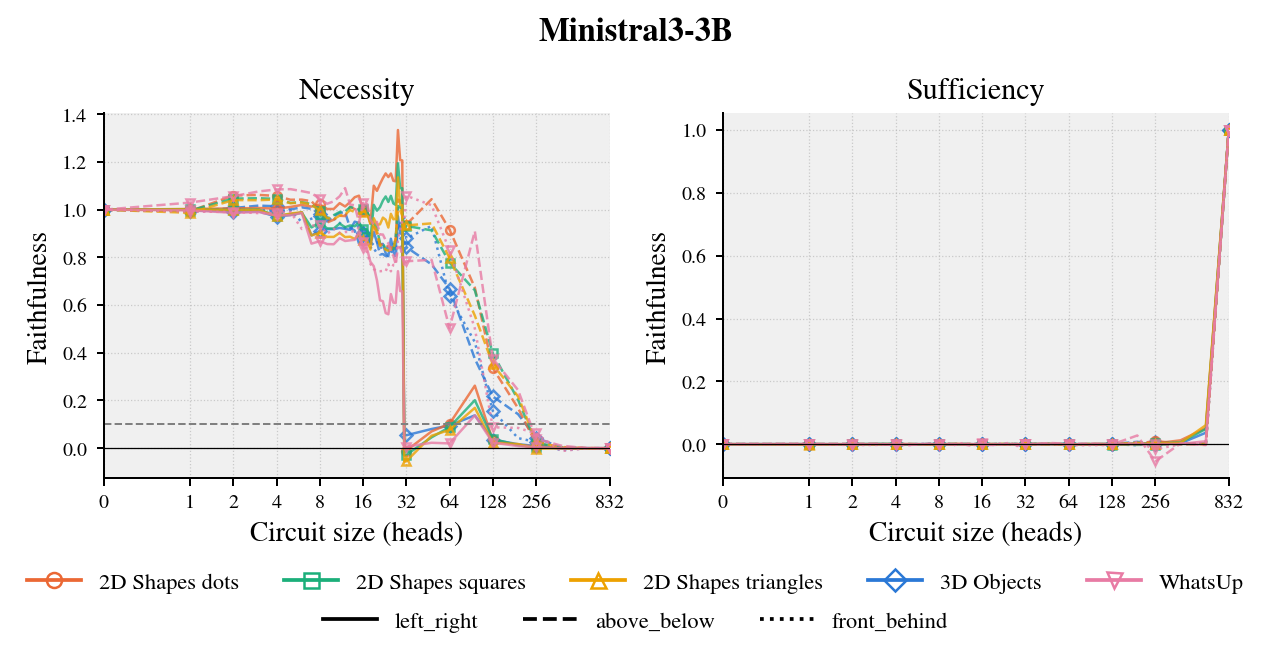}
 \caption{\textbf{Ablation - Faithfulness Analysis of Ministral with Random Head Rankings.} We repeat the necessity (left) and sufficiency (right) analysis using three random head rankings. Compared to the recovery ranking, the necessity curves are noisier. Additionally, more heads must be ablated before faithfulness drops, which means that small circuits consisting of heads with high recovery scores describe the model's behavior more accurately.}
  \label{fig:faithfulness_ministral_control}
\end{figure}

\begin{figure}
  \centering
  \includegraphics[width=\linewidth]{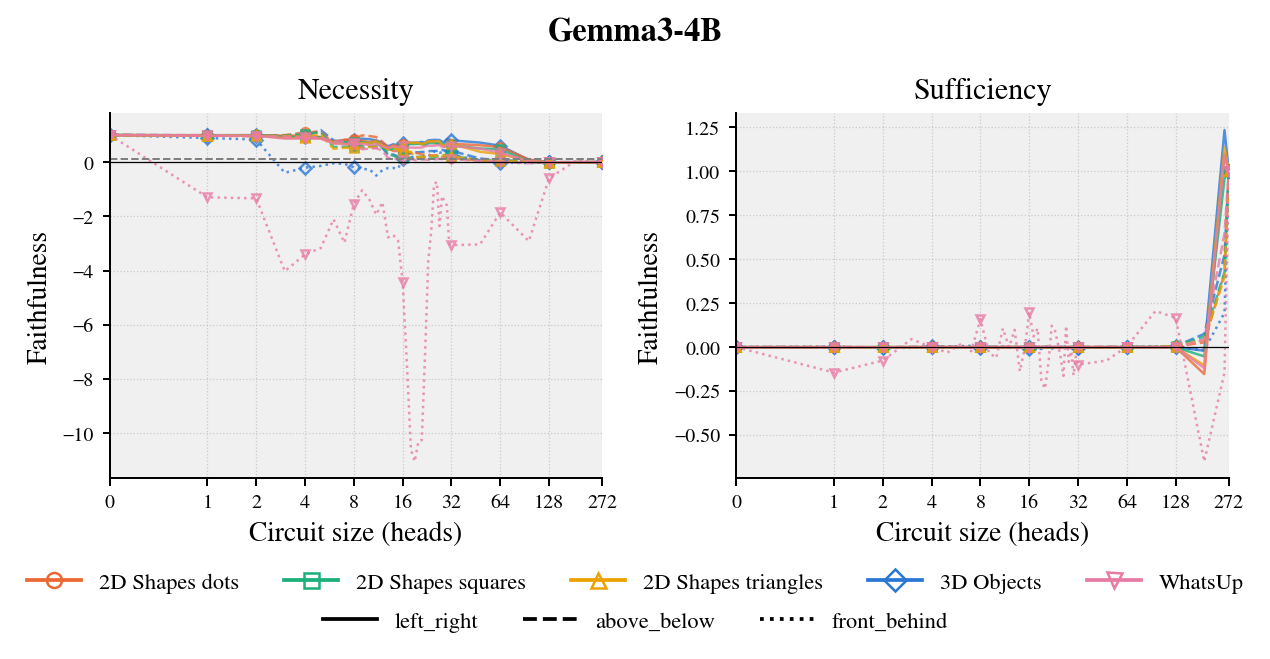}
  \includegraphics[width=\linewidth]{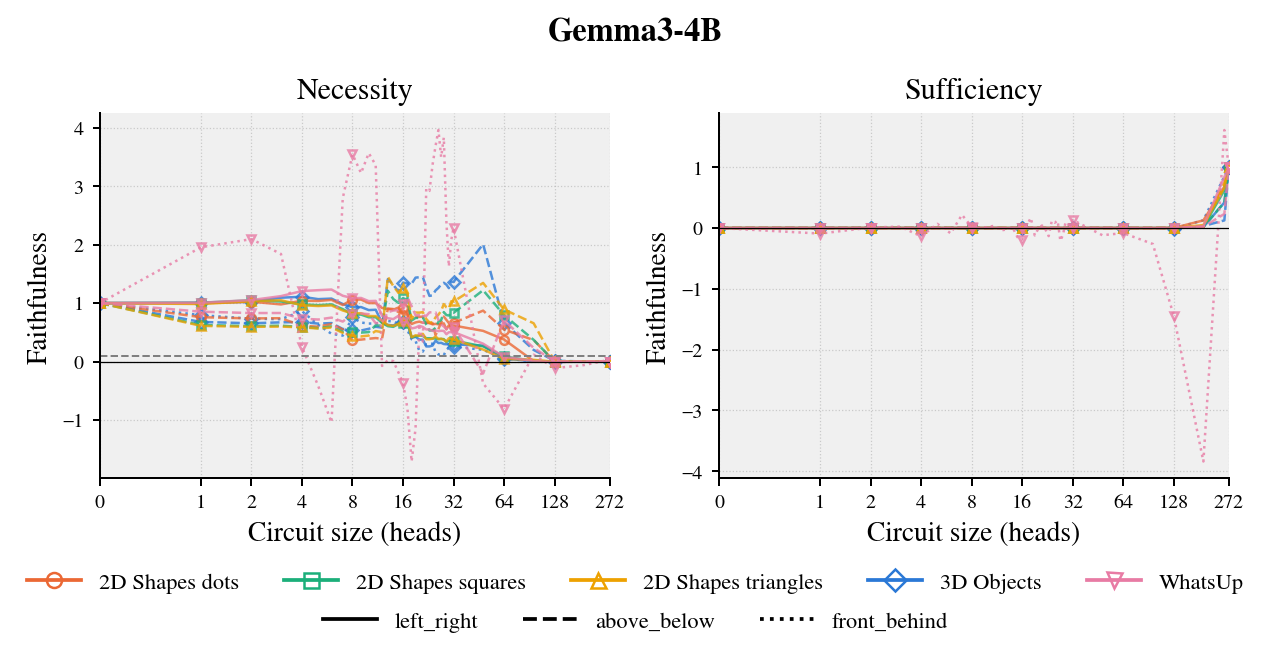}
  \includegraphics[width=\linewidth]{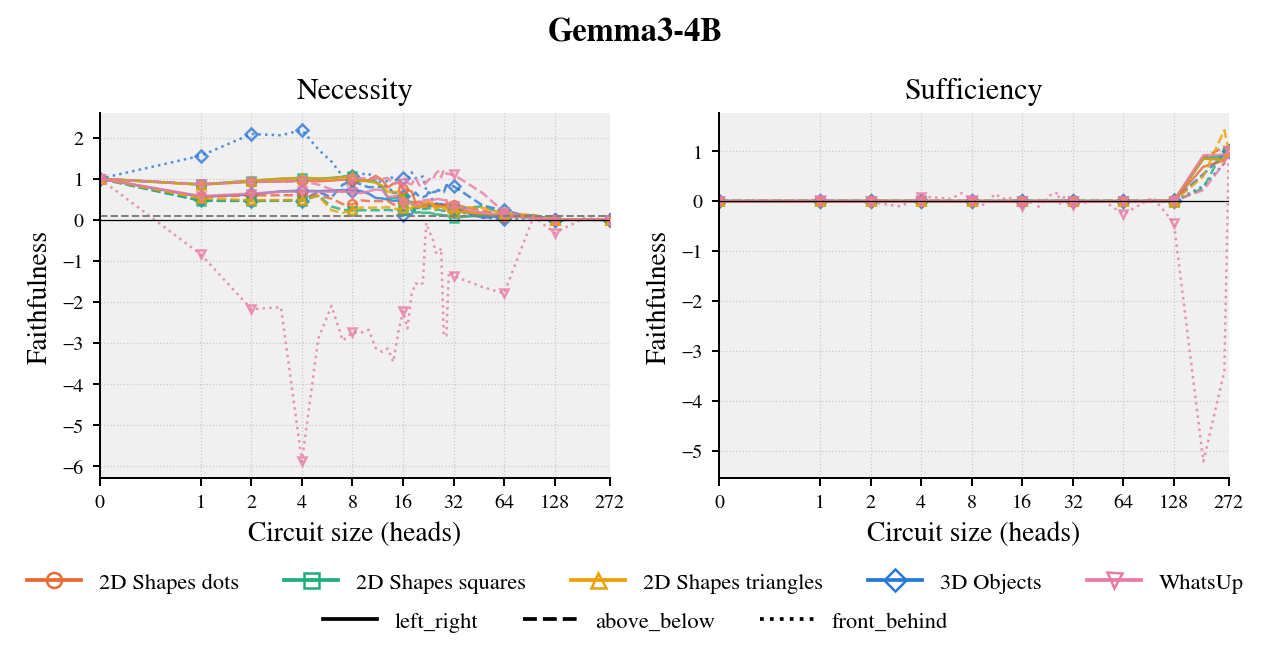}
  \caption{\textbf{Ablation - Faithfulness Analysis of Gemma with Random Head Rankings.} We repeat the necessity (left) and sufficiency (right) analysis using three random head rankings. Compared to the recovery ranking, the necessity curves are noisier. Additionally, more heads must be ablated before faithfulness drops, which means that small circuits consisting of heads with high recovery scores describe the model's behavior more accurately than these circuits of random heads.}
  \label{fig:faithfulness_gemma_control}
\end{figure}

\begin{figure}[h]
  \centering
    \includegraphics[width=\linewidth]{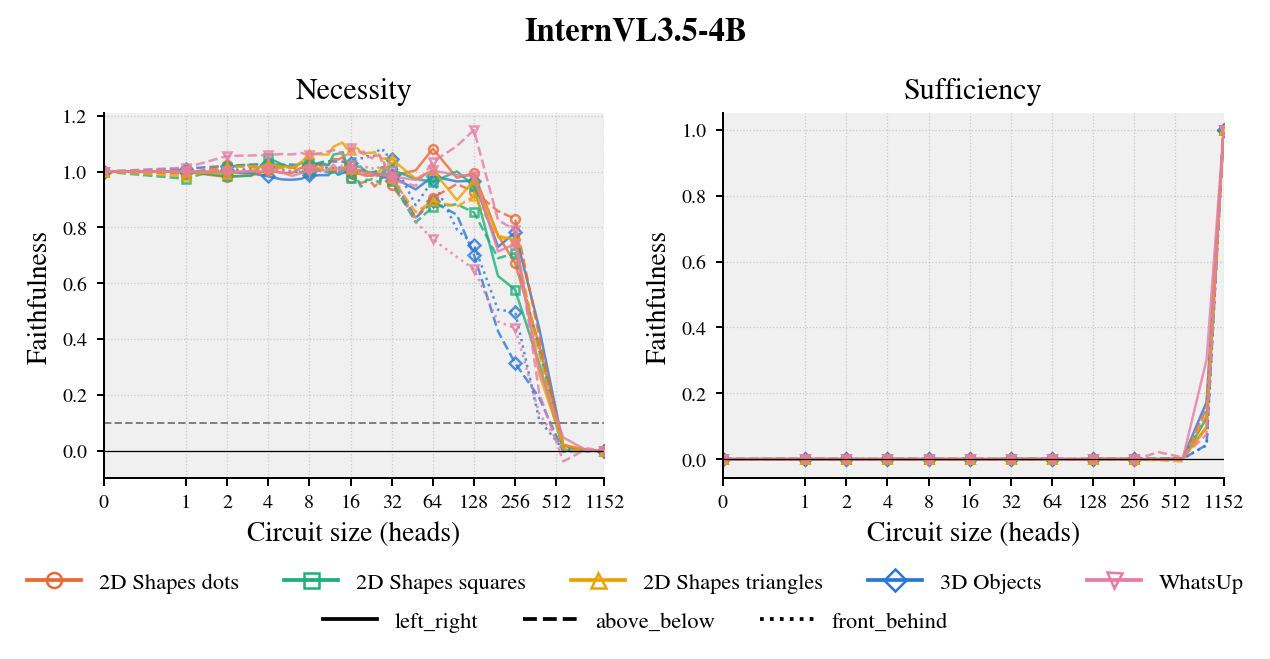}
    \includegraphics[width=\linewidth]{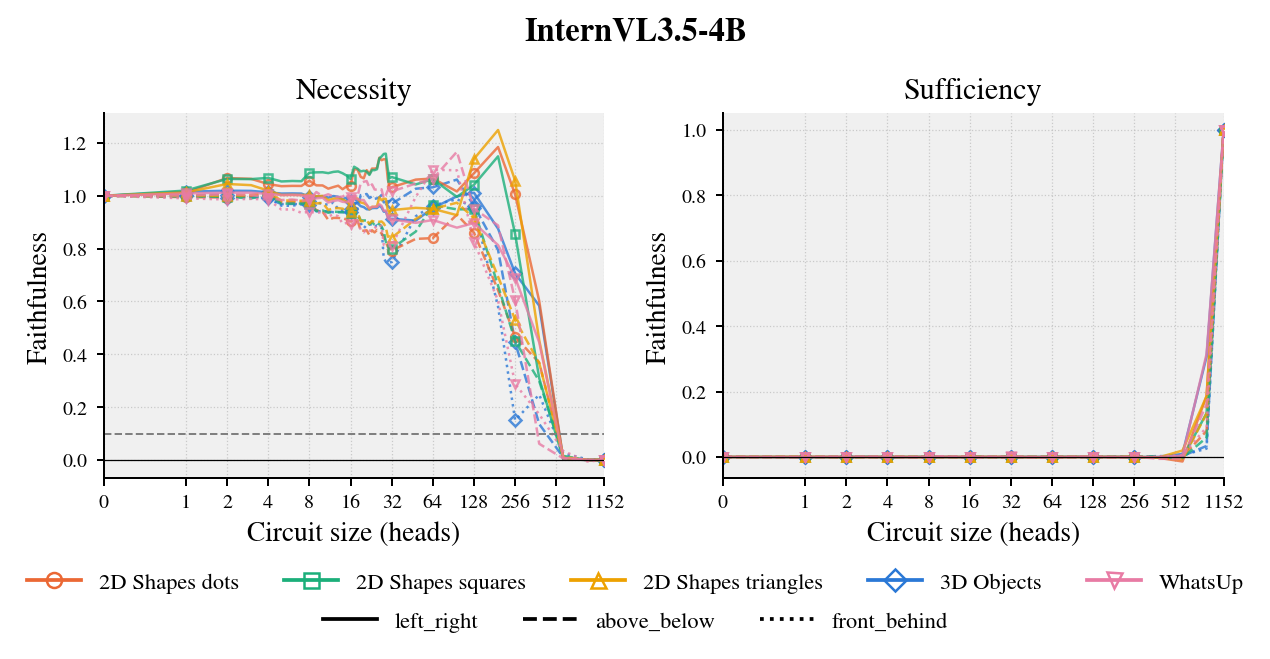}
    \includegraphics[width=\linewidth]{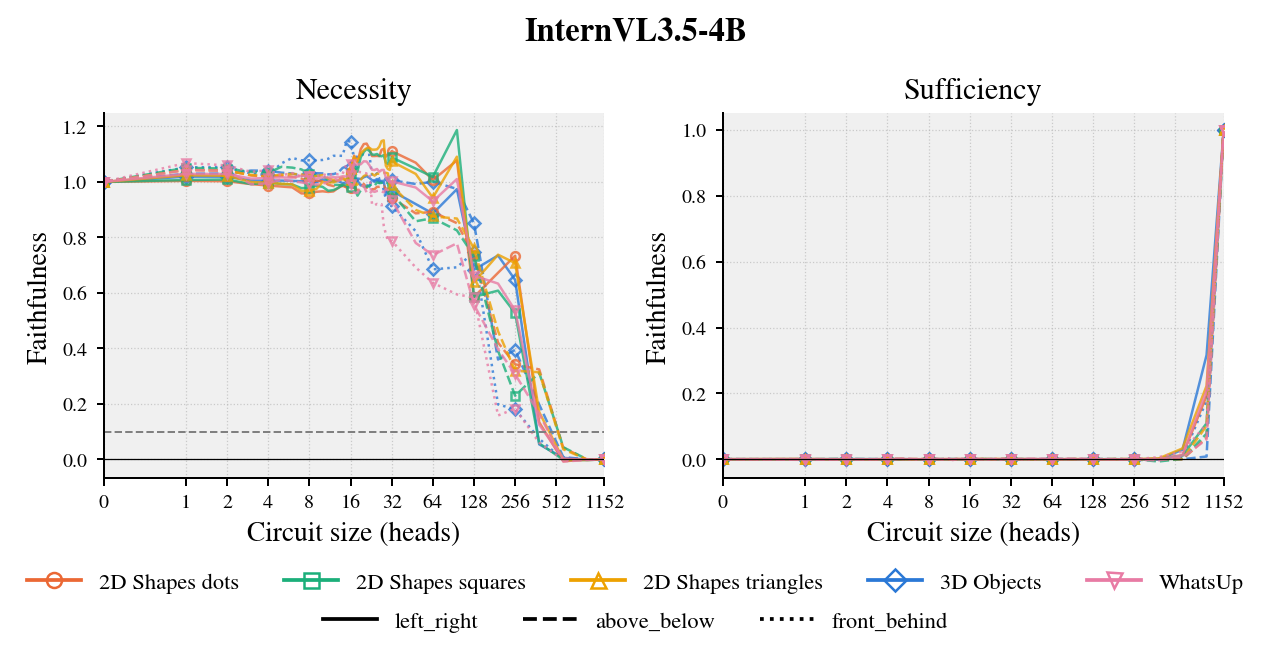}
 \caption{\textbf{Ablation - Faithfulness Analysis of InternVL with Random Head Rankings.}  We repeat the necessity (left) and sufficiency (right) analysis using three random head rankings. Compared to the recovery ranking, more heads must be ablated before faithfulness drops, which means that small circuits consisting of heads with high recovery scores describe the model's behavior more accurately.}
 \label{fig:faithfulness_internvl_control}
\end{figure}

\Cref{fig:faithfulness_ministral_control},~\Cref{fig:faithfulness_gemma_control}, and~\Cref{fig:faithfulness_internvl_control} visualize the results of Ministral, Gemma, and InternVL, respectively. Ministral's necessity curves for the random circuits vary more strongly across the datasets than for the circuits including recovery-ranked heads. Additionally, the results vary across control runs. Necessity still lies close to one in all control runs when we ablate $7$ random heads. The model's behavior in directional relation tasks breaks (i.e., faithfulness $\leq 0.1$) when around $200$ heads are ablated, which corresponds to $\approx 24\%$ of all heads in the LM decoder. Analyzing necessity of Gemma shows outlier curves for the same datasets as in the recovery-ranked faithfulness analysis. Similar to Ministral's results, ablating a circuit of random heads has a weak impact on the model's performance (i.e., necessity stays at 1) for most circuit sizes and converges to zero only when we ablate a large set of heads. Compared to recovery-ranked circuits, InternVL's random necessity curves do not decrease steeply for small circuit sizes. The model's behavior collapses when we ablate many heads. While ablating the top-$12$ heads with the highest recovery scores has a drastic impact, ablating $12$ random heads does not decrease faithfulness drastically.~\Cref{tab:control-k10} summarizes the $k_{10}$ values for all models and datasets. The random-ranking median $k_{10}$ across datasets is 91, 576, and 208 for Gemma, InternVL, and Ministral, respectively. These numbers correspond to $33.5\%$, $50\%$, and $25\%$ of the LM decoder heads in Gemma, InternVL, and Ministral, respectively. In the control runs, the models demonstrate drastically higher $k_{10}$ values. This finding implies that the necessity collapse described in~\cref{subsec:faithfulness_experiment} and~\cref{sec:appendix_faithfulness_results} is attributable to specific heads, and not solely to circuit size.

The sufficiency increases above $0.5$ only when $70-100\%$ of the heads are not ablated. Thus, the sets of unablated heads that restore the model's behavior largely overlap between different head rankings. Consequently, the random circuits show similar sufficiency curves to the recovery-ranked circuits.

\subsection{Projective steering results}
\label{sec:appendix_steering_ablation}
\begin{figure}[tb]
  \centering
  \includegraphics[width=\linewidth]{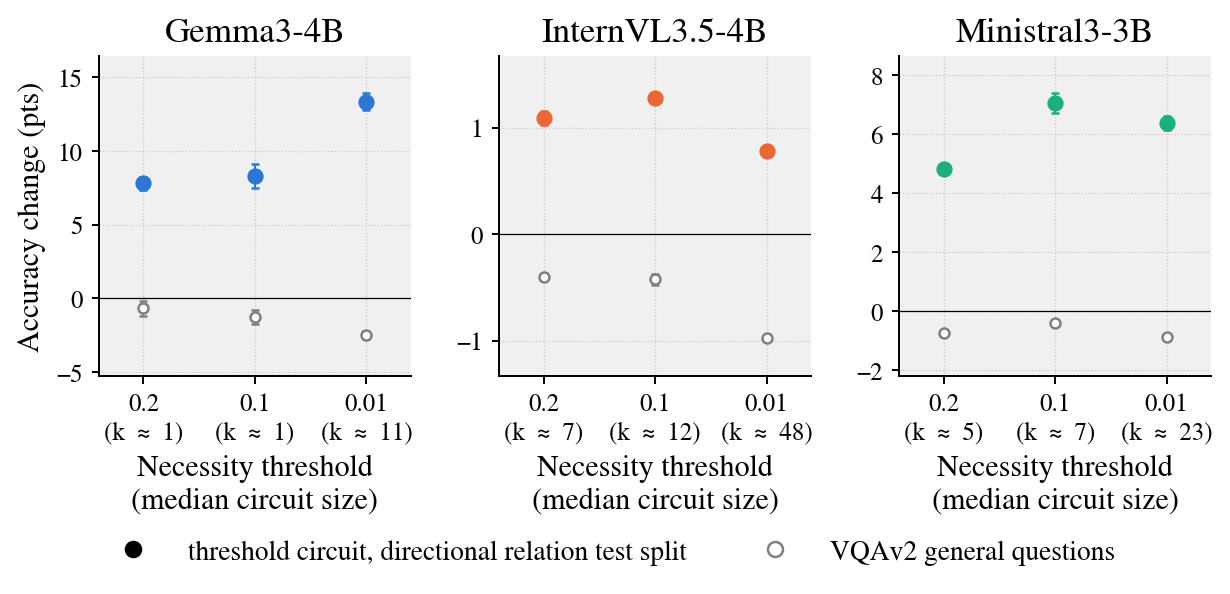}
  \caption{\textbf{Ablation - Faithfulness Threshold.} We steer the top-$k$ circuit for several faithfulness thresholds. Results show the accuracy change (in percentage points) on the directional relation test split (colored points) and on VQAv2 general questions (grey points), plotted against the median necessity faithfulness of the steered circuit. A faithfulness threshold of $0.1$ offers a good balance between improving performance on directional relation tasks and limiting degradation on VQAv2 across all models.}
  \label{fig:ablation_faith_thresh}
\end{figure}

\textbf{Faithfulness threshold.}\hspace{0.2cm}In the main experiment, we choose the median $k_{10}$, i.e., the smallest size $k$ at which necessity falls below $0.1$ and stays below $0.1$ at every larger circuit size, as a criterion to select the steering targets. Thus, we set the necessity faithfulness threshold to $0.1$. In this ablation study, we evaluate how sensitive steering is to this choice. We run the ablation with Gemma, InternVL, and Ministral. As datasets, we include all directional relation datasets and VQAv2 as a representative set for the general VQA tasks.
We test alternative thresholds, $0.2$ and $0.01$, for the following reasons. First, we double the threshold to $0.2$. At this threshold, $80\%$ of the model's behavior is removed, but the heads excluded from the corresponding circuit still have an effect. At threshold $0.1$, the excluded heads have a negligible impact. Hence, setting the threshold to $0.2$ tests whether a smaller circuit has the same effect as the circuit chosen based on the $0.1$ threshold. Additionally, we choose $0.01$ as a ``practical zero''. Setting the threshold to exactly $0$ is not possible because the necessity curve fluctuates and never stays strictly at $0$. Thus, 0.01 (i.e., $99\%$ of the model's behavior removed) is the smallest threshold at which the faithfulness stays below that value for all larger circuit sizes. For each threshold, we determine, per dataset, the circuit size $k$, which is the smallest number of top-ranked heads for which necessity stays below (i.e., falls below the threshold at that size and does not rise above it at any larger size). We compute the median of $k$ across datasets. Then, we steer the corresponding top-$k$ heads using the same steering parameters as in the main experiment. We repeat the experiment three times with three random seeds and report the accuracy change in percentage points averaged over seeds.~\Cref{fig:ablation_faith_thresh} presents the results. It visualizes accuracy change per faithfulness threshold on the directional relation test sets (blue, orange, and green points) and on the VQAv2 dataset (gray points). Gemma achieves the highest accuracy gain ($13.3$\,pp) at threshold $0.01$, but the performance degradation on VQAv2 also reaches its maximum ($-2.5$\,pp). Thus, a threshold of $0.1$ is a good trade-off, as it increases the accuracy gain compared to the threshold $0.2$ and harms performance on VQAv2 less than the $0.01$-threshold. The threshold of $0.2$ yields the smallest accuracy drop on VQAv2 for Gemma and could have been chosen as a conservative alternative to $0.1$. In contrast, InternVL and Ministral show higher accuracy gains on directional relation tasks at threshold $0.1$ without increasing harm on VQAv2. For these two models, threshold $0.1$ is the best choice regarding performance on directional relation and VQAv2 tasks. While the decrease on VQAv2 stays comparable at thresholds of $0.2$ and $0.1$, the accuracy gain on the directional relation tasks increases. Setting the threshold to $0.01$ leads to a larger set of steering targets (48 heads for InternVL, 23 heads for Ministral) without improving performance. 

Lastly, we perform a run where we steer only the top-$1$ head in InternVL and Ministral (corresponding faithfulness thresholds of $0.8$ and $0.7$, respectively). For InternVL, the accuracy changes by $0.1$\,pp on directional relation prediction and by $-0.2$\,pp on VQAv2. For Ministral, the accuracy changes by $4.2$\,pp on directional relation prediction and by $-0.7$\,pp on VQAv2. This ablation study underlines that setting the faithfulness threshold to $0.1$ is a valuable choice.

\textbf{Steering random heads.}\hspace{0.2cm}We run an ablation study where we steer a set of random heads with projective steering. The number of heads is the same as in the steering experiment with causally important heads ($1$ head for Gemma, $7$ heads for Ministral, $12$ heads for InternVL). The heads are sampled uniformly at random from all LM decoder heads, excluding those steered in the main experiment. We use the same training set to compute the steering parameters $\mu$ and $v$, and use the same steering strength, $\alpha=2.75$. We run each experiment five times with five different seeds to report the mean and standard deviation of accuracy and F1 score across five runs.

~\Cref{tab:acc-control-gemma} to~\Cref{tab:f1-control-vqa-ministral} summarize the results, which show that both performance improvements on directional relation tasks and side effects on general VQA tasks depend on the steered heads. Steering random heads does not reproduce our gains in accuracy and F1 score. On average, the accuracy changes by $0.1$\,pp, $-0.1$\,pp, and $-0.4$\,pp on the directional relation tasks for Gemma, InternVL, and Ministral, respectively. For example, the control runs decrease Ministral's accuracy on the 2D Shapes dataset, where the main steering experiment achieves accuracy gains up to $21.3$\,pp on average. The F1 scores change on average by $0.1$\,pp, $-0.1$\,pp, and $-0.5$\,pp on the directional relation tasks for Gemma, InternVL, and Ministral, respectively. These changes are drastically lower than for the causally important heads (accuracy gain: $7.9$\,pp for Gemma, $1.4$\,pp for InternVL, $6.5$\,pp for Ministral; F1 score gain: $7.2$\,pp for Gemma, $6.3$\,pp for Ministral, and $2.0$\,pp for InternVL). Moreover, the gains in the control runs lie within the std in most cases, whereas the gains from steering causal heads exceed the std. The control runs show high std overall, which means steering random heads leads to unstable, inconsistent, and unpredictable results. The hyperparameter search results and the results in~\cref{tab:acc-internvl} show that over-steering the directional relation heads can also negatively impact directional relation prediction, which underlines their crucial role in the task. In contrast, the random heads have a less negative effect as they are largely unrelated to the task. On general VQA tasks, accuracy and F1 change on average by $0.0$\,pp for Gemma, $-0.1$\,pp for InternVL, and $0.1$\,pp for Ministral. 

In summary, this ablation study shows that the performance improvements on the directional relation tasks reported in the main experiment are attributable to specific, causally relevant heads rather than to steering a certain number of heads. However, the study also suggests that the directional relation heads have a stronger impact on general VQA capabilities than random heads. We hypothesize that steering directional relation heads can have fewer side effects on general VQA tasks when we lower the steering strength. We leave the investigation of the relationship between different steering strengths and general VQA performance for future work.

\clearpage

\begin{table*}
\centering
\caption{\textbf{Accuracy (\%) of Gemma-3-4B.} Results on the directional relation test sets (columns) achieved by the different methods (rows). Results are the mean $\pm$ standard deviation over five seeds. Arrows give the change of the mean against the unsteered first row. Best mean per column in bold.}
\label{tab:acc-gemma}
\small
\fithalves{%
\begin{tabular}{l cccccc}\toprule
 & \multicolumn{6}{c}{2D Shapes}\\
\cmidrule(lr){2-7}
 & \multicolumn{2}{c}{dots} & \multicolumn{2}{c}{squares} & \multicolumn{2}{c}{triangles} \\
\cmidrule(lr){2-3}\cmidrule(lr){4-5}\cmidrule(lr){6-7}
Method & L--R & A--B & L--R & A--B & L--R & A--B \\
\midrule
Gemma-3-4B (unsteered) & 87.7\fstd{0.8} & 78.5\fstd{1.0} & 88.3\fstd{1.1} & 63.4\fstd{0.9} & 84.5\fstd{0.5} & 59.6\fstd{0.4} \\
\textsc{ScalingVis}~\cite{adaptvis} & 88.1\fstd{0.9}\fup{0.4} & 95.4\fstd{0.5}\fup{16.9} & 90.0\fstd{0.9}\fup{1.7} & 69.9\fstd{0.5}\fup{6.5} & 85.1\fstd{0.6}\fup{0.6} & \textbf{78.3}\fstd{1.1}\fup{18.7} \\
\textsc{AdaptVis}~\cite{adaptvis} & 87.8\fstd{0.5}\fup{0.1} & 95.4\fstd{0.5}\fup{16.9} & 90.0\fstd{0.9}\fup{1.7} & 69.7\fstd{0.5}\fup{6.3} & 85.1\fstd{0.6}\fup{0.6} & \textbf{78.3}\fstd{1.1}\fup{18.7} \\
Shift steering~\cite{attention_in_space,ITI} & 87.9\fstd{0.7}\fup{0.2} & 75.5\fstd{0.6}\fdown{3.0} & 88.2\fstd{0.8}\fdown{0.1} & 63.4\fstd{0.9} & 83.9\fstd{0.5}\fdown{0.6} & 58.8\fstd{0.4}\fdown{0.8} \\
\rowcolor{gray!15}Projective steering & \textbf{98.6}\fstd{0.3}\fup{10.9} & \textbf{97.3}\fstd{0.3}\fup{18.8} & \textbf{92.3}\fstd{1.7}\fup{4.0} & \textbf{77.5}\fstd{0.3}\fup{14.1} & \textbf{98.6}\fstd{0.5}\fup{14.1} & 72.9\fstd{1.2}\fup{13.3} \\
\bottomrule\end{tabular}
}{%
\begin{tabular}{l ccc ccc}\toprule
 & \multicolumn{3}{c}{3D Objects} & \multicolumn{3}{c}{WhatsUp}\\
\cmidrule(lr){2-4}\cmidrule(lr){5-7}
Method & L--R & A--B & F--B & L--R & A--B & F--B \\
\midrule
Gemma-3-4B (unsteered) & 88.1\fstd{1.1} & 51.0\fstd{0.4} & \textbf{50.0}\fstd{0.1} & 70.5\fstd{0.7} & 61.3\fstd{1.0} & \textbf{50.4}\fstd{0.2} \\
\textsc{ScalingVis}~\cite{adaptvis} & 89.2\fstd{1.1}\fup{1.1} & 53.2\fstd{0.4}\fup{2.2} & 49.9\fstd{0.6}\fdown{0.1} & 70.6\fstd{0.6}\fup{0.1} & 62.9\fstd{1.0}\fup{1.6} & 50.2\fstd{0.4}\fdown{0.2} \\
\textsc{AdaptVis}~\cite{adaptvis} & 89.2\fstd{1.1}\fup{1.1} & 53.2\fstd{0.4}\fup{2.2} & 49.9\fstd{0.3}\fdown{0.1} & 70.6\fstd{0.7}\fup{0.1} & 62.5\fstd{1.6}\fup{1.2} & 50.2\fstd{0.4}\fdown{0.2} \\
Shift steering~\cite{attention_in_space,ITI} & 87.5\fstd{0.8}\fdown{0.6} & 50.8\fstd{0.3}\fdown{0.2} & 49.9\fstd{0.2}\fdown{0.1} & 70.2\fstd{0.7}\fdown{0.3} & 61.3\fstd{1.0} & \textbf{50.4}\fstd{0.2} \\
\rowcolor{gray!15}Projective steering & \textbf{93.3}\fstd{0.8}\fup{5.2} & \textbf{53.8}\fstd{0.5}\fup{2.8} & 49.9\fstd{0.2}\fdown{0.1} & \textbf{73.9}\fstd{1.4}\fup{3.4} & \textbf{69.6}\fstd{0.8}\fup{8.3} & \textbf{50.4}\fstd{0.2} \\
\bottomrule\end{tabular}
}
\end{table*}

\begin{table*}
\centering
\caption{\textbf{Accuracy (\%) of InternVL-3.5-4B.} Layout as in Table~\ref{tab:acc-gemma}. --: shift steering was not run, because every train stimulus is answered correctly for all five seeds, so the shift has no wrong answers to be fitted on.}
\label{tab:acc-internvl}
\small
\fithalves{%
\begin{tabular}{l cccccc}\toprule
 & \multicolumn{6}{c}{2D Shapes}\\
\cmidrule(lr){2-7}
 & \multicolumn{2}{c}{dots} & \multicolumn{2}{c}{squares} & \multicolumn{2}{c}{triangles} \\
\cmidrule(lr){2-3}\cmidrule(lr){4-5}\cmidrule(lr){6-7}
Method & L--R & A--B & L--R & A--B & L--R & A--B \\
\midrule
InternVL-3.5-4B (unsteered) & \textbf{100.0}\fstd{0.0} & \textbf{100.0}\fstd{0.0} & \textbf{100.0}\fstd{0.0} & \textbf{100.0}\fstd{0.0} & \textbf{100.0}\fstd{0.0} & \textbf{100.0}\fstd{0.0} \\
\textsc{ScalingVis}~\cite{adaptvis} & \textbf{100.0}\fstd{0.0} & \textbf{100.0}\fstd{0.0} & \textbf{100.0}\fstd{0.0} & \textbf{100.0}\fstd{0.0} & \textbf{100.0}\fstd{0.0} & \textbf{100.0}\fstd{0.0} \\
\textsc{AdaptVis}~\cite{adaptvis} & \textbf{100.0}\fstd{0.0} & \textbf{100.0}\fstd{0.0} & \textbf{100.0}\fstd{0.0} & \textbf{100.0}\fstd{0.0} & \textbf{100.0}\fstd{0.0} & \textbf{100.0}\fstd{0.0} \\
Shift steering~\cite{attention_in_space,ITI} & -- & -- & -- & -- & -- & -- \\
\rowcolor{gray!15}Projective steering & \textbf{100.0}\fstd{0.0} & 99.6\fstd{0.3}\fdown{0.4} & \textbf{100.0}\fstd{0.0} & \textbf{100.0}\fstd{0.0} & \textbf{100.0}\fstd{0.0} & \textbf{100.0}\fstd{0.0} \\
\bottomrule\end{tabular}
}{%
\begin{tabular}{l ccc ccc}\toprule
 & \multicolumn{3}{c}{3D Objects} & \multicolumn{3}{c}{WhatsUp}\\
\cmidrule(lr){2-4}\cmidrule(lr){5-7}
Method & L--R & A--B & F--B & L--R & A--B & F--B \\
\midrule
InternVL-3.5-4B (unsteered) & \textbf{100.0}\fstd{0.0} & \textbf{100.0}\fstd{0.0} & \textbf{100.0}\fstd{0.0} & \textbf{100.0}\fstd{0.0} & 62.7\fstd{1.3} & 95.8\fstd{0.8} \\
\textsc{ScalingVis}~\cite{adaptvis} & \textbf{100.0}\fstd{0.0} & \textbf{100.0}\fstd{0.0} & \textbf{100.0}\fstd{0.0} & \textbf{100.0}\fstd{0.0} & 64.0\fstd{0.9}\fup{1.3} & 97.3\fstd{1.0}\fup{1.5} \\
\textsc{AdaptVis}~\cite{adaptvis} & \textbf{100.0}\fstd{0.0} & \textbf{100.0}\fstd{0.0} & \textbf{100.0}\fstd{0.0} & \textbf{100.0}\fstd{0.0} & 64.0\fstd{0.9}\fup{1.3} & 97.3\fstd{1.0}\fup{1.5} \\
Shift steering~\cite{attention_in_space,ITI} & -- & -- & -- & -- & 62.5\fstd{1.2}\fdown{0.2} & 95.8\fstd{0.8} \\
\rowcolor{gray!15}Projective steering & \textbf{100.0}\fstd{0.0} & 99.9\fstd{0.1}\fdown{0.1} & \textbf{100.0}\fstd{0.0} & 99.8\fstd{0.0}\fdown{0.2} & \textbf{76.1}\fstd{1.4}\fup{13.4} & \textbf{99.9}\fstd{0.2}\fup{4.1} \\
\bottomrule\end{tabular}
}
\end{table*}

\begin{table*}
\centering
\caption{\textbf{Accuracy (\%) of Ministral-3-3B}. Layout as in Table~\ref{tab:acc-gemma}.  --: shift steering was not run, because every train stimulus is answered correctly for all five seeds and shift has no wrong answers to be fitted on.}
\label{tab:acc-ministral}
\small
\fithalves{%
\begin{tabular}{l cccccc}\toprule
 & \multicolumn{6}{c}{2D Shapes}\\
\cmidrule(lr){2-7}
 & \multicolumn{2}{c}{dots} & \multicolumn{2}{c}{squares} & \multicolumn{2}{c}{triangles} \\
\cmidrule(lr){2-3}\cmidrule(lr){4-5}\cmidrule(lr){6-7}
Method & L--R & A--B & L--R & A--B & L--R & A--B \\
\midrule
Ministral-3-3B (unsteered) & 67.3\fstd{0.8} & 99.3\fstd{0.2} & 69.0\fstd{0.8} & \textbf{100.0}\fstd{0.0} & 76.7\fstd{0.9} & \textbf{100.0}\fstd{0.0} \\
\textsc{ScalingVis}~\cite{adaptvis} & 65.3\fstd{0.7}\fdown{2.0} & 99.4\fstd{0.2}\fup{0.1} & 68.2\fstd{0.8}\fdown{0.8} & \textbf{100.0}\fstd{0.0} & 75.3\fstd{0.7}\fdown{1.4} & \textbf{100.0}\fstd{0.0} \\
\textsc{AdaptVis}~\cite{adaptvis} & 65.2\fstd{0.8}\fdown{2.1} & 99.3\fstd{0.3} & 68.2\fstd{0.8}\fdown{0.8} & \textbf{100.0}\fstd{0.0} & 75.5\fstd{0.8}\fdown{1.2} & \textbf{100.0}\fstd{0.0} \\
Shift steering~\cite{attention_in_space,ITI} & 66.1\fstd{1.0}\fdown{1.2} & 99.2\fstd{0.2}\fdown{0.1} & 68.7\fstd{0.8}\fdown{0.3} & -- & 76.4\fstd{1.1}\fdown{0.3} & -- \\
\rowcolor{gray!15}Projective steering & \textbf{88.6}\fstd{1.1}\fup{21.3} & \textbf{100.0}\fstd{0.0}\fup{0.7} & \textbf{88.2}\fstd{1.0}\fup{19.2} & \textbf{100.0}\fstd{0.0} & \textbf{89.1}\fstd{1.5}\fup{12.4} & \textbf{100.0}\fstd{0.0} \\
\bottomrule\end{tabular}
}{%
\begin{tabular}{l ccc ccc}\toprule
 & \multicolumn{3}{c}{3D Objects} & \multicolumn{3}{c}{WhatsUp}\\
\cmidrule(lr){2-4}\cmidrule(lr){5-7}
Method & L--R & A--B & F--B & L--R & A--B & F--B \\
\midrule
Ministral-3-3B (unsteered) & \textbf{100.0}\fstd{0.0} & \textbf{100.0}\fstd{0.0} & \textbf{100.0}\fstd{0.0} & 84.6\fstd{0.7} & 55.2\fstd{1.4} & 67.4\fstd{2.4} \\
\textsc{ScalingVis}~\cite{adaptvis} & \textbf{100.0}\fstd{0.0} & \textbf{100.0}\fstd{0.0} & \textbf{100.0}\fstd{0.0} & 84.0\fstd{0.9}\fdown{0.6} & 57.0\fstd{0.9}\fup{1.8} & 69.5\fstd{2.0}\fup{2.1} \\
\textsc{AdaptVis}~\cite{adaptvis} & \textbf{100.0}\fstd{0.0} & \textbf{100.0}\fstd{0.0} & \textbf{100.0}\fstd{0.0} & 84.9\fstd{1.0}\fup{0.3} & 57.4\fstd{1.0}\fup{2.2} & 69.2\fstd{2.1}\fup{1.8} \\
Shift steering~\cite{attention_in_space,ITI} & -- & -- & -- & \textbf{85.3}\fstd{0.8}\fup{0.7} & 54.9\fstd{1.3}\fdown{0.3} & 67.0\fstd{2.3}\fdown{0.4} \\
\rowcolor{gray!15}Projective steering & \textbf{100.0}\fstd{0.0} & 97.7\fstd{0.4}\fdown{2.3} & \textbf{100.0}\fstd{0.0} & 84.1\fstd{1.2}\fdown{0.5} & \textbf{63.1}\fstd{1.6}\fup{7.9} & \textbf{87.6}\fstd{1.8}\fup{20.2} \\
\bottomrule\end{tabular}
}
\end{table*}

\begin{table*}[tp]\centering
\caption{\textbf{Accuracy (\%) on GQA~\cite{GQA} and VQAv2~\cite{VQA} for Gemma-3-4B.} Steered with the steering parameters fitted on the directional relation training set of each column. Mean $\pm$ standard deviation over five seeds. Arrows give the change of the mean against the unsteered row of the same dataset. Best mean per column and dataset in bold. }
\label{tab:acc-vqa-gemma}
\small
\fithalves{%
\begin{tabular}{l cccccc}\toprule
 & \multicolumn{6}{c}{2D Shapes}\\
\cmidrule(lr){2-7}
 & \multicolumn{2}{c}{dots} & \multicolumn{2}{c}{squares} & \multicolumn{2}{c}{triangles} \\
\cmidrule(lr){2-3}\cmidrule(lr){4-5}\cmidrule(lr){6-7}
Method & L--R & A--B & L--R & A--B & L--R & A--B \\
\midrule
GQA (unsteered) & \textbf{60.6}\fstd{0.0} & 60.6\fstd{0.0} & \textbf{60.6}\fstd{0.0} & 60.6\fstd{0.0} & \textbf{60.6}\fstd{0.0} & 60.6\fstd{0.0} \\
\textsc{ScalingVis}~\cite{adaptvis} & 59.2\fstd{0.0}\fdown{1.4} & 56.7\fstd{0.0}\fdown{3.9} & 59.2\fstd{0.0}\fdown{1.4} & 56.7\fstd{0.0}\fdown{3.9} & 59.2\fstd{0.0}\fdown{1.4} & 56.7\fstd{0.0}\fdown{3.9} \\
\textsc{AdaptVis}~\cite{adaptvis} & 59.3\fstd{0.3}\fdown{1.3} & 56.7\fstd{0.0}\fdown{3.9} & 59.2\fstd{0.0}\fdown{1.4} & 56.8\fstd{0.1}\fdown{3.8} & 59.2\fstd{0.0}\fdown{1.4} & 56.7\fstd{0.0}\fdown{3.9} \\
Shift steering~\cite{attention_in_space,ITI} & \textbf{60.6}\fstd{0.1} & 60.5\fstd{0.0}\fdown{0.1} & \textbf{60.6}\fstd{0.0} & 60.5\fstd{0.0}\fdown{0.1} & \textbf{60.6}\fstd{0.0} & 60.5\fstd{0.1}\fdown{0.1} \\
\rowcolor{gray!15}Projective steering & 60.5\fstd{0.1}\fdown{0.1} & \textbf{62.5}\fstd{0.1}\fup{1.9} & 60.5\fstd{0.1}\fdown{0.1} & \textbf{62.7}\fstd{0.1}\fup{2.1} & 60.1\fstd{0.2}\fdown{0.5} & \textbf{62.9}\fstd{0.1}\fup{2.3} \\
\bottomrule\end{tabular}
}{%
\begin{tabular}{l ccc ccc}\toprule
 & \multicolumn{3}{c}{3D Objects} & \multicolumn{3}{c}{WhatsUp}\\
\cmidrule(lr){2-4}\cmidrule(lr){5-7}
Method & L--R & A--B & F--B & L--R & A--B & F--B \\
\midrule
GQA (unsteered) & 60.4\fstd{0.0} & 60.4\fstd{0.0} & 60.4\fstd{0.0} & 60.4\fstd{0.0} & 60.4\fstd{0.0} & 60.4\fstd{0.0} \\
\textsc{ScalingVis}~\cite{adaptvis} & 58.9\fstd{0.0}\fdown{1.5} & 54.4\fstd{0.0}\fdown{6.0} & 59.9\fstd{3.0}\fdown{0.5} & 60.8\fstd{1.0}\fup{0.4} & 58.9\fstd{0.0}\fdown{1.5} & 57.1\fstd{3.7}\fdown{3.3} \\
\textsc{AdaptVis}~\cite{adaptvis} & 58.9\fstd{0.0}\fdown{1.5} & 54.5\fstd{0.1}\fdown{5.9} & 59.4\fstd{1.7}\fdown{1.0} & 59.2\fstd{0.1}\fdown{1.2} & 59.0\fstd{0.2}\fdown{1.4} & 56.2\fstd{2.5}\fdown{4.2} \\
Shift steering~\cite{attention_in_space,ITI} & 60.6\fstd{0.0}\fup{0.2} & 60.5\fstd{0.0}\fup{0.1} & 60.5\fstd{0.0}\fup{0.1} & 60.6\fstd{0.1}\fup{0.2} & \textbf{60.7}\fstd{0.0}\fup{0.3} & \textbf{60.6}\fstd{0.0}\fup{0.2} \\
\rowcolor{gray!15}Projective steering & \textbf{60.8}\fstd{0.1}\fup{0.4} & \textbf{62.0}\fstd{0.1}\fup{1.6} & \textbf{60.6}\fstd{0.0}\fup{0.2} & \textbf{60.9}\fstd{0.1}\fup{0.5} & 59.6\fstd{0.2}\fdown{0.8} & 60.3\fstd{0.2}\fdown{0.1} \\
\midrule
VQAv2 (unsteered) & 81.6\fstd{0.0} & 81.6\fstd{0.0} & 81.6\fstd{0.0} & 81.6\fstd{0.0} & 81.6\fstd{0.0} & \textbf{81.6}\fstd{0.0} \\
\bottomrule\end{tabular}
}
\end{table*}

\begin{table*}[tp]\centering
\caption{\textbf{Accuracy (\%) on POPE (random, popular, adversarial)~\cite{POPE} for Gemma-3-4B.} Layout as in Table~\ref{tab:acc-vqa-gemma}. }
\label{tab:acc-pope-gemma}
\small
\fithalves{%
\begin{tabular}{l cccccc}\toprule
 & \multicolumn{6}{c}{2D Shapes}\\
\cmidrule(lr){2-7}
 & \multicolumn{2}{c}{dots} & \multicolumn{2}{c}{squares} & \multicolumn{2}{c}{triangles} \\
\cmidrule(lr){2-3}\cmidrule(lr){4-5}\cmidrule(lr){6-7}
Method & L--R & A--B & L--R & A--B & L--R & A--B \\
\midrule
POPE random (unsteered) & \textbf{88.9}\fstd{0.0} & \textbf{88.9}\fstd{0.0} & \textbf{88.9}\fstd{0.0} & \textbf{88.9}\fstd{0.0} & \textbf{88.9}\fstd{0.0} & \textbf{88.9}\fstd{0.0} \\
\textsc{ScalingVis}~\cite{adaptvis} & 88.2\fstd{0.0}\fdown{0.7} & 82.9\fstd{0.0}\fdown{6.0} & 88.2\fstd{0.0}\fdown{0.7} & 82.9\fstd{0.0}\fdown{6.0} & 88.2\fstd{0.0}\fdown{0.7} & 82.9\fstd{0.0}\fdown{6.0} \\
\textsc{AdaptVis}~\cite{adaptvis} & 88.2\fstd{0.0}\fdown{0.7} & 82.9\fstd{0.0}\fdown{6.0} & 88.2\fstd{0.0}\fdown{0.7} & 82.9\fstd{0.0}\fdown{6.0} & 88.2\fstd{0.0}\fdown{0.7} & 82.9\fstd{0.0}\fdown{6.0} \\
Shift steering~\cite{attention_in_space,ITI} & \textbf{88.9}\fstd{0.0} & 88.8\fstd{0.0}\fdown{0.1} & \textbf{88.9}\fstd{0.0} & 88.8\fstd{0.0}\fdown{0.1} & \textbf{88.9}\fstd{0.0} & 88.8\fstd{0.0}\fdown{0.1} \\
\rowcolor{gray!15}Projective steering & 88.7\fstd{0.1}\fdown{0.2} & 88.4\fstd{0.0}\fdown{0.5} & 88.8\fstd{0.0}\fdown{0.1} & 88.4\fstd{0.0}\fdown{0.5} & 88.8\fstd{0.0}\fdown{0.1} & 88.5\fstd{0.0}\fdown{0.4} \\
\bottomrule\end{tabular}
}{%
\begin{tabular}{l ccc ccc}\toprule
 & \multicolumn{3}{c}{3D Objects} & \multicolumn{3}{c}{WhatsUp}\\
\cmidrule(lr){2-4}\cmidrule(lr){5-7}
Method & L--R & A--B & F--B & L--R & A--B & F--B \\
\midrule
POPE random (unsteered) & \textbf{88.9}\fstd{0.0} & \textbf{88.9}\fstd{0.0} & \textbf{88.9}\fstd{0.0} & 88.9\fstd{0.0} & \textbf{88.9}\fstd{0.0} & \textbf{88.9}\fstd{0.0} \\
\textsc{ScalingVis}~\cite{adaptvis} & 88.1\fstd{0.0}\fdown{0.8} & 73.9\fstd{0.0}\fdown{15.0} & 85.5\fstd{6.5}\fdown{3.4} & 88.5\fstd{0.2}\fdown{0.4} & 88.1\fstd{0.0}\fdown{0.8} & 79.7\fstd{8.0}\fdown{9.2} \\
\textsc{AdaptVis}~\cite{adaptvis} & 88.1\fstd{0.0}\fdown{0.8} & 73.9\fstd{0.0}\fdown{15.0} & 82.9\fstd{4.9}\fdown{6.0} & 88.1\fstd{0.0}\fdown{0.8} & 88.1\fstd{0.0}\fdown{0.8} & 79.6\fstd{7.8}\fdown{9.3} \\
Shift steering~\cite{attention_in_space,ITI} & \textbf{88.9}\fstd{0.0} & \textbf{88.9}\fstd{0.0} & \textbf{88.9}\fstd{0.0} & \textbf{89.0}\fstd{0.0}\fup{0.1} & \textbf{88.9}\fstd{0.0} & \textbf{88.9}\fstd{0.0} \\
\rowcolor{gray!15}Projective steering & 88.7\fstd{0.0}\fdown{0.2} & 88.6\fstd{0.0}\fdown{0.3} & 88.8\fstd{0.0}\fdown{0.1} & 88.7\fstd{0.0}\fdown{0.2} & 88.5\fstd{0.0}\fdown{0.4} & 88.8\fstd{0.1}\fdown{0.1} \\
\midrule
POPE popular (unsteered) & \textbf{83.2}\fstd{0.0} & 83.2\fstd{0.0} & 83.2\fstd{0.0} & 83.2\fstd{0.0} & 83.2\fstd{0.0} & \textbf{83.2}\fstd{0.0} \\
\bottomrule\end{tabular}
}
\end{table*}

\begin{table*}[tp]\centering
\caption{\textbf{Accuracy (\%) on GQA and VQAv2 for InternVL-3.5-4B}. Layout as in Table~\ref{tab:acc-vqa-gemma}. --: shift steering not run, because every train stimulus is answered correctly for all five seeds, so the shift has no wrong answers to be fitted on.}
\label{tab:acc-vqa-internvl}
\small
\fithalves{%
\begin{tabular}{l cccccc}\toprule
 & \multicolumn{6}{c}{2D Shapes}\\
\cmidrule(lr){2-7}
 & \multicolumn{2}{c}{dots} & \multicolumn{2}{c}{squares} & \multicolumn{2}{c}{triangles} \\
\cmidrule(lr){2-3}\cmidrule(lr){4-5}\cmidrule(lr){6-7}
Method & L--R & A--B & L--R & A--B & L--R & A--B \\
\midrule
GQA (unsteered) & \textbf{74.6}\fstd{0.0} & \textbf{74.6}\fstd{0.0} & \textbf{74.6}\fstd{0.0} & \textbf{74.6}\fstd{0.0} & \textbf{74.6}\fstd{0.0} & \textbf{74.6}\fstd{0.0} \\
\textsc{ScalingVis}~\cite{adaptvis} & 74.2\fstd{0.0}\fdown{0.4} & 74.2\fstd{0.0}\fdown{0.4} & 74.2\fstd{0.0}\fdown{0.4} & 74.2\fstd{0.0}\fdown{0.4} & 74.2\fstd{0.0}\fdown{0.4} & 74.2\fstd{0.0}\fdown{0.4} \\
\textsc{AdaptVis}~\cite{adaptvis} & 74.3\fstd{0.0}\fdown{0.3} & 74.3\fstd{0.0}\fdown{0.3} & 74.3\fstd{0.0}\fdown{0.3} & 74.3\fstd{0.0}\fdown{0.3} & 74.3\fstd{0.0}\fdown{0.3} & 74.3\fstd{0.0}\fdown{0.3} \\
Shift steering~\cite{attention_in_space,ITI} & -- & -- & -- & -- & -- & -- \\
\rowcolor{gray!15}Projective steering & 71.6\fstd{0.2}\fdown{3.0} & 72.3\fstd{0.1}\fdown{2.3} & 71.6\fstd{0.1}\fdown{3.0} & 72.5\fstd{0.2}\fdown{2.1} & 72.4\fstd{0.2}\fdown{2.2} & 72.3\fstd{0.1}\fdown{2.3} \\
\bottomrule\end{tabular}
}{%
\begin{tabular}{l ccc ccc}\toprule
 & \multicolumn{3}{c}{3D Objects} & \multicolumn{3}{c}{WhatsUp}\\
\cmidrule(lr){2-4}\cmidrule(lr){5-7}
Method & L--R & A--B & F--B & L--R & A--B & F--B \\
\midrule
GQA (unsteered) & \textbf{74.3}\fstd{0.0} & \textbf{74.3}\fstd{0.0} & \textbf{74.3}\fstd{0.0} & \textbf{74.3}\fstd{0.0} & 74.3\fstd{0.0} & 74.3\fstd{0.0} \\
\textsc{ScalingVis}~\cite{adaptvis} & 74.0\fstd{0.0}\fdown{0.3} & 74.0\fstd{0.0}\fdown{0.3} & 74.0\fstd{0.0}\fdown{0.3} & 74.1\fstd{0.0}\fdown{0.2} & 74.3\fstd{0.0} & 74.0\fstd{0.0}\fdown{0.3} \\
\textsc{AdaptVis}~\cite{adaptvis} & 74.0\fstd{0.0}\fdown{0.3} & 74.0\fstd{0.0}\fdown{0.3} & 74.0\fstd{0.0}\fdown{0.3} & \textbf{74.3}\fstd{0.0} & 74.3\fstd{0.0} & 74.0\fstd{0.0}\fdown{0.3} \\
Shift steering~\cite{attention_in_space,ITI} & -- & -- & -- & -- & \textbf{74.4}\fstd{0.2}\fup{0.1} & \textbf{74.5}\fstd{0.1}\fup{0.2} \\
\rowcolor{gray!15}Projective steering & 73.1\fstd{0.1}\fdown{1.2} & 71.2\fstd{0.1}\fdown{3.1} & 71.0\fstd{0.1}\fdown{3.3} & 72.1\fstd{0.2}\fdown{2.2} & 72.3\fstd{0.2}\fdown{2.0} & 72.0\fstd{0.1}\fdown{2.3} \\
\midrule
VQAv2 (unsteered) & \textbf{88.7}\fstd{0.0} & \textbf{88.7}\fstd{0.0} & \textbf{88.7}\fstd{0.0} & \textbf{88.7}\fstd{0.0} & \textbf{88.7}\fstd{0.0} & \textbf{88.7}\fstd{0.0} \\
\bottomrule\end{tabular}
}
\end{table*}

\begin{table*}[tp]\centering
\caption{\textbf{Accuracy (\%) on POPE (random, popular, adversarial) for InternVL-3.5-4B.} Layout as in Table~\ref{tab:acc-vqa-gemma}. --: shift steering not run, because every train stimulus is answered correctly for all five seeds, so the shift has no wrong answers to be fitted on.}
\label{tab:acc-pope-internvl}
\small
\fithalves{%
\begin{tabular}{l cccccc}\toprule
 & \multicolumn{6}{c}{2D Shapes}\\
\cmidrule(lr){2-7}
 & \multicolumn{2}{c}{dots} & \multicolumn{2}{c}{squares} & \multicolumn{2}{c}{triangles} \\
\cmidrule(lr){2-3}\cmidrule(lr){4-5}\cmidrule(lr){6-7}
Method & L--R & A--B & L--R & A--B & L--R & A--B \\
\midrule
POPE random (unsteered) & 91.9\fstd{0.0} & 91.9\fstd{0.0} & 91.9\fstd{0.0} & 91.9\fstd{0.0} & 91.9\fstd{0.0} & 91.9\fstd{0.0} \\
\textsc{ScalingVis}~\cite{adaptvis} & 91.7\fstd{0.0}\fdown{0.2} & 91.7\fstd{0.0}\fdown{0.2} & 91.7\fstd{0.0}\fdown{0.2} & 91.7\fstd{0.0}\fdown{0.2} & 91.7\fstd{0.0}\fdown{0.2} & 91.7\fstd{0.0}\fdown{0.2} \\
\textsc{AdaptVis}~\cite{adaptvis} & \textbf{92.1}\fstd{0.0}\fup{0.2} & \textbf{92.1}\fstd{0.0}\fup{0.2} & \textbf{92.1}\fstd{0.0}\fup{0.2} & \textbf{92.1}\fstd{0.0}\fup{0.2} & \textbf{92.1}\fstd{0.0}\fup{0.2} & \textbf{92.1}\fstd{0.0}\fup{0.2} \\
Shift steering~\cite{attention_in_space,ITI} & -- & -- & -- & -- & -- & -- \\
\rowcolor{gray!15}Projective steering & 90.9\fstd{0.1}\fdown{1.0} & 91.0\fstd{0.1}\fdown{0.9} & 91.0\fstd{0.1}\fdown{0.9} & 91.0\fstd{0.0}\fdown{0.9} & 91.0\fstd{0.0}\fdown{0.9} & 91.0\fstd{0.1}\fdown{0.9} \\
\bottomrule\end{tabular}
}{%
\begin{tabular}{l ccc ccc}\toprule
 & \multicolumn{3}{c}{3D Objects} & \multicolumn{3}{c}{WhatsUp}\\
\cmidrule(lr){2-4}\cmidrule(lr){5-7}
Method & L--R & A--B & F--B & L--R & A--B & F--B \\
\midrule
POPE random (unsteered) & 91.9\fstd{0.0} & 91.9\fstd{0.0} & 91.9\fstd{0.0} & 91.9\fstd{0.0} & 91.9\fstd{0.0} & \textbf{91.9}\fstd{0.0} \\
\textsc{ScalingVis}~\cite{adaptvis} & 91.7\fstd{0.0}\fdown{0.2} & 91.7\fstd{0.0}\fdown{0.2} & 91.7\fstd{0.0}\fdown{0.2} & 91.7\fstd{0.0}\fdown{0.2} & \textbf{92.1}\fstd{0.0}\fup{0.2} & \textbf{91.9}\fstd{0.0} \\
\textsc{AdaptVis}~\cite{adaptvis} & \textbf{92.1}\fstd{0.0}\fup{0.2} & \textbf{92.1}\fstd{0.0}\fup{0.2} & \textbf{92.1}\fstd{0.0}\fup{0.2} & \textbf{92.1}\fstd{0.0}\fup{0.2} & \textbf{92.1}\fstd{0.0}\fup{0.2} & \textbf{91.9}\fstd{0.0} \\
Shift steering~\cite{attention_in_space,ITI} & -- & -- & -- & -- & 91.9\fstd{0.0} & 91.8\fstd{0.1}\fdown{0.1} \\
\rowcolor{gray!15}Projective steering & 91.3\fstd{0.1}\fdown{0.6} & 90.6\fstd{0.0}\fdown{1.3} & 91.1\fstd{0.0}\fdown{0.8} & 90.9\fstd{0.0}\fdown{1.0} & 91.7\fstd{0.1}\fdown{0.2} & 91.5\fstd{0.1}\fdown{0.4} \\
\midrule
POPE popular (unsteered) & 86.5\fstd{0.0} & 86.5\fstd{0.0} & 86.5\fstd{0.0} & 86.5\fstd{0.0} & 86.5\fstd{0.0} & 86.5\fstd{0.0} \\
\bottomrule\end{tabular}
}
\end{table*}

\begin{table*}
\centering
\caption{\textbf{Accuracy (\%) on GQA and VQAv2 for Ministral-3-3B}. Layout as in Table~\ref{tab:acc-vqa-gemma}. --: shift steering not run, because every train stimulus is answered correctly for all five seeds, so the shift has no wrong answers to be fitted on.}
\label{tab:acc-vqa-ministral}
\small
\fithalves{%
\begin{tabular}{l cccccc}\toprule
 & \multicolumn{6}{c}{2D Shapes}\\
\cmidrule(lr){2-7}
 & \multicolumn{2}{c}{dots} & \multicolumn{2}{c}{squares} & \multicolumn{2}{c}{triangles} \\
\cmidrule(lr){2-3}\cmidrule(lr){4-5}\cmidrule(lr){6-7}
Method & L--R & A--B & L--R & A--B & L--R & A--B \\
\midrule
GQA (unsteered) & 64.2\fstd{0.0} & 64.2\fstd{0.0} & \textbf{64.2}\fstd{0.0} & 64.2\fstd{0.0} & 64.2\fstd{0.0} & \textbf{64.2}\fstd{0.0} \\
\textsc{ScalingVis}~\cite{adaptvis} & \textbf{64.3}\fstd{0.0}\fup{0.1} & \textbf{64.3}\fstd{0.0}\fup{0.1} & 64.1\fstd{0.0}\fdown{0.1} & 64.0\fstd{0.0}\fdown{0.2} & 64.1\fstd{0.1}\fdown{0.1} & 64.0\fstd{0.0}\fdown{0.2} \\
\textsc{AdaptVis}~\cite{adaptvis} & \textbf{64.3}\fstd{0.0}\fup{0.1} & \textbf{64.3}\fstd{0.0}\fup{0.1} & 64.1\fstd{0.0}\fdown{0.1} & 64.1\fstd{0.0}\fdown{0.1} & 64.1\fstd{0.0}\fdown{0.1} & 64.1\fstd{0.0}\fdown{0.1} \\
Shift steering~\cite{attention_in_space,ITI} & 64.1\fstd{0.1}\fdown{0.1} & 64.0\fstd{0.1}\fdown{0.2} & \textbf{64.2}\fstd{0.0} & -- & \textbf{64.3}\fstd{0.1}\fup{0.1} & -- \\
\rowcolor{gray!15}Projective steering & 64.0\fstd{0.1}\fdown{0.2} & 64.0\fstd{0.1}\fdown{0.2} & 63.6\fstd{0.2}\fdown{0.6} & \textbf{64.3}\fstd{0.1}\fup{0.1} & 64.1\fstd{0.2}\fdown{0.1} & 63.9\fstd{0.1}\fdown{0.3} \\
\bottomrule\end{tabular}
}{%
\begin{tabular}{l ccc ccc}\toprule
 & \multicolumn{3}{c}{3D Objects} & \multicolumn{3}{c}{WhatsUp}\\
\cmidrule(lr){2-4}\cmidrule(lr){5-7}
Method & L--R & A--B & F--B & L--R & A--B & F--B \\
\midrule
GQA (unsteered) & \textbf{64.0}\fstd{0.0} & \textbf{64.0}\fstd{0.0} & \textbf{64.0}\fstd{0.0} & 64.0\fstd{0.0} & \textbf{64.0}\fstd{0.0} & 64.0\fstd{0.0} \\
\textsc{ScalingVis}~\cite{adaptvis} & \textbf{64.0}\fstd{0.0} & \textbf{64.0}\fstd{0.0} & \textbf{64.0}\fstd{0.0} & 64.1\fstd{0.1}\fup{0.1} & 63.0\fstd{0.5}\fdown{1.0} & 62.8\fstd{0.0}\fdown{1.2} \\
\textsc{AdaptVis}~\cite{adaptvis} & \textbf{64.0}\fstd{0.0} & \textbf{64.0}\fstd{0.0} & \textbf{64.0}\fstd{0.0} & 64.0\fstd{0.0} & 63.2\fstd{0.6}\fdown{0.8} & 63.1\fstd{0.6}\fdown{0.9} \\
Shift steering~\cite{attention_in_space,ITI} & -- & -- & -- & \textbf{64.2}\fstd{0.1}\fup{0.2} & \textbf{64.0}\fstd{0.1} & \textbf{64.2}\fstd{0.1}\fup{0.2} \\
\rowcolor{gray!15}Projective steering & 63.9\fstd{0.0}\fdown{0.1} & 63.8\fstd{0.1}\fdown{0.2} & 63.0\fstd{0.1}\fdown{1.0} & 63.9\fstd{0.1}\fdown{0.1} & 63.6\fstd{0.2}\fdown{0.4} & 63.7\fstd{0.3}\fdown{0.3} \\
\midrule
VQAv2 (unsteered) & \textbf{83.9}\fstd{0.0} & \textbf{83.9}\fstd{0.0} & \textbf{83.9}\fstd{0.0} & \textbf{83.9}\fstd{0.0} & 83.9\fstd{0.0} & \textbf{83.9}\fstd{0.0} \\
\bottomrule\end{tabular}
}
\end{table*}

\begin{table*}
\centering
\caption{\textbf{Accuracy (\%) on POPE (random, popular, adversarial) for Ministral-3-3B.} Layout as in Table~\ref{tab:acc-vqa-gemma}. --: shift steering not run, because every train stimulus is answered correctly for all five seeds, so the shift has no wrong answers to be fitted on.}
\label{tab:acc-pope-ministral}
\small
\fithalves{%
\begin{tabular}{l cccccc}\toprule
 & \multicolumn{6}{c}{2D Shapes}\\
\cmidrule(lr){2-7}
 & \multicolumn{2}{c}{dots} & \multicolumn{2}{c}{squares} & \multicolumn{2}{c}{triangles} \\
\cmidrule(lr){2-3}\cmidrule(lr){4-5}\cmidrule(lr){6-7}
Method & L--R & A--B & L--R & A--B & L--R & A--B \\
\midrule
POPE random (unsteered) & 84.2\fstd{0.0} & 84.2\fstd{0.0} & 84.2\fstd{0.0} & 84.2\fstd{0.0} & 84.2\fstd{0.0} & 84.2\fstd{0.0} \\
\textsc{ScalingVis}~\cite{adaptvis} & \textbf{84.4}\fstd{0.0}\fup{0.2} & \textbf{84.4}\fstd{0.0}\fup{0.2} & 84.2\fstd{0.0} & \textbf{84.9}\fstd{0.0}\fup{0.7} & 84.2\fstd{0.1} & \textbf{84.9}\fstd{0.0}\fup{0.7} \\
\textsc{AdaptVis}~\cite{adaptvis} & \textbf{84.4}\fstd{0.0}\fup{0.2} & \textbf{84.4}\fstd{0.0}\fup{0.2} & 84.2\fstd{0.0} & 84.2\fstd{0.0} & 84.2\fstd{0.0} & 84.2\fstd{0.0} \\
Shift steering~\cite{attention_in_space,ITI} & 84.3\fstd{0.0}\fup{0.1} & 84.3\fstd{0.0}\fup{0.1} & \textbf{84.4}\fstd{0.1}\fup{0.2} & -- & \textbf{84.4}\fstd{0.1}\fup{0.2} & -- \\
\rowcolor{gray!15}Projective steering & 84.1\fstd{0.1}\fdown{0.1} & 83.0\fstd{0.1}\fdown{1.2} & 83.5\fstd{0.0}\fdown{0.7} & 83.7\fstd{0.0}\fdown{0.5} & 83.5\fstd{0.1}\fdown{0.7} & 83.4\fstd{0.0}\fdown{0.8} \\
\bottomrule\end{tabular}
}{%
\begin{tabular}{l ccc ccc}\toprule
 & \multicolumn{3}{c}{3D Objects} & \multicolumn{3}{c}{WhatsUp}\\
\cmidrule(lr){2-4}\cmidrule(lr){5-7}
Method & L--R & A--B & F--B & L--R & A--B & F--B \\
\midrule
POPE random (unsteered) & 84.2\fstd{0.0} & 84.2\fstd{0.0} & 84.2\fstd{0.0} & 84.2\fstd{0.0} & 84.2\fstd{0.0} & 84.2\fstd{0.0} \\
\textsc{ScalingVis}~\cite{adaptvis} & \textbf{84.9}\fstd{0.0}\fup{0.7} & \textbf{84.9}\fstd{0.0}\fup{0.7} & \textbf{84.9}\fstd{0.0}\fup{0.7} & 84.3\fstd{0.1}\fup{0.1} & 83.7\fstd{0.7}\fdown{0.5} & 83.4\fstd{0.0}\fdown{0.8} \\
\textsc{AdaptVis}~\cite{adaptvis} & 84.2\fstd{0.0} & 84.2\fstd{0.0} & 84.2\fstd{0.0} & \textbf{84.9}\fstd{0.0}\fup{0.7} & 83.9\fstd{0.8}\fdown{0.3} & 83.6\fstd{0.5}\fdown{0.6} \\
Shift steering~\cite{attention_in_space,ITI} & -- & -- & -- & 84.3\fstd{0.1}\fup{0.1} & 84.3\fstd{0.0}\fup{0.1} & \textbf{84.3}\fstd{0.0}\fup{0.1} \\
\rowcolor{gray!15}Projective steering & 83.3\fstd{0.0}\fdown{0.9} & 83.0\fstd{0.1}\fdown{1.2} & 83.3\fstd{0.1}\fdown{0.9} & 83.6\fstd{0.1}\fdown{0.6} & \textbf{84.4}\fstd{0.7}\fup{0.2} & 84.1\fstd{0.3}\fdown{0.1} \\
\midrule
POPE popular (unsteered) & 83.1\fstd{0.0} & 83.1\fstd{0.0} & 83.1\fstd{0.0} & 83.1\fstd{0.0} & \textbf{83.1}\fstd{0.0} & \textbf{83.1}\fstd{0.0} \\
\bottomrule\end{tabular}
}
\end{table*}

\begin{table*}[tp]\centering
\caption{\textbf{Macro-F1 ($\%$) on the directional relations test split for Gemma-3-4B.} Mean $\pm$ standard deviation over five random seeds. Arrows indicate the change of the mean against the unsteered baseline (first row). Best mean per column marked in bold. }
\label{tab:f1-gemma}
\small
\fithalves{%
\begin{tabular}{l cccccc}\toprule
 & \multicolumn{6}{c}{2D Shapes}\\
\cmidrule(lr){2-7}
 & \multicolumn{2}{c}{dots} & \multicolumn{2}{c}{squares} & \multicolumn{2}{c}{triangles} \\
\cmidrule(lr){2-3}\cmidrule(lr){4-5}\cmidrule(lr){6-7}
Method & L--R & A--B & L--R & A--B & L--R & A--B \\
\midrule
Gemma-3-4B (unsteered) & 87.7\fstd{0.8} & 77.4\fstd{1.1} & 88.1\fstd{1.1} & 57.8\fstd{1.3} & 84.2\fstd{0.6} & 51.7\fstd{0.7} \\
\textsc{ScalingVis}~\cite{adaptvis}& 88.0\fstd{0.9}\fup{0.3} & 95.4\fstd{0.5}\fup{18.0} & 90.0\fstd{0.9}\fup{1.9} & 66.9\fstd{0.7}\fup{9.1} & 84.9\fstd{0.7}\fup{0.7} & \textbf{77.3}\fstd{1.3}\fup{25.6} \\
\textsc{AdaptVis}~\cite{adaptvis}& 87.8\fstd{0.6}\fup{0.1} & 95.4\fstd{0.5}\fup{18.0} & 90.0\fstd{0.9}\fup{1.9} & 66.7\fstd{0.7}\fup{8.9} & 84.9\fstd{0.7}\fup{0.7} & \textbf{77.3}\fstd{1.3}\fup{25.6} \\
Shift steering~\cite{attention_in_space,ITI} & 87.9\fstd{0.7}\fup{0.2} & 74.0\fstd{0.7}\fdown{3.4} & 88.0\fstd{0.8}\fdown{0.1} & 57.7\fstd{1.4}\fdown{0.1} & 83.6\fstd{0.6}\fdown{0.6} & 50.4\fstd{0.6}\fdown{1.3} \\
\rowcolor{gray!15}Projective steering & \textbf{98.6}\fstd{0.3}\fup{10.9} & \textbf{97.3}\fstd{0.3}\fup{19.9} & \textbf{92.3}\fstd{1.7}\fup{4.2} & \textbf{76.3}\fstd{0.4}\fup{18.5} & \textbf{98.6}\fstd{0.5}\fup{14.4} & 70.7\fstd{1.5}\fup{19.0} \\
\bottomrule\end{tabular}
}{%
\begin{tabular}{l ccc ccc}\toprule
 & \multicolumn{3}{c}{3D Objects} & \multicolumn{3}{c}{WhatsUp}\\
\cmidrule(lr){2-4}\cmidrule(lr){5-7}
Method & L--R & A--B & F--B & L--R & A--B & F--B \\
\midrule
Gemma-3-4B (unsteered) & 87.9\fstd{1.1} & 35.4\fstd{0.8} & 33.8\fstd{0.2} & 68.3\fstd{0.8} & 54.7\fstd{1.7} & 34.2\fstd{0.5} \\
\textsc{ScalingVis}~\cite{adaptvis}& 89.1\fstd{1.2}\fup{1.2} & \textbf{50.5}\fstd{0.5}\fup{15.1} & 35.3\fstd{3.2}\fup{1.5} & 68.3\fstd{0.8} & 57.3\fstd{1.6}\fup{2.6} & \textbf{42.6}\fstd{8.1}\fup{8.4} \\
\textsc{AdaptVis}~\cite{adaptvis}& 89.1\fstd{1.2}\fup{1.2} & 50.4\fstd{0.5}\fup{15.0} & \textbf{42.1}\fstd{0.9}\fup{8.3} & 68.4\fstd{0.8}\fup{0.1} & 56.5\fstd{2.6}\fup{1.8} & \textbf{42.6}\fstd{8.1}\fup{8.4} \\
Shift steering~\cite{attention_in_space,ITI} & 87.3\fstd{0.8}\fdown{0.6} & 35.1\fstd{0.6}\fdown{0.3} & 33.7\fstd{0.2}\fdown{0.1} & 67.9\fstd{0.9}\fdown{0.4} & 54.7\fstd{1.7} & 34.2\fstd{0.5} \\
\rowcolor{gray!15}Projective steering & \textbf{93.3}\fstd{0.8}\fup{5.4} & 41.2\fstd{0.9}\fup{5.8} & 33.7\fstd{0.2}\fdown{0.1} & \textbf{73.5}\fstd{1.5}\fup{5.2} & \textbf{66.8}\fstd{0.8}\fup{12.1} & 34.2\fstd{0.5} \\
\bottomrule\end{tabular}
}
\end{table*}

\begin{table*}[tp]\centering
\caption{\textbf{Macro-F1 ($\%$) on the directional relations test split for InternVL-3.5-4B.} Layout as in Table~\ref{tab:f1-gemma}. --: shift steering not run, because every train sample is answered correctly, so the shift has no wrong answers and is undefined in this case.}
\label{tab:f1-internvl}
\small
\fithalves{%
\begin{tabular}{l cccccc}\toprule
 & \multicolumn{6}{c}{2D Shapes}\\
\cmidrule(lr){2-7}
 & \multicolumn{2}{c}{dots} & \multicolumn{2}{c}{squares} & \multicolumn{2}{c}{triangles} \\
\cmidrule(lr){2-3}\cmidrule(lr){4-5}\cmidrule(lr){6-7}
Method & L--R & A--B & L--R & A--B & L--R & A--B \\
\midrule
InternVL-3.5-4B (unsteered) & \textbf{100.0}\fstd{0.0} & \textbf{100.0}\fstd{0.0} & \textbf{100.0}\fstd{0.0} & \textbf{100.0}\fstd{0.0} & \textbf{100.0}\fstd{0.0} & \textbf{100.0}\fstd{0.0} \\
\textsc{ScalingVis}~\cite{adaptvis}& \textbf{100.0}\fstd{0.0} & \textbf{100.0}\fstd{0.0} & \textbf{100.0}\fstd{0.0} & \textbf{100.0}\fstd{0.0} & \textbf{100.0}\fstd{0.0} & \textbf{100.0}\fstd{0.0} \\
\textsc{AdaptVis}~\cite{adaptvis}& \textbf{100.0}\fstd{0.0} & \textbf{100.0}\fstd{0.0} & \textbf{100.0}\fstd{0.0} & \textbf{100.0}\fstd{0.0} & \textbf{100.0}\fstd{0.0} & \textbf{100.0}\fstd{0.0} \\
Shift steering~\cite{attention_in_space,ITI} & -- & -- & -- & -- & -- & -- \\
\rowcolor{gray!15}Projective steering & \textbf{100.0}\fstd{0.0} & 99.6\fstd{0.3}\fdown{0.4} & \textbf{100.0}\fstd{0.0} & \textbf{100.0}\fstd{0.0} & \textbf{100.0}\fstd{0.0} & \textbf{100.0}\fstd{0.0} \\
\bottomrule\end{tabular}
}{%
\begin{tabular}{l ccc ccc}\toprule
 & \multicolumn{3}{c}{3D Objects} & \multicolumn{3}{c}{WhatsUp}\\
\cmidrule(lr){2-4}\cmidrule(lr){5-7}
Method & L--R & A--B & F--B & L--R & A--B & F--B \\
\midrule
InternVL-3.5-4B (unsteered) & \textbf{100.0}\fstd{0.0} & \textbf{100.0}\fstd{0.0} & \textbf{100.0}\fstd{0.0} & \textbf{100.0}\fstd{0.0} & 62.3\fstd{1.3} & 95.8\fstd{0.8} \\
\textsc{ScalingVis}~\cite{adaptvis}& \textbf{100.0}\fstd{0.0} & \textbf{100.0}\fstd{0.0} & \textbf{100.0}\fstd{0.0} & \textbf{100.0}\fstd{0.0} & 63.5\fstd{0.9}\fup{1.2} & 97.3\fstd{1.0}\fup{1.5} \\
\textsc{AdaptVis}~\cite{adaptvis}& \textbf{100.0}\fstd{0.0} & \textbf{100.0}\fstd{0.0} & \textbf{100.0}\fstd{0.0} & \textbf{100.0}\fstd{0.0} & 63.5\fstd{0.9}\fup{1.2} & 97.3\fstd{1.0}\fup{1.5} \\
Shift steering~\cite{attention_in_space,ITI} & -- & -- & -- & -- & 62.0\fstd{1.2}\fdown{0.3} & 95.8\fstd{0.8} \\
\rowcolor{gray!15}Projective steering & \textbf{100.0}\fstd{0.0} & 99.9\fstd{0.1}\fdown{0.1} & \textbf{100.0}\fstd{0.0} & 99.8\fstd{0.0}\fdown{0.2} & \textbf{74.9}\fstd{1.5}\fup{12.6} & \textbf{99.9}\fstd{0.2}\fup{4.1} \\
\bottomrule\end{tabular}
}
\end{table*}

\begin{table*}[tp]\centering
\caption{\textbf{Macro-F1 ($\%$) on the directional relations test split for Ministral-3-3B.} Layout as in Table~\ref{tab:f1-gemma}. --: Shift steering not run on these datasets, because every training sample is answered correctly and the shift has no wrong answers to be fitted on.}
\label{tab:f1-ministral}
\small
\fithalves{%
\begin{tabular}{l cccccc}\toprule
 & \multicolumn{6}{c}{2D Shapes}\\
\cmidrule(lr){2-7}
 & \multicolumn{2}{c}{dots} & \multicolumn{2}{c}{squares} & \multicolumn{2}{c}{triangles} \\
\cmidrule(lr){2-3}\cmidrule(lr){4-5}\cmidrule(lr){6-7}
Method & L--R & A--B & L--R & A--B & L--R & A--B \\
\midrule
Ministral-3-3B (unsteered) & 63.3\fstd{1.2} & 99.3\fstd{0.2} & 65.7\fstd{1.0} & \textbf{100.0}\fstd{0.0} & 75.3\fstd{1.1} & \textbf{100.0}\fstd{0.0} \\
\textsc{ScalingVis}~\cite{adaptvis}& 60.6\fstd{1.0}\fdown{2.7} & 99.4\fstd{0.2}\fup{0.1} & 64.6\fstd{1.1}\fdown{1.1} & \textbf{100.0}\fstd{0.0} & 73.7\fstd{0.8}\fdown{1.6} & \textbf{100.0}\fstd{0.0} \\
\textsc{AdaptVis}~\cite{adaptvis}& 60.4\fstd{1.1}\fdown{2.9} & 99.3\fstd{0.3} & 64.6\fstd{1.1}\fdown{1.1} & \textbf{100.0}\fstd{0.0} & 73.9\fstd{1.0}\fdown{1.4} & \textbf{100.0}\fstd{0.0} \\
Shift steering~\cite{attention_in_space,ITI} & 61.6\fstd{1.4}\fdown{1.7} & 99.2\fstd{0.2}\fdown{0.1} & 65.3\fstd{1.0}\fdown{0.4} & -- & 75.0\fstd{1.2}\fdown{0.3} & -- \\
\rowcolor{gray!15}Projective steering & \textbf{88.5}\fstd{1.1}\fup{25.2} & \textbf{100.0}\fstd{0.0}\fup{0.7} & \textbf{88.1}\fstd{1.1}\fup{22.4} & \textbf{100.0}\fstd{0.0} & \textbf{88.9}\fstd{1.6}\fup{13.6} & \textbf{100.0}\fstd{0.0} \\
\bottomrule\end{tabular}
}{%
\begin{tabular}{l ccc ccc}\toprule
 & \multicolumn{3}{c}{3D Objects} & \multicolumn{3}{c}{WhatsUp}\\
\cmidrule(lr){2-4}\cmidrule(lr){5-7}
Method & L--R & A--B & F--B & L--R & A--B & F--B \\
\midrule
Ministral-3-3B (unsteered) & \textbf{100.0}\fstd{0.0} & \textbf{100.0}\fstd{0.0} & \textbf{100.0}\fstd{0.0} & 84.5\fstd{0.7} & 54.3\fstd{1.5} & 63.4\fstd{3.4} \\
\textsc{ScalingVis}~\cite{adaptvis}& \textbf{100.0}\fstd{0.0} & \textbf{100.0}\fstd{0.0} & \textbf{100.0}\fstd{0.0} & 83.9\fstd{1.0}\fdown{0.6} & 56.5\fstd{1.6}\fup{2.2} & 66.3\fstd{2.7}\fup{2.9} \\
\textsc{AdaptVis}~\cite{adaptvis}& \textbf{100.0}\fstd{0.0} & \textbf{100.0}\fstd{0.0} & \textbf{100.0}\fstd{0.0} & 84.6\fstd{1.0}\fup{0.1} & 57.2\fstd{0.9}\fup{2.9} & 65.9\fstd{2.8}\fup{2.5} \\
Shift steering~\cite{attention_in_space,ITI} & -- & -- & -- & \textbf{85.3}\fstd{0.8}\fup{0.8} & 53.9\fstd{1.3}\fdown{0.4} & 62.8\fstd{3.3}\fdown{0.6} \\
\rowcolor{gray!15}Projective steering & \textbf{100.0}\fstd{0.0} & 97.7\fstd{0.4}\fdown{2.3} & \textbf{100.0}\fstd{0.0} & 84.0\fstd{1.2}\fdown{0.5} & \textbf{62.9}\fstd{1.6}\fup{8.6} & \textbf{87.4}\fstd{1.9}\fup{24.0} \\
\bottomrule\end{tabular}
}
\end{table*}

\begin{table*}[tp]
\centering
\caption{\textbf{F1 ($\%$) of the answer ``yes'' on GQA and VQAv2 for Gemma-3-4B.} The steering parameters are fitted on the directional relation training set of each column. Results are the mean $\pm$ standard deviation over five random seeds. Arrows give the change of the mean against the unsteered row of the same dataset. Best mean per column and dataset marked in bold.}
\label{tab:f1-vqa-gemma}
\small
\fithalves{%
\begin{tabular}{l cccccc}\toprule
 & \multicolumn{6}{c}{2D Shapes}\\
\cmidrule(lr){2-7}
 & \multicolumn{2}{c}{dots} & \multicolumn{2}{c}{squares} & \multicolumn{2}{c}{triangles} \\
\cmidrule(lr){2-3}\cmidrule(lr){4-5}\cmidrule(lr){6-7}
Method & L--R & A--B & L--R & A--B & L--R & A--B \\
\midrule
GQA (unsteered) & \textbf{67.8}\fstd{0.0} & 67.8\fstd{0.0} & \textbf{67.8}\fstd{0.0} & 67.8\fstd{0.0} & 67.8\fstd{0.0} & 67.8\fstd{0.0} \\
\textsc{ScalingVis}~\cite{adaptvis}& 67.7\fstd{0.0}\fdown{0.1} & \textbf{68.2}\fstd{0.0}\fup{0.4} & 67.7\fstd{0.0}\fdown{0.1} & \textbf{68.2}\fstd{0.0}\fup{0.4} & 67.7\fstd{0.0}\fdown{0.1} & \textbf{68.2}\fstd{0.0}\fup{0.4} \\
\textsc{AdaptVis}~\cite{adaptvis}& 67.6\fstd{0.1}\fdown{0.2} & \textbf{68.2}\fstd{0.0}\fup{0.4} & 67.7\fstd{0.0}\fdown{0.1} & \textbf{68.2}\fstd{0.0}\fup{0.4} & 67.7\fstd{0.0}\fdown{0.1} & \textbf{68.2}\fstd{0.0}\fup{0.4} \\
Shift steering~\cite{attention_in_space,ITI} & \textbf{67.8}\fstd{0.0} & 67.9\fstd{0.0}\fup{0.1} & \textbf{67.8}\fstd{0.0} & 67.8\fstd{0.0} & 67.8\fstd{0.0} & 67.8\fstd{0.1} \\
\rowcolor{gray!15}Projective steering & \textbf{67.8}\fstd{0.1} & 66.6\fstd{0.0}\fdown{1.2} & 67.6\fstd{0.0}\fdown{0.2} & 67.1\fstd{0.1}\fdown{0.7} & \textbf{67.9}\fstd{0.1}\fup{0.1} & 67.1\fstd{0.1}\fdown{0.7} \\
\bottomrule\end{tabular}
}{%
\begin{tabular}{l ccc ccc}\toprule
 & \multicolumn{3}{c}{3D Objects} & \multicolumn{3}{c}{WhatsUp}\\
\cmidrule(lr){2-4}\cmidrule(lr){5-7}
Method & L--R & A--B & F--B & L--R & A--B & F--B \\
\midrule
GQA (unsteered) & 67.7\fstd{0.0} & 67.7\fstd{0.0} & 67.7\fstd{0.0} & 67.7\fstd{0.0} & 67.7\fstd{0.0} & \textbf{67.7}\fstd{0.0} \\
\textsc{ScalingVis}~\cite{adaptvis}& 67.4\fstd{0.0}\fdown{0.3} & 61.2\fstd{0.0}\fdown{6.5} & 66.2\fstd{2.8}\fdown{1.5} & 67.6\fstd{0.1}\fdown{0.1} & 67.4\fstd{0.0}\fdown{0.3} & 63.7\fstd{3.4}\fdown{4.0} \\
\textsc{AdaptVis}~\cite{adaptvis}& 67.4\fstd{0.0}\fdown{0.3} & 61.2\fstd{0.0}\fdown{6.5} & 65.3\fstd{1.8}\fdown{2.4} & 67.3\fstd{0.1}\fdown{0.4} & 67.4\fstd{0.0}\fdown{0.3} & 63.7\fstd{3.4}\fdown{4.0} \\
Shift steering~\cite{attention_in_space,ITI} & 67.8\fstd{0.0}\fup{0.1} & \textbf{67.8}\fstd{0.0}\fup{0.1} & \textbf{67.8}\fstd{0.0}\fup{0.1} & \textbf{67.8}\fstd{0.1}\fup{0.1} & \textbf{67.8}\fstd{0.0}\fup{0.1} & \textbf{67.7}\fstd{0.0} \\
\rowcolor{gray!15}Projective steering & \textbf{67.9}\fstd{0.0}\fup{0.2} & 67.3\fstd{0.0}\fdown{0.4} & 67.6\fstd{0.0}\fdown{0.1} & \textbf{67.8}\fstd{0.1}\fup{0.1} & 67.2\fstd{0.1}\fdown{0.5} & \textbf{67.7}\fstd{0.1} \\
\midrule
VQAv2 (unsteered) & 83.4\fstd{0.0} & 83.4\fstd{0.0} & 83.4\fstd{0.0} & 83.4\fstd{0.0} & 83.4\fstd{0.0} & \textbf{83.4}\fstd{0.0} \\
\bottomrule\end{tabular}
}
\end{table*}

\begin{table*}[tp]\centering
\caption{\textbf{F1 ($\%$) of the answer ``yes'' on POPE (random, popular, adversarial) for Gemma-3-4B.} Layout as in Table~\ref{tab:f1-vqa-gemma}.}
\label{tab:f1-pope-gemma}
\small
\fithalves{%
\begin{tabular}{l cccccc}\toprule
 & \multicolumn{6}{c}{2D Shapes}\\
\cmidrule(lr){2-7}
 & \multicolumn{2}{c}{dots} & \multicolumn{2}{c}{squares} & \multicolumn{2}{c}{triangles} \\
\cmidrule(lr){2-3}\cmidrule(lr){4-5}\cmidrule(lr){6-7}
Method & L--R & A--B & L--R & A--B & L--R & A--B \\
\midrule
POPE random (unsteered) & \textbf{88.7}\fstd{0.0} & \textbf{88.7}\fstd{0.0} & \textbf{88.7}\fstd{0.0} & \textbf{88.7}\fstd{0.0} & \textbf{88.7}\fstd{0.0} & \textbf{88.7}\fstd{0.0} \\
\textsc{ScalingVis}~\cite{adaptvis}& 88.2\fstd{0.0}\fdown{0.5} & 84.7\fstd{0.0}\fdown{4.0} & 88.2\fstd{0.0}\fdown{0.5} & 84.7\fstd{0.0}\fdown{4.0} & 88.2\fstd{0.0}\fdown{0.5} & 84.7\fstd{0.0}\fdown{4.0} \\
\textsc{AdaptVis}~\cite{adaptvis}& 88.2\fstd{0.0}\fdown{0.5} & 84.7\fstd{0.0}\fdown{4.0} & 88.2\fstd{0.0}\fdown{0.5} & 84.7\fstd{0.0}\fdown{4.0} & 88.2\fstd{0.0}\fdown{0.5} & 84.7\fstd{0.0}\fdown{4.0} \\
Shift steering~\cite{attention_in_space,ITI} & 88.6\fstd{0.0}\fdown{0.1} & 88.6\fstd{0.0}\fdown{0.1} & 88.6\fstd{0.0}\fdown{0.1} & 88.6\fstd{0.0}\fdown{0.1} & \textbf{88.7}\fstd{0.0} & 88.6\fstd{0.0}\fdown{0.1} \\
\rowcolor{gray!15}Projective steering & 88.5\fstd{0.1}\fdown{0.2} & 87.9\fstd{0.0}\fdown{0.8} & 88.6\fstd{0.1}\fdown{0.1} & 88.0\fstd{0.0}\fdown{0.7} & 88.6\fstd{0.0}\fdown{0.1} & 88.0\fstd{0.0}\fdown{0.7} \\
\bottomrule\end{tabular}
}{%
\begin{tabular}{l ccc ccc}\toprule
 & \multicolumn{3}{c}{3D Objects} & \multicolumn{3}{c}{WhatsUp}\\
\cmidrule(lr){2-4}\cmidrule(lr){5-7}
Method & L--R & A--B & F--B & L--R & A--B & F--B \\
\midrule
POPE random (unsteered) & \textbf{88.7}\fstd{0.0} & \textbf{88.7}\fstd{0.0} & \textbf{88.7}\fstd{0.0} & 88.7\fstd{0.0} & \textbf{88.7}\fstd{0.0} & \textbf{88.7}\fstd{0.0} \\
\textsc{ScalingVis}~\cite{adaptvis}& 88.2\fstd{0.0}\fdown{0.5} & 75.2\fstd{0.0}\fdown{13.5} & 85.5\fstd{5.7}\fdown{3.2} & 88.2\fstd{0.0}\fdown{0.5} & 88.2\fstd{0.0}\fdown{0.5} & 80.4\fstd{7.0}\fdown{8.3} \\
\textsc{AdaptVis}~\cite{adaptvis}& 88.2\fstd{0.0}\fdown{0.5} & 75.2\fstd{0.0}\fdown{13.5} & 83.0\fstd{4.3}\fdown{5.7} & 88.2\fstd{0.0}\fdown{0.5} & 88.2\fstd{0.0}\fdown{0.5} & 80.4\fstd{7.1}\fdown{8.3} \\
Shift steering~\cite{attention_in_space,ITI} & 88.6\fstd{0.0}\fdown{0.1} & 88.6\fstd{0.0}\fdown{0.1} & \textbf{88.7}\fstd{0.0} & \textbf{88.8}\fstd{0.0}\fup{0.1} & \textbf{88.7}\fstd{0.0} & \textbf{88.7}\fstd{0.0} \\
\rowcolor{gray!15}Projective steering & 88.6\fstd{0.0}\fdown{0.1} & 88.3\fstd{0.0}\fdown{0.4} & 88.6\fstd{0.0}\fdown{0.1} & 88.5\fstd{0.0}\fdown{0.2} & 88.5\fstd{0.0}\fdown{0.2} & 88.6\fstd{0.1}\fdown{0.1} \\
\midrule
POPE popular (unsteered) & \textbf{83.8}\fstd{0.0} & \textbf{83.8}\fstd{0.0} & 83.8\fstd{0.0} & 83.8\fstd{0.0} & 83.8\fstd{0.0} & \textbf{83.8}\fstd{0.0} \\
\bottomrule\end{tabular}
}
\end{table*}

\begin{table*}[tp]\centering
\caption{\textbf{F1 ($\%$) of the answer ``yes'' on GQA and VQAv2 for InternVL-3.5-4B.} Layout as in Table~\ref{tab:f1-vqa-gemma}.  --: shift steering not run, because every train stimulus is answered correctly for all five seeds and the shift has no wrong answers to be fitted on in this case.}
\label{tab:f1-vqa-internvl}
\small
\fithalves{%
\begin{tabular}{l cccccc}\toprule
 & \multicolumn{6}{c}{2D Shapes}\\
\cmidrule(lr){2-7}
 & \multicolumn{2}{c}{dots} & \multicolumn{2}{c}{squares} & \multicolumn{2}{c}{triangles} \\
\cmidrule(lr){2-3}\cmidrule(lr){4-5}\cmidrule(lr){6-7}
Method & L--R & A--B & L--R & A--B & L--R & A--B \\
\midrule
GQA (unsteered) & \textbf{74.6}\fstd{0.0} & \textbf{74.6}\fstd{0.0} & \textbf{74.6}\fstd{0.0} & \textbf{74.6}\fstd{0.0} & \textbf{74.6}\fstd{0.0} & \textbf{74.6}\fstd{0.0} \\
\textsc{ScalingVis}~\cite{adaptvis}& 74.0\fstd{0.0}\fdown{0.6} & 74.0\fstd{0.0}\fdown{0.6} & 74.0\fstd{0.0}\fdown{0.6} & 74.0\fstd{0.0}\fdown{0.6} & 74.0\fstd{0.0}\fdown{0.6} & 74.0\fstd{0.0}\fdown{0.6} \\
\textsc{AdaptVis}~\cite{adaptvis}& 74.1\fstd{0.0}\fdown{0.5} & 74.1\fstd{0.0}\fdown{0.5} & 74.1\fstd{0.0}\fdown{0.5} & 74.1\fstd{0.0}\fdown{0.5} & 74.1\fstd{0.0}\fdown{0.5} & 74.1\fstd{0.0}\fdown{0.5} \\
Shift steering~\cite{attention_in_space,ITI} & -- & -- & -- & -- & -- & -- \\
\rowcolor{gray!15}Projective steering & 73.6\fstd{0.3}\fdown{1.0} & 72.2\fstd{0.1}\fdown{2.4} & 73.8\fstd{0.1}\fdown{0.8} & 72.2\fstd{0.2}\fdown{2.4} & 73.7\fstd{0.2}\fdown{0.9} & 73.2\fstd{0.1}\fdown{1.4} \\
\bottomrule\end{tabular}
}{%
\begin{tabular}{l ccc ccc}\toprule
 & \multicolumn{3}{c}{3D Objects} & \multicolumn{3}{c}{WhatsUp}\\
\cmidrule(lr){2-4}\cmidrule(lr){5-7}
Method & L--R & A--B & F--B & L--R & A--B & F--B \\
\midrule
GQA (unsteered) & \textbf{74.2}\fstd{0.0} & \textbf{74.2}\fstd{0.0} & \textbf{74.2}\fstd{0.0} & \textbf{74.3}\fstd{0.0} & \textbf{74.3}\fstd{0.0} & \textbf{74.3}\fstd{0.0} \\
\textsc{ScalingVis}~\cite{adaptvis}& 73.9\fstd{0.0}\fdown{0.3} & 73.9\fstd{0.0}\fdown{0.3} & 73.9\fstd{0.0}\fdown{0.3} & 73.8\fstd{0.0}\fdown{0.5} & 74.0\fstd{0.0}\fdown{0.3} & 73.2\fstd{0.0}\fdown{1.1} \\
\textsc{AdaptVis}~\cite{adaptvis}& 73.8\fstd{0.0}\fdown{0.4} & 73.8\fstd{0.0}\fdown{0.4} & 73.8\fstd{0.0}\fdown{0.4} & 74.0\fstd{0.0}\fdown{0.3} & 74.0\fstd{0.0}\fdown{0.3} & 73.2\fstd{0.0}\fdown{1.1} \\
Shift steering~\cite{attention_in_space,ITI} & -- & -- & -- & -- & \textbf{74.3}\fstd{0.1} & \textbf{74.3}\fstd{0.1} \\
\rowcolor{gray!15}Projective steering & 73.0\fstd{0.1}\fdown{1.2} & 72.6\fstd{0.1}\fdown{1.6} & 72.3\fstd{0.1}\fdown{1.9} & 73.2\fstd{0.2}\fdown{1.1} & 69.9\fstd{0.2}\fdown{4.4} & 73.2\fstd{0.1}\fdown{1.1} \\
\midrule
VQAv2 (unsteered) & \textbf{88.9}\fstd{0.0} & \textbf{88.9}\fstd{0.0} & \textbf{88.9}\fstd{0.0} & \textbf{88.9}\fstd{0.0} & \textbf{88.9}\fstd{0.0} & \textbf{88.9}\fstd{0.0} \\
\bottomrule\end{tabular}
}
\end{table*}

\begin{table*}[tp]\centering
\caption{\textbf{F1 ($\%$) of the answer ``yes'' on POPE (random, popular, adversarial) for InternVL-3.5-4B.} Layout as in Table~\ref{tab:f1-vqa-gemma}. --: shift steering not run, because every train stimulus is answered correctly for all five seeds, so the shift has no wrong answers to be fitted on.}
\label{tab:f1-pope-internvl}
\small
\fithalves{%
\begin{tabular}{l cccccc}\toprule
 & \multicolumn{6}{c}{2D Shapes}\\
\cmidrule(lr){2-7}
 & \multicolumn{2}{c}{dots} & \multicolumn{2}{c}{squares} & \multicolumn{2}{c}{triangles} \\
\cmidrule(lr){2-3}\cmidrule(lr){4-5}\cmidrule(lr){6-7}
Method & L--R & A--B & L--R & A--B & L--R & A--B \\
\midrule
POPE random (unsteered) & 91.8\fstd{0.0} & 91.8\fstd{0.0} & 91.8\fstd{0.0} & 91.8\fstd{0.0} & 91.8\fstd{0.0} & 91.8\fstd{0.0} \\
\textsc{ScalingVis}~\cite{adaptvis}& 91.6\fstd{0.0}\fdown{0.2} & 91.6\fstd{0.0}\fdown{0.2} & 91.6\fstd{0.0}\fdown{0.2} & 91.6\fstd{0.0}\fdown{0.2} & 91.6\fstd{0.0}\fdown{0.2} & 91.6\fstd{0.0}\fdown{0.2} \\
\textsc{AdaptVis}~\cite{adaptvis}& \textbf{92.0}\fstd{0.0}\fup{0.2} & \textbf{92.0}\fstd{0.0}\fup{0.2} & \textbf{92.0}\fstd{0.0}\fup{0.2} & \textbf{92.0}\fstd{0.0}\fup{0.2} & \textbf{92.0}\fstd{0.0}\fup{0.2} & \textbf{92.0}\fstd{0.0}\fup{0.2} \\
Shift steering~\cite{attention_in_space,ITI} & -- & -- & -- & -- & -- & -- \\
\rowcolor{gray!15}Projective steering & 91.1\fstd{0.1}\fdown{0.7} & 91.0\fstd{0.1}\fdown{0.8} & 91.2\fstd{0.1}\fdown{0.6} & 91.0\fstd{0.0}\fdown{0.8} & 91.1\fstd{0.0}\fdown{0.7} & 91.1\fstd{0.1}\fdown{0.7} \\
\bottomrule\end{tabular}
}{%
\begin{tabular}{l ccc ccc}\toprule
 & \multicolumn{3}{c}{3D Objects} & \multicolumn{3}{c}{WhatsUp}\\
\cmidrule(lr){2-4}\cmidrule(lr){5-7}
Method & L--R & A--B & F--B & L--R & A--B & F--B \\
\midrule
POPE random (unsteered) & 91.8\fstd{0.0} & 91.8\fstd{0.0} & 91.8\fstd{0.0} & 91.8\fstd{0.0} & 91.8\fstd{0.0} & \textbf{91.8}\fstd{0.0} \\
\textsc{ScalingVis}~\cite{adaptvis}& 91.6\fstd{0.0}\fdown{0.2} & 91.6\fstd{0.0}\fdown{0.2} & 91.6\fstd{0.0}\fdown{0.2} & 91.6\fstd{0.0}\fdown{0.2} & \textbf{92.0}\fstd{0.0}\fup{0.2} & 91.7\fstd{0.0}\fdown{0.1} \\
\textsc{AdaptVis}~\cite{adaptvis}& \textbf{92.0}\fstd{0.0}\fup{0.2} & \textbf{92.0}\fstd{0.0}\fup{0.2} & \textbf{92.0}\fstd{0.0}\fup{0.2} & \textbf{92.0}\fstd{0.0}\fup{0.2} & \textbf{92.0}\fstd{0.0}\fup{0.2} & 91.7\fstd{0.0}\fdown{0.1} \\
Shift steering~\cite{attention_in_space,ITI} & -- & -- & -- & -- & 91.8\fstd{0.0} & \textbf{91.8}\fstd{0.1} \\
\rowcolor{gray!15}Projective steering & 91.3\fstd{0.1}\fdown{0.5} & 90.8\fstd{0.0}\fdown{1.0} & 91.2\fstd{0.0}\fdown{0.6} & 91.0\fstd{0.0}\fdown{0.8} & 91.5\fstd{0.1}\fdown{0.3} & 91.5\fstd{0.1}\fdown{0.3} \\
\midrule
POPE popular (unsteered) & 87.1\fstd{0.0} & 87.1\fstd{0.0} & 87.1\fstd{0.0} & 87.1\fstd{0.0} & 87.1\fstd{0.0} & 87.1\fstd{0.0} \\
\bottomrule\end{tabular}
}
\end{table*}

\begin{table*}[tp]\centering
\caption{\textbf{F1 of the answer ``yes'' ($\%$) on GQA and VQAv2 for Ministral-3-3B.} Layout as in Table~\ref{tab:f1-vqa-gemma}. --: shift steering not run, because every train stimulus is answered correctly for all five seeds, which means the shift has no wrong answers to be fitted on.}
\label{tab:f1-vqa-ministral}
\small
\fithalves{%
\begin{tabular}{l cccccc}\toprule
 & \multicolumn{6}{c}{2D Shapes}\\
\cmidrule(lr){2-7}
 & \multicolumn{2}{c}{dots} & \multicolumn{2}{c}{squares} & \multicolumn{2}{c}{triangles} \\
\cmidrule(lr){2-3}\cmidrule(lr){4-5}\cmidrule(lr){6-7}
Method & L--R & A--B & L--R & A--B & L--R & A--B \\
\midrule
GQA (unsteered) & 56.5\fstd{0.0} & 56.5\fstd{0.0} & \textbf{56.5}\fstd{0.0} & 56.5\fstd{0.0} & 56.5\fstd{0.0} & 56.5\fstd{0.0} \\
\textsc{ScalingVis}~\cite{adaptvis}& \textbf{57.0}\fstd{0.0}\fup{0.5} & \textbf{57.0}\fstd{0.0}\fup{0.5} & 56.0\fstd{0.0}\fdown{0.5} & \textbf{56.9}\fstd{0.0}\fup{0.4} & 56.2\fstd{0.4}\fdown{0.3} & \textbf{56.9}\fstd{0.0}\fup{0.4} \\
\textsc{AdaptVis}~\cite{adaptvis}& \textbf{57.0}\fstd{0.0}\fup{0.5} & \textbf{57.0}\fstd{0.0}\fup{0.5} & 56.0\fstd{0.0}\fdown{0.5} & 56.0\fstd{0.0}\fdown{0.5} & 56.0\fstd{0.0}\fdown{0.5} & 56.0\fstd{0.0}\fdown{0.5} \\
Shift steering~\cite{attention_in_space,ITI} & 56.4\fstd{0.1}\fdown{0.1} & 56.0\fstd{0.1}\fdown{0.5} & \textbf{56.5}\fstd{0.1} & -- & \textbf{56.7}\fstd{0.1}\fup{0.2} & -- \\
\rowcolor{gray!15}Projective steering & 55.6\fstd{0.3}\fdown{0.9} & 53.3\fstd{0.2}\fdown{3.2} & 54.3\fstd{0.3}\fdown{2.2} & 55.0\fstd{0.1}\fdown{1.5} & 55.0\fstd{0.3}\fdown{1.5} & 53.9\fstd{0.2}\fdown{2.6} \\
\bottomrule\end{tabular}
}{%
\begin{tabular}{l ccc ccc}\toprule
 & \multicolumn{3}{c}{3D Objects} & \multicolumn{3}{c}{WhatsUp}\\
\cmidrule(lr){2-4}\cmidrule(lr){5-7}
Method & L--R & A--B & F--B & L--R & A--B & F--B \\
\midrule
GQA (unsteered) & 56.3\fstd{0.0} & 56.3\fstd{0.0} & 56.3\fstd{0.0} & 56.3\fstd{0.0} & \textbf{56.3}\fstd{0.0} & 56.3\fstd{0.0} \\
\textsc{ScalingVis}~\cite{adaptvis}& \textbf{56.7}\fstd{0.0}\fup{0.4} & \textbf{56.7}\fstd{0.0}\fup{0.4} & \textbf{56.7}\fstd{0.0}\fup{0.4} & 56.4\fstd{0.6}\fup{0.1} & 53.1\fstd{2.0}\fdown{3.2} & 52.2\fstd{0.0}\fdown{4.1} \\
\textsc{AdaptVis}~\cite{adaptvis}& 55.9\fstd{0.0}\fdown{0.4} & 55.9\fstd{0.0}\fdown{0.4} & 55.9\fstd{0.0}\fdown{0.4} & \textbf{56.7}\fstd{0.0}\fup{0.4} & 53.9\fstd{2.3}\fdown{2.4} & 53.1\fstd{2.1}\fdown{3.2} \\
Shift steering~\cite{attention_in_space,ITI} & -- & -- & -- & 56.5\fstd{0.1}\fup{0.2} & 56.0\fstd{0.1}\fdown{0.3} & \textbf{56.4}\fstd{0.1}\fup{0.1} \\
\rowcolor{gray!15}Projective steering & 54.8\fstd{0.1}\fdown{1.5} & 52.5\fstd{0.2}\fdown{3.8} & 52.3\fstd{0.1}\fdown{4.0} & 55.2\fstd{0.3}\fdown{1.1} & 55.6\fstd{1.1}\fdown{0.7} & 56.0\fstd{0.9}\fdown{0.3} \\
\midrule
VQAv2 (unsteered) & \textbf{82.2}\fstd{0.0} & \textbf{82.2}\fstd{0.0} & \textbf{82.2}\fstd{0.0} & \textbf{82.2}\fstd{0.0} & 82.2\fstd{0.0} & \textbf{82.2}\fstd{0.0} \\
\bottomrule\end{tabular}
}
\end{table*}

\begin{table*}[tp]\centering
\caption{\textbf{F1 ($\%$) of the answer ``yes'' on POPE (random, popular, adversarial) for Ministral-3-3B.} Layout as in Table~\ref{tab:f1-vqa-gemma}. --: shift steering not run, because every train stimulus is answered correctly for all five seeds, so the shift has no wrong answers to be fitted on.}
\label{tab:f1-pope-ministral}
\small
\fithalves{%
\begin{tabular}{l cccccc}\toprule
 & \multicolumn{6}{c}{2D Shapes}\\
\cmidrule(lr){2-7}
 & \multicolumn{2}{c}{dots} & \multicolumn{2}{c}{squares} & \multicolumn{2}{c}{triangles} \\
\cmidrule(lr){2-3}\cmidrule(lr){4-5}\cmidrule(lr){6-7}
Method & L--R & A--B & L--R & A--B & L--R & A--B \\
\midrule
POPE random (unsteered) & 81.6\fstd{0.0} & 81.6\fstd{0.0} & 81.6\fstd{0.0} & 81.6\fstd{0.0} & 81.6\fstd{0.0} & 81.6\fstd{0.0} \\
\textsc{ScalingVis}~\cite{adaptvis}& \textbf{81.9}\fstd{0.0}\fup{0.3} & \textbf{81.9}\fstd{0.0}\fup{0.3} & 81.6\fstd{0.0} & \textbf{82.5}\fstd{0.0}\fup{0.9} & 81.6\fstd{0.2} & \textbf{82.5}\fstd{0.0}\fup{0.9} \\
\textsc{AdaptVis}~\cite{adaptvis}& \textbf{81.9}\fstd{0.0}\fup{0.3} & \textbf{81.9}\fstd{0.0}\fup{0.3} & 81.6\fstd{0.0} & 81.6\fstd{0.0} & 81.6\fstd{0.0} & 81.6\fstd{0.0} \\
Shift steering~\cite{attention_in_space,ITI} & 81.7\fstd{0.0}\fup{0.1} & 81.6\fstd{0.0} & \textbf{81.8}\fstd{0.1}\fup{0.2} & -- & \textbf{81.8}\fstd{0.1}\fup{0.2} & -- \\
\rowcolor{gray!15}Projective steering & 81.4\fstd{0.1}\fdown{0.2} & 79.7\fstd{0.1}\fdown{1.9} & 80.7\fstd{0.0}\fdown{0.9} & 80.8\fstd{0.0}\fdown{0.8} & 80.6\fstd{0.1}\fdown{1.0} & 80.4\fstd{0.1}\fdown{1.2} \\
\bottomrule\end{tabular}
}{%
\begin{tabular}{l ccc ccc}\toprule
 & \multicolumn{3}{c}{3D Objects} & \multicolumn{3}{c}{WhatsUp}\\
\cmidrule(lr){2-4}\cmidrule(lr){5-7}
Method & L--R & A--B & F--B & L--R & A--B & F--B \\
\midrule
POPE random (unsteered) & 81.6\fstd{0.0} & 81.6\fstd{0.0} & 81.6\fstd{0.0} & 81.6\fstd{0.0} & 81.6\fstd{0.0} & 81.6\fstd{0.0} \\
\textsc{ScalingVis}~\cite{adaptvis}& \textbf{82.5}\fstd{0.0}\fup{0.9} & \textbf{82.5}\fstd{0.0}\fup{0.9} & \textbf{82.5}\fstd{0.0}\fup{0.9} & 81.7\fstd{0.2}\fup{0.1} & 80.8\fstd{1.0}\fdown{0.8} & 80.3\fstd{0.0}\fdown{1.3} \\
\textsc{AdaptVis}~\cite{adaptvis}& 81.6\fstd{0.0} & 81.6\fstd{0.0} & 81.6\fstd{0.0} & \textbf{82.5}\fstd{0.0}\fup{0.9} & 81.2\fstd{1.1}\fdown{0.4} & 80.7\fstd{0.7}\fdown{0.9} \\
Shift steering~\cite{attention_in_space,ITI} & -- & -- & -- & 81.8\fstd{0.1}\fup{0.2} & 81.6\fstd{0.0} & \textbf{81.7}\fstd{0.0}\fup{0.1} \\
\rowcolor{gray!15}Projective steering & 80.3\fstd{0.0}\fdown{1.3} & 79.7\fstd{0.1}\fdown{1.9} & 80.2\fstd{0.1}\fdown{1.4} & 80.7\fstd{0.1}\fdown{0.9} & \textbf{81.9}\fstd{0.9}\fup{0.3} & 81.5\fstd{0.5}\fdown{0.1} \\
\midrule
POPE popular (unsteered) & 80.6\fstd{0.0} & 80.6\fstd{0.0} & 80.6\fstd{0.0} & 80.6\fstd{0.0} & \textbf{80.6}\fstd{0.0} & \textbf{80.6}\fstd{0.0} \\
\bottomrule\end{tabular}
}
\end{table*}

\clearpage
\begin{table*}[h]
\centering
\caption{\textbf{Accuracy (\%) of the control runs} of projective steering on the directional relation test split for \textbf{Gemma-3-4B.} Mean $\pm$ standard deviation over five seeds. Arrows indicate the change of the mean against the unsteered first row. We mark the best mean per column in bold. $^{\ddag}$Control: the projective steering at the same strength ($\alpha = 2.75$) applied to $1$ randomly drawn head (as many as the circuit has), drawn per seed from the heads outside the circuit.}
\label{tab:acc-control-gemma}
\small
\fithalves{%
\begin{tabular}{l cccccc}\toprule
 & \multicolumn{6}{c}{2D Shapes}\\
\cmidrule(lr){2-7}
 & \multicolumn{2}{c}{dots} & \multicolumn{2}{c}{squares} & \multicolumn{2}{c}{triangles} \\
\cmidrule(lr){2-3}\cmidrule(lr){4-5}\cmidrule(lr){6-7}
Method & L--R & A--B & L--R & A--B & L--R & A--B \\
\midrule
Gemma-3-4B (unsteered) & 87.7\fstd{0.8} & 78.5\fstd{1.0} & \textbf{88.3}\fstd{1.1} & \textbf{63.4}\fstd{0.9} & \textbf{84.5}\fstd{0.5} & \textbf{59.6}\fstd{0.4} \\
Control steering$^{\ddag}$ & \textbf{87.8}\fstd{1.0}\fup{0.1} & \textbf{79.5}\fstd{4.5}\fup{1.0} & 88.1\fstd{1.7}\fdown{0.2} & \textbf{63.4}\fstd{1.0} & 83.9\fstd{2.4}\fdown{0.6} & 59.5\fstd{0.9}\fdown{0.1} \\
\bottomrule\end{tabular}
}{%
\begin{tabular}{l ccc ccc}\toprule
 & \multicolumn{3}{c}{3D Objects} & \multicolumn{3}{c}{WhatsUp}\\
\cmidrule(lr){2-4}\cmidrule(lr){5-7}
Method & L--R & A--B & F--B & L--R & A--B & F--B \\
\midrule
Gemma-3-4B (unsteered) & \textbf{88.1}\fstd{1.1} & \textbf{51.0}\fstd{0.4} & \textbf{50.0}\fstd{0.1} & \textbf{70.5}\fstd{0.7} & 61.3\fstd{1.0} & \textbf{50.4}\fstd{0.2} \\
Control steering$^{\ddag}$ & \textbf{88.1}\fstd{1.1} & \textbf{51.0}\fstd{0.4} & \textbf{50.0}\fstd{0.2} & 70.2\fstd{0.9}\fdown{0.3} & \textbf{62.4}\fstd{3.1}\fup{1.1} & \textbf{50.4}\fstd{0.2} \\
\bottomrule\end{tabular}
}
\end{table*}

\begin{table*}[tp]
\centering
\caption{\textbf{Accuracy (\%) of the control runs} of projective steering on the directional relation test split for \textbf{InternVL-3.5-4B.} Layout as in Table~\ref{tab:acc-control-gemma}. $^{\ddag}$Control: the projective steering at the same strength ($\alpha = 2.75$) applied to $12$ randomly drawn heads (as many as the circuit has), drawn per seed from the heads outside the circuit.}
\label{tab:acc-control-internvl}
\small
\fithalves{%
\begin{tabular}{l cccccc}\toprule
 & \multicolumn{6}{c}{2D Shapes}\\
\cmidrule(lr){2-7}
 & \multicolumn{2}{c}{dots} & \multicolumn{2}{c}{squares} & \multicolumn{2}{c}{triangles} \\
\cmidrule(lr){2-3}\cmidrule(lr){4-5}\cmidrule(lr){6-7}
Method & L--R & A--B & L--R & A--B & L--R & A--B \\
\midrule
InternVL-3.5-4B (unsteered) & \textbf{100.0}\fstd{0.0} & \textbf{100.0}\fstd{0.0} & \textbf{100.0}\fstd{0.0} & \textbf{100.0}\fstd{0.0} & \textbf{100.0}\fstd{0.0} & \textbf{100.0}\fstd{0.0} \\
Control steering$^{\ddag}$ & \textbf{100.0}\fstd{0.0} & \textbf{100.0}\fstd{0.0} & \textbf{100.0}\fstd{0.0} & \textbf{100.0}\fstd{0.0} & \textbf{100.0}\fstd{0.0} & \textbf{100.0}\fstd{0.0} \\
\bottomrule\end{tabular}
}{%
\begin{tabular}{l ccc ccc}\toprule
 & \multicolumn{3}{c}{3D Objects} & \multicolumn{3}{c}{WhatsUp}\\
\cmidrule(lr){2-4}\cmidrule(lr){5-7}
Method & L--R & A--B & F--B & L--R & A--B & F--B \\
\midrule
InternVL-3.5-4B (unsteered) & \textbf{100.0}\fstd{0.0} & \textbf{100.0}\fstd{0.0} & \textbf{100.0}\fstd{0.0} & \textbf{100.0}\fstd{0.0} & \textbf{62.7}\fstd{1.3} & \textbf{95.8}\fstd{0.8} \\
Control steering$^{\ddag}$ & \textbf{100.0}\fstd{0.0} & \textbf{100.0}\fstd{0.0} & \textbf{100.0}\fstd{0.0} & \textbf{100.0}\fstd{0.1} & 62.6\fstd{3.3}\fdown{0.1} & 95.0\fstd{2.9}\fdown{0.8} \\
\bottomrule\end{tabular}
}
\end{table*}

\begin{table*}[tp]\centering
\caption{\textbf{Accuracy (\%) of the control runs} of projective steering on the directional relation test split for \textbf{Ministral-3-3B.} Layout as in Table~\ref{tab:acc-control-gemma}. $^{\ddag}$Control: the projective steering at the same strength ($\alpha = 2.75$) applied to $7$ randomly drawn heads (as many as the circuit has), drawn per seed from the heads outside the circuit.}
\label{tab:acc-control-ministral}
\small
\fithalves{%
\begin{tabular}{l cccccc}\toprule
 & \multicolumn{6}{c}{2D Shapes}\\
\cmidrule(lr){2-7}
 & \multicolumn{2}{c}{dots} & \multicolumn{2}{c}{squares} & \multicolumn{2}{c}{triangles} \\
\cmidrule(lr){2-3}\cmidrule(lr){4-5}\cmidrule(lr){6-7}
Method & L--R & A--B & L--R & A--B & L--R & A--B \\
\midrule
Ministral-3-3B (unsteered) & \textbf{67.3}\fstd{0.8} & \textbf{99.3}\fstd{0.2} & \textbf{69.0}\fstd{0.8} & \textbf{100.0}\fstd{0.0} & \textbf{76.7}\fstd{0.9} & \textbf{100.0}\fstd{0.0} \\
Control steering$^{\ddag}$ & 65.1\fstd{7.0}\fdown{2.2} & 98.2\fstd{3.3}\fdown{1.1} & 64.8\fstd{8.4}\fdown{4.2} & 99.7\fstd{0.7}\fdown{0.3} & 73.5\fstd{5.2}\fdown{3.2} & \textbf{100.0}\fstd{0.0} \\
\bottomrule\end{tabular}
}{%
\begin{tabular}{l ccc ccc}\toprule
 & \multicolumn{3}{c}{3D Objects} & \multicolumn{3}{c}{WhatsUp}\\
\cmidrule(lr){2-4}\cmidrule(lr){5-7}
Method & L--R & A--B & F--B & L--R & A--B & F--B \\
\midrule
Ministral-3-3B (unsteered) & \textbf{100.0}\fstd{0.0} & \textbf{100.0}\fstd{0.0} & \textbf{100.0}\fstd{0.0} & 84.6\fstd{0.7} & 55.2\fstd{1.4} & 67.4\fstd{2.4} \\
Control steering$^{\ddag}$ & \textbf{100.0}\fstd{0.0} & \textbf{100.0}\fstd{0.0} & \textbf{100.0}\fstd{0.0} & \textbf{84.7}\fstd{1.1}\fup{0.1} & \textbf{56.5}\fstd{2.0}\fup{1.3} & \textbf{72.8}\fstd{4.1}\fup{5.4} \\
\bottomrule\end{tabular}
}
\end{table*}

\begin{table*}[tp]\centering
\caption{\textbf{Accuracy (\%) of the control runs} of projective steering \textbf{on the general VQA datasets for Gemma-3-4B.} We fit the steering parameters on the directional relation training set of each column. Results show mean $\pm$ standard deviation over five seeds. Arrows give the change of the mean against the unsteered row of the same dataset. Best mean per column and dataset in bold. $^{\ddag}$Control: the projective steering at the same strength ($\alpha = 2.75$) applied to $1$ randomly drawn head (as many as the circuit has), drawn per seed from the heads outside the circuit.}
\label{tab:acc-control-vqa-gemma}
\small
\fithalves{%
\begin{tabular}{l cccccc}\toprule
 & \multicolumn{6}{c}{2D Shapes}\\
\cmidrule(lr){2-7}
 & \multicolumn{2}{c}{dots} & \multicolumn{2}{c}{squares} & \multicolumn{2}{c}{triangles} \\
\cmidrule(lr){2-3}\cmidrule(lr){4-5}\cmidrule(lr){6-7}
Method & L--R & A--B & L--R & A--B & L--R & A--B \\
\midrule
GQA (unsteered) & \textbf{60.6}\fstd{0.0} & \textbf{60.6}\fstd{0.0} & \textbf{60.6}\fstd{0.0} & \textbf{60.6}\fstd{0.0} & \textbf{60.6}\fstd{0.0} & \textbf{60.6}\fstd{0.0} \\
Control steering$^{\ddag}$ & \textbf{60.6}\fstd{0.1} & \textbf{60.6}\fstd{0.2} & \textbf{60.6}\fstd{0.1} & 60.5\fstd{0.1}\fdown{0.1} & \textbf{60.6}\fstd{0.0} & 60.5\fstd{0.1}\fdown{0.1} \\
\midrule
VQAv2 (unsteered) & \textbf{81.7}\fstd{0.0} & 81.7\fstd{0.0} & \textbf{81.7}\fstd{0.0} & \textbf{81.7}\fstd{0.0} & \textbf{81.7}\fstd{0.0} & \textbf{81.7}\fstd{0.0} \\
Control steering$^{\ddag}$ & \textbf{81.7}\fstd{0.0} & \textbf{81.8}\fstd{0.1}\fup{0.1} & \textbf{81.7}\fstd{0.1} & \textbf{81.7}\fstd{0.1} & \textbf{81.7}\fstd{0.1} & \textbf{81.7}\fstd{0.1} \\
\midrule
POPE random (unsteered) & \textbf{88.9}\fstd{0.0} & \textbf{88.9}\fstd{0.0} & \textbf{88.9}\fstd{0.0} & \textbf{88.9}\fstd{0.0} & \textbf{88.9}\fstd{0.0} & \textbf{88.9}\fstd{0.0} \\
Control steering$^{\ddag}$ & \textbf{88.9}\fstd{0.0} & 88.8\fstd{0.0}\fdown{0.1} & \textbf{88.9}\fstd{0.0} & 88.8\fstd{0.0}\fdown{0.1} & \textbf{88.9}\fstd{0.0} & 88.8\fstd{0.0}\fdown{0.1} \\
\midrule
POPE popular (unsteered) & \textbf{83.2}\fstd{0.0} & 83.2\fstd{0.0} & \textbf{83.2}\fstd{0.0} & \textbf{83.2}\fstd{0.0} & \textbf{83.2}\fstd{0.0} & \textbf{83.2}\fstd{0.0} \\
Control steering$^{\ddag}$ & \textbf{83.2}\fstd{0.0} & \textbf{83.3}\fstd{0.0}\fup{0.1} & \textbf{83.2}\fstd{0.0} & \textbf{83.2}\fstd{0.0} & \textbf{83.2}\fstd{0.0} & \textbf{83.2}\fstd{0.0} \\
\midrule
POPE adversarial (unsteered) & \textbf{81.4}\fstd{0.0} & \textbf{81.4}\fstd{0.0} & \textbf{81.4}\fstd{0.0} & \textbf{81.4}\fstd{0.0} & \textbf{81.4}\fstd{0.0} & \textbf{81.4}\fstd{0.0} \\
Control steering$^{\ddag}$ & \textbf{81.4}\fstd{0.0} & \textbf{81.4}\fstd{0.0} & \textbf{81.4}\fstd{0.0} & \textbf{81.4}\fstd{0.0} & \textbf{81.4}\fstd{0.0} & \textbf{81.4}\fstd{0.0} \\
\bottomrule\end{tabular}
}{%
\begin{tabular}{l ccc ccc}\toprule
 & \multicolumn{3}{c}{3D Objects} & \multicolumn{3}{c}{WhatsUp}\\
\cmidrule(lr){2-4}\cmidrule(lr){5-7}
Method & L--R & A--B & F--B & L--R & A--B & F--B \\
\midrule
GQA (unsteered) & \textbf{60.4}\fstd{0.0} & \textbf{60.4}\fstd{0.0} & \textbf{60.4}\fstd{0.0} & 60.4\fstd{0.0} & \textbf{60.4}\fstd{0.0} & \textbf{60.4}\fstd{0.0} \\
Control steering$^{\ddag}$ & \textbf{60.4}\fstd{0.1} & \textbf{60.4}\fstd{0.1} & 60.3\fstd{0.0}\fdown{0.1} & \textbf{60.5}\fstd{0.3}\fup{0.1} & \textbf{60.4}\fstd{0.1} & 60.3\fstd{0.0}\fdown{0.1} \\
\midrule
VQAv2 (unsteered) & 81.6\fstd{0.0} & \textbf{81.6}\fstd{0.0} & \textbf{81.6}\fstd{0.0} & 81.6\fstd{0.0} & \textbf{81.6}\fstd{0.0} & \textbf{81.6}\fstd{0.0} \\
Control steering$^{\ddag}$ & \textbf{81.7}\fstd{0.1}\fup{0.1} & \textbf{81.6}\fstd{0.1} & \textbf{81.6}\fstd{0.1} & \textbf{81.7}\fstd{0.1}\fup{0.1} & \textbf{81.6}\fstd{0.1} & \textbf{81.6}\fstd{0.1} \\
\midrule
POPE random (unsteered) & \textbf{88.9}\fstd{0.0} & \textbf{88.9}\fstd{0.0} & \textbf{88.9}\fstd{0.0} & \textbf{88.9}\fstd{0.0} & \textbf{88.9}\fstd{0.0} & \textbf{88.9}\fstd{0.0} \\
Control steering$^{\ddag}$ & \textbf{88.9}\fstd{0.0} & \textbf{88.9}\fstd{0.0} & \textbf{88.9}\fstd{0.1} & \textbf{88.9}\fstd{0.0} & \textbf{88.9}\fstd{0.1} & \textbf{88.9}\fstd{0.1} \\
\midrule
POPE popular (unsteered) & 83.2\fstd{0.0} & \textbf{83.2}\fstd{0.0} & \textbf{83.2}\fstd{0.0} & 83.2\fstd{0.0} & 83.2\fstd{0.0} & \textbf{83.2}\fstd{0.0} \\
Control steering$^{\ddag}$ & \textbf{83.3}\fstd{0.1}\fup{0.1} & \textbf{83.2}\fstd{0.0} & \textbf{83.2}\fstd{0.1} & \textbf{83.3}\fstd{0.0}\fup{0.1} & \textbf{83.3}\fstd{0.1}\fup{0.1} & \textbf{83.2}\fstd{0.1} \\
\midrule
POPE adversarial (unsteered) & \textbf{81.4}\fstd{0.0} & \textbf{81.4}\fstd{0.0} & \textbf{81.4}\fstd{0.0} & \textbf{81.4}\fstd{0.0} & \textbf{81.4}\fstd{0.0} & \textbf{81.4}\fstd{0.0} \\
Control steering$^{\ddag}$ & \textbf{81.4}\fstd{0.0} & \textbf{81.4}\fstd{0.0} & \textbf{81.4}\fstd{0.0} & \textbf{81.4}\fstd{0.0} & \textbf{81.4}\fstd{0.0} & \textbf{81.4}\fstd{0.1} \\
\bottomrule\end{tabular}
}
\end{table*}

\begin{table*}[tp]\centering
\caption{\textbf{Accuracy (\%) of the control runs} of projective steering \textbf{on the general VQA datasets for InternVL-3.5-4B.} Layout as in Table~\ref{tab:acc-control-vqa-gemma}. $^{\ddag}$Control: the projective steering at the same strength ($\alpha = 2.75$) applied to $12$ randomly drawn heads (as many as the circuit has), drawn per seed from the heads outside the circuit.}
\label{tab:acc-control-vqa-internvl}
\small
\fithalves{%
\begin{tabular}{l cccccc}\toprule
 & \multicolumn{6}{c}{2D Shapes}\\
\cmidrule(lr){2-7}
 & \multicolumn{2}{c}{dots} & \multicolumn{2}{c}{squares} & \multicolumn{2}{c}{triangles} \\
\cmidrule(lr){2-3}\cmidrule(lr){4-5}\cmidrule(lr){6-7}
Method & L--R & A--B & L--R & A--B & L--R & A--B \\
\midrule
GQA (unsteered) & \textbf{74.6}\fstd{0.0} & \textbf{74.6}\fstd{0.0} & \textbf{74.6}\fstd{0.0} & \textbf{74.6}\fstd{0.0} & \textbf{74.6}\fstd{0.0} & \textbf{74.6}\fstd{0.0} \\
Control steering$^{\ddag}$ & 74.1\fstd{0.3}\fdown{0.5} & 74.1\fstd{0.3}\fdown{0.5} & 74.1\fstd{0.2}\fdown{0.5} & 74.2\fstd{0.2}\fdown{0.4} & 74.2\fstd{0.2}\fdown{0.4} & 74.2\fstd{0.4}\fdown{0.4} \\
\midrule
VQAv2 (unsteered) & 88.7\fstd{0.0} & \textbf{88.7}\fstd{0.0} & \textbf{88.7}\fstd{0.0} & \textbf{88.7}\fstd{0.0} & \textbf{88.7}\fstd{0.0} & \textbf{88.7}\fstd{0.0} \\
Control steering$^{\ddag}$ & \textbf{88.9}\fstd{0.1}\fup{0.2} & \textbf{88.7}\fstd{0.2} & \textbf{88.7}\fstd{0.2} & 88.6\fstd{0.1}\fdown{0.1} & \textbf{88.7}\fstd{0.1} & \textbf{88.7}\fstd{0.2} \\
\midrule
POPE random (unsteered) & \textbf{91.9}\fstd{0.0} & 91.9\fstd{0.0} & \textbf{91.9}\fstd{0.0} & \textbf{91.9}\fstd{0.0} & 91.9\fstd{0.0} & \textbf{91.9}\fstd{0.0} \\
Control steering$^{\ddag}$ & \textbf{91.9}\fstd{0.1} & \textbf{92.0}\fstd{0.1}\fup{0.1} & \textbf{91.9}\fstd{0.1} & \textbf{91.9}\fstd{0.1} & \textbf{92.0}\fstd{0.2}\fup{0.1} & \textbf{91.9}\fstd{0.1} \\
\midrule
POPE popular (unsteered) & \textbf{86.5}\fstd{0.0} & \textbf{86.5}\fstd{0.0} & \textbf{86.5}\fstd{0.0} & \textbf{86.5}\fstd{0.0} & \textbf{86.5}\fstd{0.0} & \textbf{86.5}\fstd{0.0} \\
Control steering$^{\ddag}$ & 86.4\fstd{0.2}\fdown{0.1} & \textbf{86.5}\fstd{0.1} & 86.4\fstd{0.1}\fdown{0.1} & 86.4\fstd{0.1}\fdown{0.1} & \textbf{86.5}\fstd{0.2} & 86.4\fstd{0.1}\fdown{0.1} \\
\midrule
POPE adversarial (unsteered) & \textbf{82.1}\fstd{0.0} & \textbf{82.1}\fstd{0.0} & \textbf{82.1}\fstd{0.0} & \textbf{82.1}\fstd{0.0} & \textbf{82.1}\fstd{0.0} & \textbf{82.1}\fstd{0.0} \\
Control steering$^{\ddag}$ & 81.9\fstd{0.2}\fdown{0.2} & \textbf{82.1}\fstd{0.2} & 81.9\fstd{0.2}\fdown{0.2} & 82.0\fstd{0.2}\fdown{0.1} & 82.0\fstd{0.3}\fdown{0.1} & 82.0\fstd{0.1}\fdown{0.1} \\
\bottomrule\end{tabular}
}{%
\begin{tabular}{l ccc ccc}\toprule
 & \multicolumn{3}{c}{3D Objects} & \multicolumn{3}{c}{WhatsUp}\\
\cmidrule(lr){2-4}\cmidrule(lr){5-7}
Method & L--R & A--B & F--B & L--R & A--B & F--B \\
\midrule
GQA (unsteered) & \textbf{74.3}\fstd{0.0} & \textbf{74.3}\fstd{0.0} & \textbf{74.3}\fstd{0.0} & \textbf{74.3}\fstd{0.0} & \textbf{74.3}\fstd{0.0} & \textbf{74.3}\fstd{0.0} \\
Control steering$^{\ddag}$ & \textbf{74.3}\fstd{0.2} & 74.1\fstd{0.3}\fdown{0.2} & 74.1\fstd{0.3}\fdown{0.2} & \textbf{74.3}\fstd{0.3} & 74.2\fstd{0.1}\fdown{0.1} & 74.2\fstd{0.2}\fdown{0.1} \\
\midrule
VQAv2 (unsteered) & \textbf{88.7}\fstd{0.0} & \textbf{88.7}\fstd{0.0} & \textbf{88.7}\fstd{0.0} & \textbf{88.7}\fstd{0.0} & 88.7\fstd{0.0} & \textbf{88.7}\fstd{0.0} \\
Control steering$^{\ddag}$ & 88.6\fstd{0.1}\fdown{0.1} & \textbf{88.7}\fstd{0.2} & \textbf{88.7}\fstd{0.2} & \textbf{88.7}\fstd{0.2} & \textbf{88.8}\fstd{0.2}\fup{0.1} & \textbf{88.7}\fstd{0.2} \\
\midrule
POPE random (unsteered) & \textbf{91.9}\fstd{0.0} & \textbf{91.9}\fstd{0.0} & \textbf{91.9}\fstd{0.0} & \textbf{91.9}\fstd{0.0} & \textbf{91.9}\fstd{0.0} & \textbf{91.9}\fstd{0.0} \\
Control steering$^{\ddag}$ & \textbf{91.9}\fstd{0.1} & \textbf{91.9}\fstd{0.1} & \textbf{91.9}\fstd{0.1} & \textbf{91.9}\fstd{0.1} & \textbf{91.9}\fstd{0.1} & \textbf{91.9}\fstd{0.1} \\
\midrule
POPE popular (unsteered) & \textbf{86.5}\fstd{0.0} & \textbf{86.5}\fstd{0.0} & 86.5\fstd{0.0} & \textbf{86.5}\fstd{0.0} & \textbf{86.5}\fstd{0.0} & \textbf{86.5}\fstd{0.0} \\
Control steering$^{\ddag}$ & \textbf{86.5}\fstd{0.1} & 86.4\fstd{0.1}\fdown{0.1} & \textbf{86.6}\fstd{0.1}\fup{0.1} & 86.4\fstd{0.1}\fdown{0.1} & 86.3\fstd{0.3}\fdown{0.2} & \textbf{86.5}\fstd{0.2} \\
\midrule
POPE adversarial (unsteered) & \textbf{82.1}\fstd{0.0} & \textbf{82.1}\fstd{0.0} & 82.1\fstd{0.0} & \textbf{82.1}\fstd{0.0} & \textbf{82.1}\fstd{0.0} & 82.1\fstd{0.0} \\
Control steering$^{\ddag}$ & 82.0\fstd{0.2}\fdown{0.1} & \textbf{82.1}\fstd{0.1} & \textbf{82.2}\fstd{0.1}\fup{0.1} & 82.0\fstd{0.1}\fdown{0.1} & \textbf{82.1}\fstd{0.2} & \textbf{82.2}\fstd{0.3}\fup{0.1} \\
\bottomrule\end{tabular}
}
\end{table*}

\begin{table*}[tp]\centering
\caption{\textbf{Accuracy (\%) of the control runs} of projective steering \textbf{on the general VQA datasets for Ministral-3-3B.} Layout as in Table~\ref{tab:acc-control-vqa-gemma}. $^{\ddag}$Control: the projective steering at the same strength ($\alpha = 2.75$) applied to $7$ randomly drawn heads (as many as the circuit has), drawn per seed from the heads outside the circuit.}
\label{tab:acc-control-vqa-ministral}
\small
\fithalves{%
\begin{tabular}{l cccccc}\toprule
 & \multicolumn{6}{c}{2D Shapes}\\
\cmidrule(lr){2-7}
 & \multicolumn{2}{c}{dots} & \multicolumn{2}{c}{squares} & \multicolumn{2}{c}{triangles} \\
\cmidrule(lr){2-3}\cmidrule(lr){4-5}\cmidrule(lr){6-7}
Method & L--R & A--B & L--R & A--B & L--R & A--B \\
\midrule
GQA (unsteered) & 64.2\fstd{0.0} & 64.2\fstd{0.0} & 64.2\fstd{0.0} & 64.2\fstd{0.0} & 64.2\fstd{0.0} & 64.2\fstd{0.0} \\
Control steering$^{\ddag}$ & \textbf{64.3}\fstd{0.2}\fup{0.1} & \textbf{64.3}\fstd{0.5}\fup{0.1} & \textbf{64.5}\fstd{0.3}\fup{0.3} & \textbf{64.3}\fstd{0.5}\fup{0.1} & \textbf{64.4}\fstd{0.2}\fup{0.2} & \textbf{64.3}\fstd{0.4}\fup{0.1} \\
\midrule
VQAv2 (unsteered) & \textbf{83.9}\fstd{0.0} & \textbf{83.9}\fstd{0.0} & \textbf{83.9}\fstd{0.0} & \textbf{83.9}\fstd{0.0} & \textbf{83.9}\fstd{0.0} & \textbf{83.9}\fstd{0.0} \\
Control steering$^{\ddag}$ & 83.8\fstd{0.1}\fdown{0.1} & 83.7\fstd{0.3}\fdown{0.2} & 83.8\fstd{0.1}\fdown{0.1} & 83.7\fstd{0.3}\fdown{0.2} & \textbf{83.9}\fstd{0.3} & 83.8\fstd{0.2}\fdown{0.1} \\
\midrule
POPE random (unsteered) & 84.2\fstd{0.0} & 84.2\fstd{0.0} & 84.2\fstd{0.0} & 84.2\fstd{0.0} & 84.2\fstd{0.0} & 84.2\fstd{0.0} \\
Control steering$^{\ddag}$ & \textbf{84.4}\fstd{0.1}\fup{0.2} & \textbf{84.5}\fstd{0.4}\fup{0.3} & \textbf{84.5}\fstd{0.2}\fup{0.3} & \textbf{84.4}\fstd{0.4}\fup{0.2} & \textbf{84.6}\fstd{0.2}\fup{0.4} & \textbf{84.5}\fstd{0.4}\fup{0.3} \\
\midrule
POPE popular (unsteered) & 83.1\fstd{0.0} & 83.1\fstd{0.0} & 83.1\fstd{0.0} & 83.1\fstd{0.0} & 83.1\fstd{0.0} & 83.1\fstd{0.0} \\
Control steering$^{\ddag}$ & \textbf{83.2}\fstd{0.1}\fup{0.1} & \textbf{83.2}\fstd{0.2}\fup{0.1} & \textbf{83.3}\fstd{0.1}\fup{0.2} & \textbf{83.2}\fstd{0.3}\fup{0.1} & \textbf{83.4}\fstd{0.2}\fup{0.3} & \textbf{83.2}\fstd{0.3}\fup{0.1} \\
\midrule
POPE adversarial (unsteered) & 81.8\fstd{0.0} & 81.8\fstd{0.0} & 81.8\fstd{0.0} & 81.8\fstd{0.0} & 81.8\fstd{0.0} & 81.8\fstd{0.0} \\
Control steering$^{\ddag}$ & \textbf{81.9}\fstd{0.1}\fup{0.1} & \textbf{81.9}\fstd{0.2}\fup{0.1} & \textbf{82.0}\fstd{0.1}\fup{0.2} & \textbf{81.9}\fstd{0.2}\fup{0.1} & \textbf{82.0}\fstd{0.2}\fup{0.2} & \textbf{82.0}\fstd{0.2}\fup{0.2} \\
\bottomrule\end{tabular}
}{%
\begin{tabular}{l ccc ccc}\toprule
 & \multicolumn{3}{c}{3D Objects} & \multicolumn{3}{c}{WhatsUp}\\
\cmidrule(lr){2-4}\cmidrule(lr){5-7}
Method & L--R & A--B & F--B & L--R & A--B & F--B \\
\midrule
GQA (unsteered) & \textbf{64.0}\fstd{0.0} & \textbf{64.0}\fstd{0.0} & 64.0\fstd{0.0} & \textbf{64.0}\fstd{0.0} & 64.0\fstd{0.0} & \textbf{64.0}\fstd{0.0} \\
Control steering$^{\ddag}$ & \textbf{64.0}\fstd{0.5} & \textbf{64.0}\fstd{0.7} & \textbf{64.1}\fstd{0.1}\fup{0.1} & \textbf{64.0}\fstd{0.5} & \textbf{64.4}\fstd{0.4}\fup{0.4} & \textbf{64.0}\fstd{0.3} \\
\midrule
VQAv2 (unsteered) & \textbf{83.9}\fstd{0.0} & \textbf{83.9}\fstd{0.0} & \textbf{83.9}\fstd{0.0} & \textbf{83.9}\fstd{0.0} & \textbf{83.9}\fstd{0.0} & \textbf{83.9}\fstd{0.0} \\
Control steering$^{\ddag}$ & 83.6\fstd{0.2}\fdown{0.3} & 83.5\fstd{0.7}\fdown{0.4} & 83.7\fstd{0.1}\fdown{0.2} & 83.5\fstd{0.2}\fdown{0.4} & 83.7\fstd{0.3}\fdown{0.2} & 83.7\fstd{0.1}\fdown{0.2} \\
\midrule
POPE random (unsteered) & \textbf{84.2}\fstd{0.0} & 84.2\fstd{0.0} & 84.2\fstd{0.0} & \textbf{84.2}\fstd{0.0} & 84.2\fstd{0.0} & 84.2\fstd{0.0} \\
Control steering$^{\ddag}$ & \textbf{84.2}\fstd{0.2} & \textbf{84.4}\fstd{0.5}\fup{0.2} & \textbf{84.4}\fstd{0.1}\fup{0.2} & \textbf{84.2}\fstd{0.1} & \textbf{84.4}\fstd{0.5}\fup{0.2} & \textbf{84.3}\fstd{0.1}\fup{0.1} \\
\midrule
POPE popular (unsteered) & \textbf{83.1}\fstd{0.0} & \textbf{83.1}\fstd{0.0} & 83.1\fstd{0.0} & \textbf{83.1}\fstd{0.0} & 83.1\fstd{0.0} & \textbf{83.1}\fstd{0.0} \\
Control steering$^{\ddag}$ & 83.0\fstd{0.1}\fdown{0.1} & \textbf{83.1}\fstd{0.3} & \textbf{83.2}\fstd{0.1}\fup{0.1} & \textbf{83.1}\fstd{0.1} & \textbf{83.2}\fstd{0.4}\fup{0.1} & \textbf{83.1}\fstd{0.1} \\
\midrule
POPE adversarial (unsteered) & \textbf{81.8}\fstd{0.0} & 81.8\fstd{0.0} & 81.8\fstd{0.0} & \textbf{81.8}\fstd{0.0} & 81.8\fstd{0.0} & 81.8\fstd{0.0} \\
Control steering$^{\ddag}$ & \textbf{81.8}\fstd{0.1} & \textbf{81.9}\fstd{0.3}\fup{0.1} & \textbf{81.9}\fstd{0.1}\fup{0.1} & \textbf{81.8}\fstd{0.1} & \textbf{82.0}\fstd{0.3}\fup{0.2} & \textbf{81.9}\fstd{0.1}\fup{0.1} \\
\bottomrule\end{tabular}
}
\end{table*}

\begin{table*}[tp]\centering
\caption{\textbf{Macro-F1 ($\%$) of the control runs} of projective steering on the directional relation test split for \textbf{Gemma3-4B.} Mean $\pm$ standard deviation over five seeds. Arrows give the change of the mean against the unsteered first row. Best mean per column in bold. $^{\ddag}$Control: the projective steering rule at the same strength ($\alpha = 2.75$) applied to $1$ randomly drawn head (as many as the circuit has), drawn per seed from the heads outside the circuit.}
\label{tab:f1-control-gemma}
\small
\fithalves{%
\begin{tabular}{l cccccc}\toprule
 & \multicolumn{6}{c}{2D Shapes}\\
\cmidrule(lr){2-7}
 & \multicolumn{2}{c}{dots} & \multicolumn{2}{c}{squares} & \multicolumn{2}{c}{triangles} \\
\cmidrule(lr){2-3}\cmidrule(lr){4-5}\cmidrule(lr){6-7}
Method & L--R & A--B & L--R & A--B & L--R & A--B \\
\midrule
Gemma-3-4B (unsteered) & 87.7\fstd{0.8} & 77.4\fstd{1.1} & \textbf{88.1}\fstd{1.1} & \textbf{57.8}\fstd{1.3} & \textbf{84.2}\fstd{0.6} & \textbf{51.7}\fstd{0.7} \\
Control steering$^{\ddag}$ & \textbf{87.8}\fstd{1.0}\fup{0.1} & \textbf{78.5}\fstd{5.1}\fup{1.1} & 87.9\fstd{1.8}\fdown{0.2} & \textbf{57.8}\fstd{1.5} & 83.5\fstd{2.6}\fdown{0.7} & 51.5\fstd{1.4}\fdown{0.2} \\
\bottomrule\end{tabular}
}{%
\begin{tabular}{l ccc ccc}\toprule
 & \multicolumn{3}{c}{3D Objects} & \multicolumn{3}{c}{WhatsUp}\\
\cmidrule(lr){2-4}\cmidrule(lr){5-7}
Method & L--R & A--B & F--B & L--R & A--B & F--B \\
\midrule
Gemma-3-4B (unsteered) & \textbf{87.9}\fstd{1.1} & 35.4\fstd{0.8} & \textbf{33.8}\fstd{0.2} & \textbf{68.3}\fstd{0.8} & 54.7\fstd{1.7} & \textbf{34.2}\fstd{0.5} \\
Control steering$^{\ddag}$ & \textbf{87.9}\fstd{1.2} & \textbf{35.5}\fstd{0.9}\fup{0.1} & 33.7\fstd{0.2}\fdown{0.1} & 67.9\fstd{1.2}\fdown{0.4} & \textbf{56.3}\fstd{4.8}\fup{1.6} & \textbf{34.2}\fstd{0.5} \\
\bottomrule\end{tabular}
}
\end{table*}

\begin{table*}[tp]\centering
\caption{\textbf{Macro-F1 ($\%$) of the control runs} of projective steering on the directional relation test split for \textbf{InternVL3.5-4B.} Layout as in Table~\ref{tab:f1-control-gemma}. $^{\ddag}$Control: the projective steering rule at the same strength ($\alpha = 2.75$) applied to $12$ randomly drawn heads (as many as the circuit has), drawn per seed from the heads outside the circuit.}
\label{tab:f1-control-internvl}
\small
\fithalves{%
\begin{tabular}{l cccccc}\toprule
 & \multicolumn{6}{c}{2D Shapes}\\
\cmidrule(lr){2-7}
 & \multicolumn{2}{c}{dots} & \multicolumn{2}{c}{squares} & \multicolumn{2}{c}{triangles} \\
\cmidrule(lr){2-3}\cmidrule(lr){4-5}\cmidrule(lr){6-7}
Method & L--R & A--B & L--R & A--B & L--R & A--B \\
\midrule
InternVL-3.5-4B (unsteered) & \textbf{100.0}\fstd{0.0} & \textbf{100.0}\fstd{0.0} & \textbf{100.0}\fstd{0.0} & \textbf{100.0}\fstd{0.0} & \textbf{100.0}\fstd{0.0} & \textbf{100.0}\fstd{0.0} \\
Control steering$^{\ddag}$ & \textbf{100.0}\fstd{0.0} & \textbf{100.0}\fstd{0.0} & \textbf{100.0}\fstd{0.0} & \textbf{100.0}\fstd{0.0} & \textbf{100.0}\fstd{0.0} & \textbf{100.0}\fstd{0.0} \\
\bottomrule\end{tabular}
}{%
\begin{tabular}{l ccc ccc}\toprule
 & \multicolumn{3}{c}{3D Objects} & \multicolumn{3}{c}{WhatsUp}\\
\cmidrule(lr){2-4}\cmidrule(lr){5-7}
Method & L--R & A--B & F--B & L--R & A--B & F--B \\
\midrule
InternVL-3.5-4B (unsteered) & \textbf{100.0}\fstd{0.0} & \textbf{100.0}\fstd{0.0} & \textbf{100.0}\fstd{0.0} & \textbf{100.0}\fstd{0.0} & \textbf{62.3}\fstd{1.3} & \textbf{95.8}\fstd{0.8} \\
Control steering$^{\ddag}$ & \textbf{100.0}\fstd{0.0} & \textbf{100.0}\fstd{0.0} & \textbf{100.0}\fstd{0.0} & \textbf{100.0}\fstd{0.1} & 62.2\fstd{3.1}\fdown{0.1} & 95.0\fstd{2.9}\fdown{0.8} \\
\bottomrule\end{tabular}
}
\end{table*}

\begin{table*}[tp]\centering
\caption{\textbf{Macro-F1 ($\%$) of the control runs }of projective steering on the directional relation test split for \textbf{Ministral3-3B.} Layout as in Table~\ref{tab:f1-control-gemma}. $^{\ddag}$Control: the projective steering at the same strength ($\alpha = 2.75$) applied to 7 randomly drawn heads (as many as the circuit has), drawn per seed from the heads outside the circuit.}
\label{tab:f1-control-ministral}
\small
\fithalves{%
\begin{tabular}{l cccccc}\toprule
 & \multicolumn{6}{c}{2D Shapes}\\
\cmidrule(lr){2-7}
 & \multicolumn{2}{c}{dots} & \multicolumn{2}{c}{squares} & \multicolumn{2}{c}{triangles} \\
\cmidrule(lr){2-3}\cmidrule(lr){4-5}\cmidrule(lr){6-7}
Method & L--R & A--B & L--R & A--B & L--R & A--B \\
\midrule
Ministral-3-3B (unsteered) & \textbf{63.3}\fstd{1.2} & \textbf{99.3}\fstd{0.2} & \textbf{65.7}\fstd{1.0} & \textbf{100.0}\fstd{0.0} & \textbf{75.3}\fstd{1.1} & \textbf{100.0}\fstd{0.0} \\
Control steering$^{\ddag}$ & 59.6\fstd{10.3}\fdown{3.7} & 98.2\fstd{3.3}\fdown{1.1} & 58.9\fstd{12.7}\fdown{6.8} & 99.7\fstd{0.7}\fdown{0.3} & 71.3\fstd{6.6}\fdown{4.0} & \textbf{100.0}\fstd{0.0} \\
\bottomrule\end{tabular}
}{%
\begin{tabular}{l ccc ccc}\toprule
 & \multicolumn{3}{c}{3D Objects} & \multicolumn{3}{c}{WhatsUp}\\
\cmidrule(lr){2-4}\cmidrule(lr){5-7}
Method & L--R & A--B & F--B & L--R & A--B & F--B \\
\midrule
Ministral-3-3B (unsteered) & \textbf{100.0}\fstd{0.0} & \textbf{100.0}\fstd{0.0} & \textbf{100.0}\fstd{0.0} & 84.5\fstd{0.7} & 54.3\fstd{1.5} & 63.4\fstd{3.4} \\
Control steering$^{\ddag}$ & \textbf{100.0}\fstd{0.0} & \textbf{100.0}\fstd{0.0} & \textbf{100.0}\fstd{0.0} & \textbf{84.6}\fstd{1.1}\fup{0.1} & \textbf{56.2}\fstd{2.2}\fup{1.9} & \textbf{70.8}\fstd{5.4}\fup{7.4} \\
\bottomrule\end{tabular}
}
\end{table*}

\begin{table*}[tp]\centering
\caption{\textbf{F1 ($\%$) of the answer ``yes'' of the control runs of project steering on the general VQA datasets for Gemma-3-4B.} Steering parameters fitted on the directional relation training set of each column. Mean $\pm$ standard deviation over five random seeds. Arrows give the change of the mean against the unsteered row of the same dataset. Best mean per column and dataset in bold. $^{\ddag}$Control: the projective steering at the same strength ($\alpha = 2.75$) applied to 1 randomly drawn head (as many as the circuit has), drawn per seed from the heads outside the circuit.}
\label{tab:f1-control-vqa-gemma}
\small
\fithalves{%
\begin{tabular}{l cccccc}\toprule
 & \multicolumn{6}{c}{2D Shapes}\\
\cmidrule(lr){2-7}
 & \multicolumn{2}{c}{dots} & \multicolumn{2}{c}{squares} & \multicolumn{2}{c}{triangles} \\
\cmidrule(lr){2-3}\cmidrule(lr){4-5}\cmidrule(lr){6-7}
Method & L--R & A--B & L--R & A--B & L--R & A--B \\
\midrule
GQA (unsteered) & \textbf{67.8}\fstd{0.0} & \textbf{67.8}\fstd{0.0} & \textbf{67.8}\fstd{0.0} & \textbf{67.8}\fstd{0.0} & \textbf{67.8}\fstd{0.0} & \textbf{67.8}\fstd{0.0} \\
Control steering$^{\ddag}$ & \textbf{67.8}\fstd{0.1} & \textbf{67.8}\fstd{0.1} & \textbf{67.8}\fstd{0.1} & 67.7\fstd{0.1}\fdown{0.1} & \textbf{67.8}\fstd{0.1} & \textbf{67.8}\fstd{0.1} \\
\midrule
VQAv2 (unsteered) & \textbf{83.5}\fstd{0.0} & 83.5\fstd{0.0} & \textbf{83.5}\fstd{0.0} & \textbf{83.5}\fstd{0.0} & \textbf{83.5}\fstd{0.0} & \textbf{83.5}\fstd{0.0} \\
Control steering$^{\ddag}$ & \textbf{83.5}\fstd{0.0} & \textbf{83.6}\fstd{0.1}\fup{0.1} & \textbf{83.5}\fstd{0.1} & \textbf{83.5}\fstd{0.0} & \textbf{83.5}\fstd{0.1} & \textbf{83.5}\fstd{0.1} \\
\midrule
POPE random (unsteered) & \textbf{88.7}\fstd{0.0} & \textbf{88.7}\fstd{0.0} & \textbf{88.7}\fstd{0.0} & \textbf{88.7}\fstd{0.0} & \textbf{88.7}\fstd{0.0} & \textbf{88.7}\fstd{0.0} \\
Control steering$^{\ddag}$ & 88.6\fstd{0.0}\fdown{0.1} & 88.6\fstd{0.0}\fdown{0.1} & 88.6\fstd{0.0}\fdown{0.1} & 88.6\fstd{0.0}\fdown{0.1} & 88.6\fstd{0.0}\fdown{0.1} & 88.6\fstd{0.0}\fdown{0.1} \\
\midrule
POPE popular (unsteered) & \textbf{83.8}\fstd{0.0} & \textbf{83.8}\fstd{0.0} & \textbf{83.8}\fstd{0.0} & \textbf{83.8}\fstd{0.0} & \textbf{83.8}\fstd{0.0} & \textbf{83.8}\fstd{0.0} \\
Control steering$^{\ddag}$ & \textbf{83.8}\fstd{0.0} & \textbf{83.8}\fstd{0.0} & \textbf{83.8}\fstd{0.0} & \textbf{83.8}\fstd{0.0} & \textbf{83.8}\fstd{0.0} & \textbf{83.8}\fstd{0.0} \\
\midrule
POPE adversarial (unsteered) & \textbf{82.4}\fstd{0.0} & \textbf{82.4}\fstd{0.0} & \textbf{82.4}\fstd{0.0} & \textbf{82.4}\fstd{0.0} & \textbf{82.4}\fstd{0.0} & \textbf{82.4}\fstd{0.0} \\
Control steering$^{\ddag}$ & 82.3\fstd{0.0}\fdown{0.1} & 82.3\fstd{0.0}\fdown{0.1} & 82.3\fstd{0.0}\fdown{0.1} & 82.3\fstd{0.0}\fdown{0.1} & 82.3\fstd{0.0}\fdown{0.1} & 82.3\fstd{0.0}\fdown{0.1} \\
\bottomrule\end{tabular}
}{%
\begin{tabular}{l ccc ccc}\toprule
 & \multicolumn{3}{c}{3D Objects} & \multicolumn{3}{c}{WhatsUp}\\
\cmidrule(lr){2-4}\cmidrule(lr){5-7}
Method & L--R & A--B & F--B & L--R & A--B & F--B \\
\midrule
GQA (unsteered) & \textbf{67.7}\fstd{0.0} & \textbf{67.7}\fstd{0.0} & \textbf{67.7}\fstd{0.0} & \textbf{67.7}\fstd{0.0} & \textbf{67.7}\fstd{0.0} & \textbf{67.7}\fstd{0.0} \\
Control steering$^{\ddag}$ & \textbf{67.7}\fstd{0.1} & \textbf{67.7}\fstd{0.1} & \textbf{67.7}\fstd{0.0} & \textbf{67.7}\fstd{0.1} & \textbf{67.7}\fstd{0.1} & 67.6\fstd{0.0}\fdown{0.1} \\
\midrule
VQAv2 (unsteered) & 83.4\fstd{0.0} & \textbf{83.4}\fstd{0.0} & \textbf{83.4}\fstd{0.0} & 83.4\fstd{0.0} & \textbf{83.4}\fstd{0.0} & \textbf{83.4}\fstd{0.0} \\
Control steering$^{\ddag}$ & \textbf{83.5}\fstd{0.1}\fup{0.1} & \textbf{83.4}\fstd{0.1} & \textbf{83.4}\fstd{0.1} & \textbf{83.5}\fstd{0.0}\fup{0.1} & \textbf{83.4}\fstd{0.1} & \textbf{83.4}\fstd{0.0} \\
\midrule
POPE random (unsteered) & \textbf{88.7}\fstd{0.0} & \textbf{88.7}\fstd{0.0} & \textbf{88.7}\fstd{0.0} & \textbf{88.7}\fstd{0.0} & \textbf{88.7}\fstd{0.0} & \textbf{88.7}\fstd{0.0} \\
Control steering$^{\ddag}$ & \textbf{88.7}\fstd{0.0} & \textbf{88.7}\fstd{0.0} & \textbf{88.7}\fstd{0.1} & \textbf{88.7}\fstd{0.0} & \textbf{88.7}\fstd{0.1} & \textbf{88.7}\fstd{0.1} \\
\midrule
POPE popular (unsteered) & 83.8\fstd{0.0} & \textbf{83.8}\fstd{0.0} & \textbf{83.8}\fstd{0.0} & 83.8\fstd{0.0} & 83.8\fstd{0.0} & \textbf{83.8}\fstd{0.0} \\
Control steering$^{\ddag}$ & \textbf{83.9}\fstd{0.1}\fup{0.1} & \textbf{83.8}\fstd{0.0} & \textbf{83.8}\fstd{0.1} & \textbf{83.9}\fstd{0.0}\fup{0.1} & \textbf{83.9}\fstd{0.0}\fup{0.1} & \textbf{83.8}\fstd{0.1} \\
\midrule
POPE adversarial (unsteered) & \textbf{82.4}\fstd{0.0} & \textbf{82.4}\fstd{0.0} & \textbf{82.4}\fstd{0.0} & \textbf{82.4}\fstd{0.0} & \textbf{82.4}\fstd{0.0} & \textbf{82.4}\fstd{0.0} \\
Control steering$^{\ddag}$ & \textbf{82.4}\fstd{0.0} & \textbf{82.4}\fstd{0.0} & \textbf{82.4}\fstd{0.0} & \textbf{82.4}\fstd{0.0} & \textbf{82.4}\fstd{0.0} & \textbf{82.4}\fstd{0.1} \\
\bottomrule\end{tabular}
}
\end{table*}

\begin{table*}[tp]\centering
\caption{\textbf{F1 ($\%$) of the answer ``yes'' of the control runs of project steering on the general VQA datasets for InternVL-3.5-4B.} Layout as in Table~\ref{tab:f1-control-vqa-gemma}. $^{\ddag}$Control: the projective steering at the same strength ($\alpha = 2.75$) applied to 12 randomly drawn heads (as many as the circuit has), drawn per seed from the heads outside the circuit.}
\label{tab:f1-control-vqa-internvl}
\small
\fithalves{%
\begin{tabular}{l cccccc}\toprule
 & \multicolumn{6}{c}{2D Shapes}\\
\cmidrule(lr){2-7}
 & \multicolumn{2}{c}{dots} & \multicolumn{2}{c}{squares} & \multicolumn{2}{c}{triangles} \\
\cmidrule(lr){2-3}\cmidrule(lr){4-5}\cmidrule(lr){6-7}
Method & L--R & A--B & L--R & A--B & L--R & A--B \\
\midrule
GQA (unsteered) & \textbf{74.6}\fstd{0.0} & \textbf{74.6}\fstd{0.0} & \textbf{74.6}\fstd{0.0} & \textbf{74.6}\fstd{0.0} & \textbf{74.6}\fstd{0.0} & \textbf{74.6}\fstd{0.0} \\
Control steering$^{\ddag}$ & 73.9\fstd{0.3}\fdown{0.7} & 73.9\fstd{0.5}\fdown{0.7} & 74.0\fstd{0.2}\fdown{0.6} & 74.0\fstd{0.3}\fdown{0.6} & 74.0\fstd{0.2}\fdown{0.6} & 74.0\fstd{0.5}\fdown{0.6} \\
\midrule
VQAv2 (unsteered) & 88.9\fstd{0.0} & \textbf{88.9}\fstd{0.0} & \textbf{88.9}\fstd{0.0} & \textbf{88.9}\fstd{0.0} & \textbf{88.9}\fstd{0.0} & \textbf{88.9}\fstd{0.0} \\
Control steering$^{\ddag}$ & \textbf{89.1}\fstd{0.1}\fup{0.2} & \textbf{88.9}\fstd{0.1} & \textbf{88.9}\fstd{0.2} & 88.8\fstd{0.1}\fdown{0.1} & \textbf{88.9}\fstd{0.1} & \textbf{88.9}\fstd{0.2} \\
\midrule
POPE random (unsteered) & \textbf{91.8}\fstd{0.0} & 91.8\fstd{0.0} & \textbf{91.8}\fstd{0.0} & \textbf{91.8}\fstd{0.0} & 91.8\fstd{0.0} & \textbf{91.8}\fstd{0.0} \\
Control steering$^{\ddag}$ & \textbf{91.8}\fstd{0.1} & \textbf{92.0}\fstd{0.1}\fup{0.2} & \textbf{91.8}\fstd{0.1} & \textbf{91.8}\fstd{0.1} & \textbf{91.9}\fstd{0.2}\fup{0.1} & \textbf{91.8}\fstd{0.1} \\
\midrule
POPE popular (unsteered) & \textbf{87.1}\fstd{0.0} & \textbf{87.1}\fstd{0.0} & \textbf{87.1}\fstd{0.0} & \textbf{87.1}\fstd{0.0} & \textbf{87.1}\fstd{0.0} & \textbf{87.1}\fstd{0.0} \\
Control steering$^{\ddag}$ & 87.0\fstd{0.1}\fdown{0.1} & \textbf{87.1}\fstd{0.1} & 87.0\fstd{0.1}\fdown{0.1} & 87.0\fstd{0.1}\fdown{0.1} & \textbf{87.1}\fstd{0.2} & \textbf{87.1}\fstd{0.1} \\
\midrule
POPE adversarial (unsteered) & \textbf{83.6}\fstd{0.0} & \textbf{83.6}\fstd{0.0} & \textbf{83.6}\fstd{0.0} & \textbf{83.6}\fstd{0.0} & \textbf{83.6}\fstd{0.0} & \textbf{83.6}\fstd{0.0} \\
Control steering$^{\ddag}$ & 83.4\fstd{0.1}\fdown{0.2} & \textbf{83.6}\fstd{0.1} & 83.4\fstd{0.1}\fdown{0.2} & 83.5\fstd{0.2}\fdown{0.1} & \textbf{83.6}\fstd{0.2} & 83.5\fstd{0.0}\fdown{0.1} \\
\bottomrule\end{tabular}
}{%
\begin{tabular}{l ccc ccc}\toprule
 & \multicolumn{3}{c}{3D Objects} & \multicolumn{3}{c}{WhatsUp}\\
\cmidrule(lr){2-4}\cmidrule(lr){5-7}
Method & L--R & A--B & F--B & L--R & A--B & F--B \\
\midrule
GQA (unsteered) & \textbf{74.2}\fstd{0.0} & \textbf{74.2}\fstd{0.0} & \textbf{74.2}\fstd{0.0} & \textbf{74.3}\fstd{0.0} & \textbf{74.3}\fstd{0.0} & \textbf{74.3}\fstd{0.0} \\
Control steering$^{\ddag}$ & \textbf{74.2}\fstd{0.3} & 73.9\fstd{0.5}\fdown{0.3} & 73.9\fstd{0.3}\fdown{0.3} & 74.2\fstd{0.3}\fdown{0.1} & 73.9\fstd{0.2}\fdown{0.4} & 73.9\fstd{0.2}\fdown{0.4} \\
\midrule
VQAv2 (unsteered) & \textbf{88.9}\fstd{0.0} & 88.9\fstd{0.0} & \textbf{88.9}\fstd{0.0} & \textbf{88.9}\fstd{0.0} & 88.9\fstd{0.0} & \textbf{88.9}\fstd{0.0} \\
Control steering$^{\ddag}$ & 88.8\fstd{0.1}\fdown{0.1} & \textbf{89.0}\fstd{0.2}\fup{0.1} & \textbf{88.9}\fstd{0.2} & \textbf{88.9}\fstd{0.2} & \textbf{89.0}\fstd{0.3}\fup{0.1} & \textbf{88.9}\fstd{0.2} \\
\midrule
POPE random (unsteered) & \textbf{91.8}\fstd{0.0} & \textbf{91.8}\fstd{0.0} & \textbf{91.8}\fstd{0.0} & \textbf{91.8}\fstd{0.0} & \textbf{91.8}\fstd{0.0} & \textbf{91.8}\fstd{0.0} \\
Control steering$^{\ddag}$ & \textbf{91.8}\fstd{0.1} & \textbf{91.8}\fstd{0.1} & \textbf{91.8}\fstd{0.1} & \textbf{91.8}\fstd{0.1} & \textbf{91.8}\fstd{0.1} & \textbf{91.8}\fstd{0.1} \\
\midrule
POPE popular (unsteered) & \textbf{87.1}\fstd{0.0} & \textbf{87.1}\fstd{0.0} & 87.1\fstd{0.0} & \textbf{87.1}\fstd{0.0} & \textbf{87.1}\fstd{0.0} & \textbf{87.1}\fstd{0.0} \\
Control steering$^{\ddag}$ & \textbf{87.1}\fstd{0.1} & \textbf{87.1}\fstd{0.1} & \textbf{87.2}\fstd{0.1}\fup{0.1} & \textbf{87.1}\fstd{0.1} & 86.9\fstd{0.2}\fdown{0.2} & \textbf{87.1}\fstd{0.2} \\
\midrule
POPE adversarial (unsteered) & \textbf{83.6}\fstd{0.0} & \textbf{83.6}\fstd{0.0} & 83.6\fstd{0.0} & \textbf{83.6}\fstd{0.0} & \textbf{83.6}\fstd{0.0} & \textbf{83.6}\fstd{0.0} \\
Control steering$^{\ddag}$ & 83.5\fstd{0.1}\fdown{0.1} & \textbf{83.6}\fstd{0.1} & \textbf{83.7}\fstd{0.1}\fup{0.1} & 83.5\fstd{0.1}\fdown{0.1} & 83.5\fstd{0.2}\fdown{0.1} & \textbf{83.6}\fstd{0.2} \\
\bottomrule\end{tabular}
}
\end{table*}

\begin{table*}[tp]\centering
\caption{\textbf{F1 (in $\%$) of the answer ``yes'' of the control runs of projective steering on the general VQA datasets for Ministral-3-3B.}  $^{\ddag}$Control: the projective steering at the same strength ($\alpha = 2.75$) applied to $7$ randomly drawn heads (as many as the circuit has), drawn per seed from the heads outside the circuit.}
\label{tab:f1-control-vqa-ministral}
\small
\fithalves{%
\begin{tabular}{l cccccc}\toprule
 & \multicolumn{6}{c}{2D Shapes}\\
\cmidrule(lr){2-7}
 & \multicolumn{2}{c}{dots} & \multicolumn{2}{c}{squares} & \multicolumn{2}{c}{triangles} \\
\cmidrule(lr){2-3}\cmidrule(lr){4-5}\cmidrule(lr){6-7}
Method & L--R & A--B & L--R & A--B & L--R & A--B \\
\midrule
GQA (unsteered) & 56.5\fstd{0.0} & 56.5\fstd{0.0} & 56.5\fstd{0.0} & 56.5\fstd{0.0} & 56.5\fstd{0.0} & 56.5\fstd{0.0} \\
Control steering$^{\ddag}$ & \textbf{56.9}\fstd{0.6}\fup{0.4} & \textbf{57.0}\fstd{1.6}\fup{0.5} & \textbf{57.4}\fstd{0.7}\fup{0.9} & \textbf{57.1}\fstd{1.3}\fup{0.6} & \textbf{57.4}\fstd{0.6}\fup{0.9} & \textbf{57.1}\fstd{1.2}\fup{0.6} \\
\midrule
VQAv2 (unsteered) & \textbf{82.2}\fstd{0.0} & \textbf{82.2}\fstd{0.0} & \textbf{82.2}\fstd{0.0} & \textbf{82.2}\fstd{0.0} & 82.2\fstd{0.0} & \textbf{82.2}\fstd{0.0} \\
Control steering$^{\ddag}$ & \textbf{82.2}\fstd{0.2} & 82.1\fstd{0.5}\fdown{0.1} & \textbf{82.2}\fstd{0.2} & 82.1\fstd{0.4}\fdown{0.1} & \textbf{82.3}\fstd{0.4}\fup{0.1} & \textbf{82.2}\fstd{0.4} \\
\midrule
POPE random (unsteered) & 81.6\fstd{0.0} & 81.6\fstd{0.0} & 81.6\fstd{0.0} & 81.6\fstd{0.0} & 81.6\fstd{0.0} & 81.6\fstd{0.0} \\
Control steering$^{\ddag}$ & \textbf{81.9}\fstd{0.2}\fup{0.3} & \textbf{82.0}\fstd{0.6}\fup{0.4} & \textbf{82.1}\fstd{0.3}\fup{0.5} & \textbf{81.9}\fstd{0.6}\fup{0.3} & \textbf{82.1}\fstd{0.4}\fup{0.5} & \textbf{82.0}\fstd{0.6}\fup{0.4} \\
\midrule
POPE popular (unsteered) & 80.6\fstd{0.0} & 80.6\fstd{0.0} & 80.6\fstd{0.0} & 80.6\fstd{0.0} & 80.6\fstd{0.0} & 80.6\fstd{0.0} \\
Control steering$^{\ddag}$ & \textbf{80.8}\fstd{0.2}\fup{0.2} & \textbf{80.8}\fstd{0.4}\fup{0.2} & \textbf{80.9}\fstd{0.2}\fup{0.3} & \textbf{80.8}\fstd{0.5}\fup{0.2} & \textbf{81.0}\fstd{0.3}\fup{0.4} & \textbf{80.8}\fstd{0.5}\fup{0.2} \\
\midrule
POPE adversarial (unsteered) & 79.4\fstd{0.0} & 79.4\fstd{0.0} & 79.4\fstd{0.0} & 79.4\fstd{0.0} & 79.4\fstd{0.0} & 79.4\fstd{0.0} \\
Control steering$^{\ddag}$ & \textbf{79.5}\fstd{0.2}\fup{0.1} & \textbf{79.6}\fstd{0.4}\fup{0.2} & \textbf{79.7}\fstd{0.2}\fup{0.3} & \textbf{79.6}\fstd{0.4}\fup{0.2} & \textbf{79.7}\fstd{0.3}\fup{0.3} & \textbf{79.7}\fstd{0.4}\fup{0.3} \\
\bottomrule\end{tabular}
}{%
\begin{tabular}{l ccc ccc}\toprule
 & \multicolumn{3}{c}{3D Objects} & \multicolumn{3}{c}{WhatsUp}\\
\cmidrule(lr){2-4}\cmidrule(lr){5-7}
Method & L--R & A--B & F--B & L--R & A--B & F--B \\
\midrule
GQA (unsteered) & \textbf{56.3}\fstd{0.0} & \textbf{56.3}\fstd{0.0} & 56.3\fstd{0.0} & \textbf{56.3}\fstd{0.0} & 56.3\fstd{0.0} & \textbf{56.3}\fstd{0.0} \\
Control steering$^{\ddag}$ & 55.8\fstd{1.8}\fdown{0.5} & \textbf{56.3}\fstd{2.2} & \textbf{56.4}\fstd{0.5}\fup{0.1} & 55.7\fstd{1.1}\fdown{0.6} & \textbf{57.1}\fstd{1.2}\fup{0.8} & 56.2\fstd{0.8}\fdown{0.1} \\
\midrule
VQAv2 (unsteered) & \textbf{82.2}\fstd{0.0} & \textbf{82.2}\fstd{0.0} & \textbf{82.2}\fstd{0.0} & \textbf{82.2}\fstd{0.0} & \textbf{82.2}\fstd{0.0} & \textbf{82.2}\fstd{0.0} \\
Control steering$^{\ddag}$ & 81.9\fstd{0.3}\fdown{0.3} & 81.8\fstd{1.0}\fdown{0.4} & 82.0\fstd{0.1}\fdown{0.2} & 81.7\fstd{0.2}\fdown{0.5} & 81.9\fstd{0.5}\fdown{0.3} & 82.0\fstd{0.1}\fdown{0.2} \\
\midrule
POPE random (unsteered) & \textbf{81.6}\fstd{0.0} & 81.6\fstd{0.0} & 81.6\fstd{0.0} & \textbf{81.6}\fstd{0.0} & 81.6\fstd{0.0} & 81.6\fstd{0.0} \\
Control steering$^{\ddag}$ & 81.5\fstd{0.3}\fdown{0.1} & \textbf{81.9}\fstd{0.8}\fup{0.3} & \textbf{81.9}\fstd{0.2}\fup{0.3} & 81.5\fstd{0.1}\fdown{0.1} & \textbf{81.8}\fstd{0.7}\fup{0.2} & \textbf{81.7}\fstd{0.2}\fup{0.1} \\
\midrule
POPE popular (unsteered) & \textbf{80.6}\fstd{0.0} & 80.6\fstd{0.0} & 80.6\fstd{0.0} & \textbf{80.6}\fstd{0.0} & 80.6\fstd{0.0} & \textbf{80.6}\fstd{0.0} \\
Control steering$^{\ddag}$ & 80.5\fstd{0.3}\fdown{0.1} & \textbf{80.7}\fstd{0.6}\fup{0.1} & \textbf{80.7}\fstd{0.2}\fup{0.1} & 80.5\fstd{0.1}\fdown{0.1} & \textbf{80.7}\fstd{0.6}\fup{0.1} & \textbf{80.6}\fstd{0.2} \\
\midrule
POPE adversarial (unsteered) & \textbf{79.4}\fstd{0.0} & 79.4\fstd{0.0} & 79.4\fstd{0.0} & \textbf{79.4}\fstd{0.0} & 79.4\fstd{0.0} & 79.4\fstd{0.0} \\
Control steering$^{\ddag}$ & 79.3\fstd{0.1}\fdown{0.1} & \textbf{79.6}\fstd{0.5}\fup{0.2} & \textbf{79.6}\fstd{0.1}\fup{0.2} & 79.3\fstd{0.1}\fdown{0.1} & \textbf{79.6}\fstd{0.5}\fup{0.2} & \textbf{79.5}\fstd{0.1}\fup{0.1} \\
\bottomrule\end{tabular}
}
\end{table*}